\documentclass[10pt,twocolumn,letterpaper]{article}

\usepackage[pagenumbers]{iccv} 

\definecolor{iccvblue}{rgb}{0.21,0.49,0.74}
\usepackage[pagebackref,breaklinks,colorlinks,allcolors=iccvblue]{hyperref}

\usepackage{array}
\usepackage{booktabs}
\usepackage{multirow}
\usepackage{makecell}
\usepackage{pifont}
\newcommand{\cmark}{\ding{51}}%
\newcommand{\xmark}{\ding{55}}%
\usepackage{rotating}
\usepackage{tikz}
\usepackage{graphicx}
\usepackage{hyperref}
\hypersetup{
    colorlinks=true,            
    linkcolor=blue,             
    filecolor=magenta,          
    urlcolor=cyan,              
    pdftitle={Overleaf Example},
    pdfpagemode=FullScreen,
    }
\usepackage[misc]{ifsym}

\usepackage{algorithm}
\usepackage{listings}
\usepackage{xcolor}
\usepackage{tocloft}
\usepackage{caption}
\makeatletter
\def\thanks#1{\protected@xdef\@thanks{\@thanks
		\protect\footnotetext{#1}}}
\makeatother

\def\paperID{****} 
\def\confName{ICCV}
\def\confYear{2025}

\title{Falcon: A Remote Sensing Vision-Language Foundation Model \\(Technical Report)}

\author{Kelu Yao$^{\ast}$, Nuo Xu$^{\ast}$, Rong Yang$^{\ast}$, Yingying Xu$^{\ast}$, \\ Zhuoyan Gao, Titinunt Kitrungrotsakul, Yi Ren, Pu Zhang, Jin Wang, Ning Wei, Chao Li $^\textrm{\Letter}$ \thanks{$\ast$: Equal contribution. $\textrm{\Letter}$: Corresponding author.}\\
Research Center for Space Computing System, ZhejiangLab, Hangzhou, China\\
{\tt\small \{yaokelu, nuo.xu, yang\_rong, cs\_ying, lichao\}@zhejianglab.org}
}

\begin{document}
\maketitle

\begin{abstract}
This paper introduces a holistic vision-language foundation model tailored for remote sensing, named Falcon.
Falcon offers a unified, prompt-based paradigm that effectively executes comprehensive and complex remote sensing tasks. 
Falcon demonstrates powerful understanding and reasoning abilities at the image, region, and pixel levels.
Specifically, given simple natural language instructions and remote sensing images, Falcon can produce impressive results in text form across 14 distinct tasks, i.e., image classification, object detection, segmentation, image captioning, and etc.
To facilitate Falcon's training and empower its representation capacity to encode rich spatial and semantic information, we developed Falcon\_SFT, a large-scale, multi-task, instruction-tuning dataset in the field of remote sensing. 
The Falcon\_SFT dataset consists of approximately 78 million high-quality data samples, covering 5.6 million multi-spatial resolution and multi-view remote sensing images with diverse instructions. 
It features hierarchical annotations and undergoes manual sampling verification to ensure high data quality and reliability.
Extensive comparative experiments are conducted, which verify that Falcon achieves remarkable performance over 67 datasets and 14 tasks, despite having only 0.7B parameters.
We release the complete dataset, code, and model weights at \href{https://github.com/TianHuiLab/Falcon}{https://github.com/TianHuiLab/Falcon}, hoping to help further develop the open-source community.
\end{abstract}    
\addtocontents{toc}{\protect\setcounter{tocdepth}{-1}}
\section{Introduction}
\label{sec:intro}
Large vision language models (LVLMs) have demonstrated remarkable success in various vision-language tasks on natural images \cite{clip,flamingo,llava,zhu2023minigpt,dai2023instructblip}.
However, due to the significant domain and embedded knowledge gap between the natural images and remote sensing images, developing a remote sensing foundational vision-language model remains a substantial challenge.
To this end, previous studies \cite{kuckreja2024geochat,muhtar2024lhrs,liu2024remoteclip,zhang2024rs5m,hu2023rsgpt} usually focused on learning vision-language models that excel in specific remote sensing tasks, limiting their adaptability for more diverse and complex scenarios.
With the ongoing advancement of Artificial General Intelligence (AGI) systems, creating a foundational remote sensing model with comprehensive understanding and reasoning capabilities is of significant value.


\begin{figure}[tbp]
  \centering
   \includegraphics[width=1\linewidth]{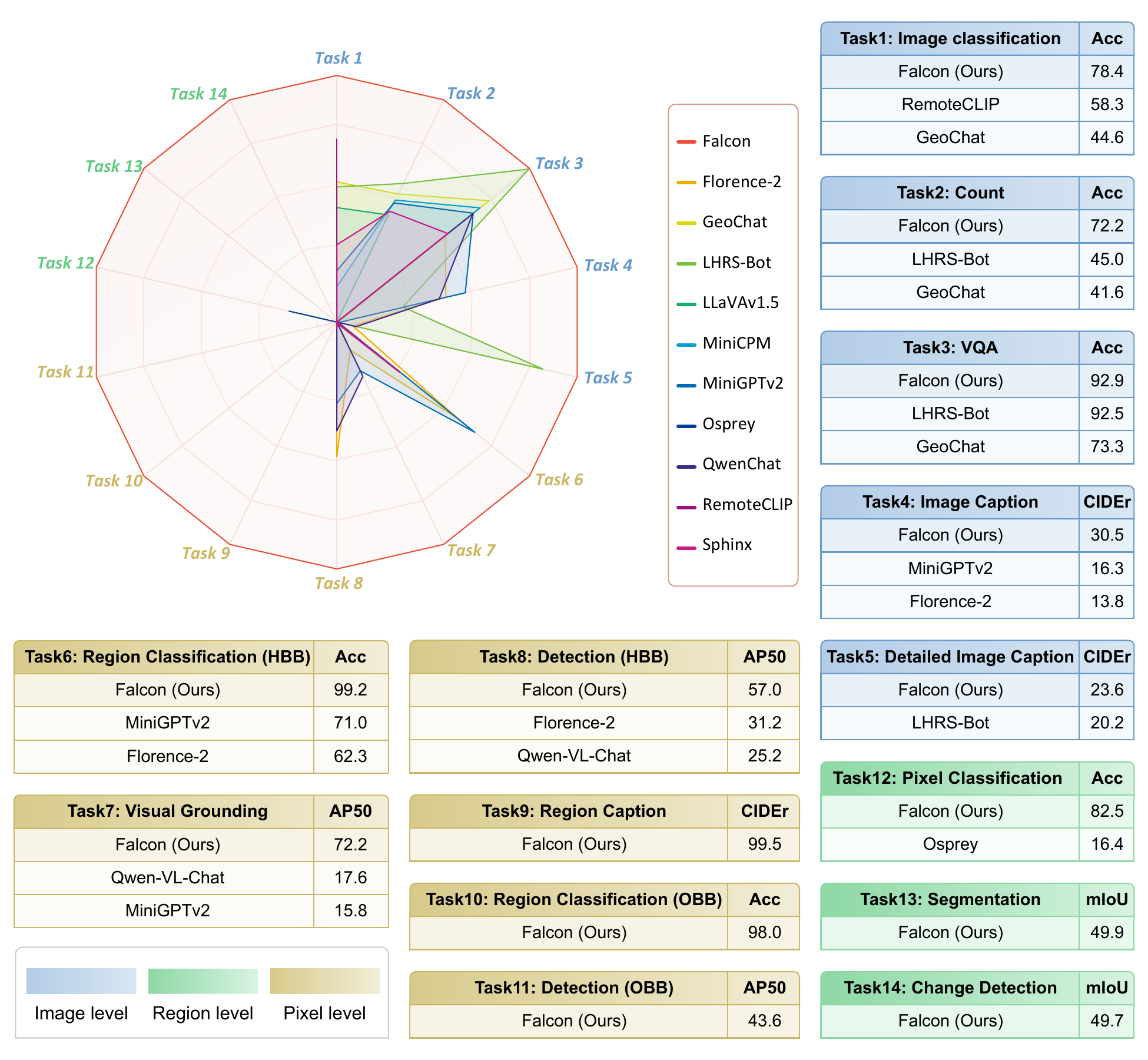}
   \vspace{-7mm}
   \caption{An overall performance comparison between Falcon and 10 state-of-the-art models across 14 remote sensing tasks at image, region, and pixel levels. Results demonstrate that Falcon outperformed existing models, showcasing superior and more comprehensive understanding and reasoning capabilities.}
   \vspace{-5mm}
   \label{fig:first_image}
\end{figure}

However, attaining such a foundational remote sensing model still faces significant challenges, which we summarize as follows: 
1) Existing models did not feature a universal representation for diverse remote sensing tasks, 
often failing to facilitate the learning of comprehensive perceptual and reasoning abilities;
2) The absence of a large-scale, high-quality, multi-task dataset for training also limits the ability of current remote sensing models to learn robust and generalized representations.

\begin{figure*}[tbp]
  \centering
   \includegraphics[width=0.85\linewidth]{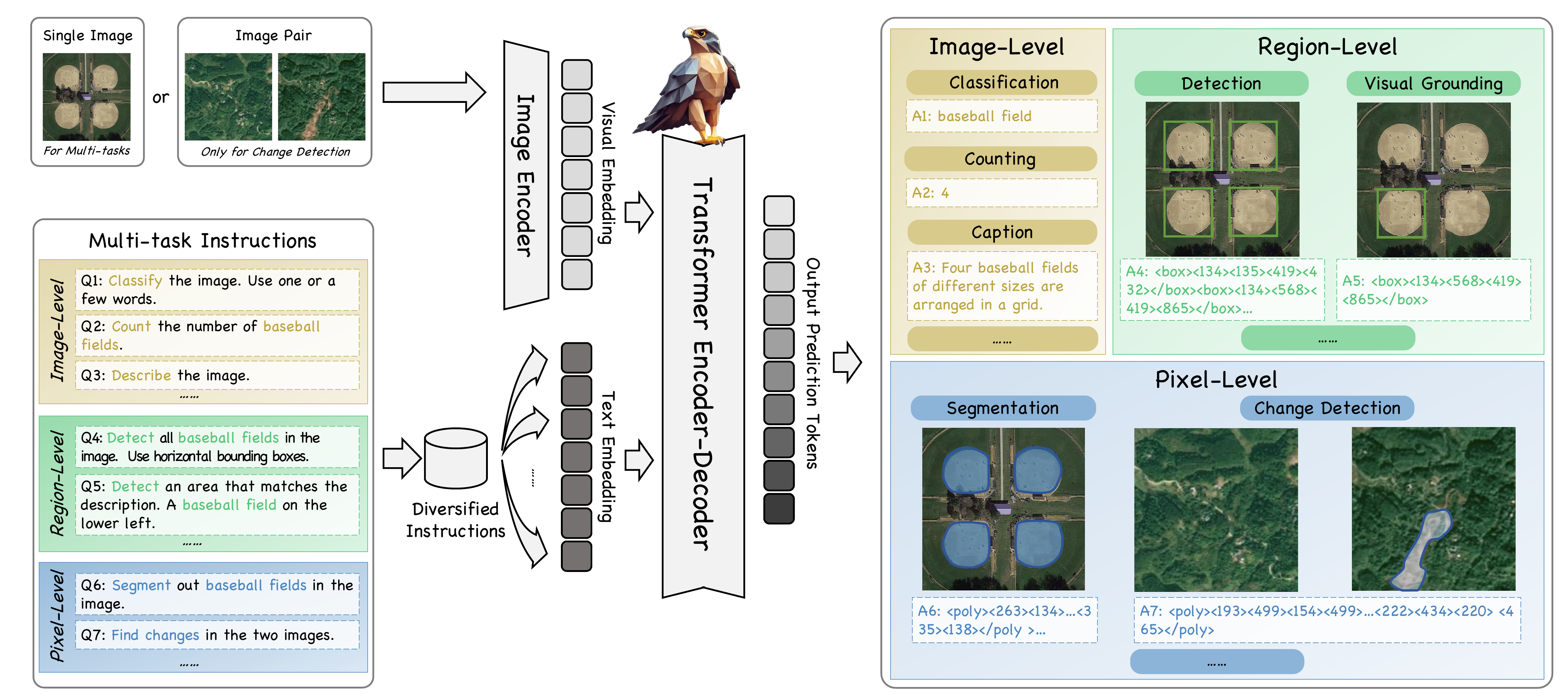}

   \vspace{-2mm}
   \caption{The overview of Falcon model architecture. Given a single image or an image pair (for the task of change detection), Falcon can follow diverse multi-task instructions, generating a universal textual representation suitable for various remote sensing tasks. As shown in the figure, Falcon correctly distinguishes the category of the given image, provides the spatial bounding boxes/segmentations masks for the given objects and even detects subtle changes across images, highlighting its comprehensive capabilities for remote sensing.}
   \label{fig:model_arc}
   \vspace{-4mm}
\end{figure*}

To address the above challenges, we first propose Falcon, a versatile vision-language foundation model with comprehensive perceptual and reasoning abilities tailored for remote sensing.
In particular, Falcon features a unified architecture for multitask learning, bridging image-level, region-level, and pixel-level reasoning and understanding abilities in one model.
To the best of our knowledge, Falcon is the first remote sensing VLM capable of performing 14 diverse understanding and reasoning tasks across image, region, and pixel levels simultaneously. 
We hereby provide an ability comparison among various remote sensing VLMs and Falcon in Tab.\;\ref{vlms feature}.
Compared with Falcon, previous models like GeoChat \cite{kuckreja2024geochat} and RSGPT \cite{hu2023rsgpt} can only support a limited scope of remote sensing tasks, narrowing their application scenarios.

The crucial challenge for designing Falcon is learning universal representation for diverse remote sensing tasks.
Inspired by the latest research in natural image area \cite{xiao2024florence, wang2023visionllmlargelanguagemodel, OMGLLaVA, wu2024visionllm2}, we utilize a unified network architecture to seamlessly integrate spatial hierarchy and sematic granularity information into a universal representation.
The architecture consists of an image encoder and a multi-modality encoder-decoder.
This design aligns the vision and language representations, and offers a unified framework to various remote sensing tasks without additional module designs.
Besides, to further enhance the instruction understanding capability of Falcon, we propose a dynamic prompt training strategy that leverages multiple differently phrased versions of each instruction.
In this way, given user's prompts and remote sensing images, Falcon can produce results in a unified textual form across a wide range of tasks, \textit{e.g.}, image classification, object detection, segmentation, image captioning, change detection, and etc.

\begin{figure*}[htbp]
  \centering
   \includegraphics[width=0.93\linewidth]{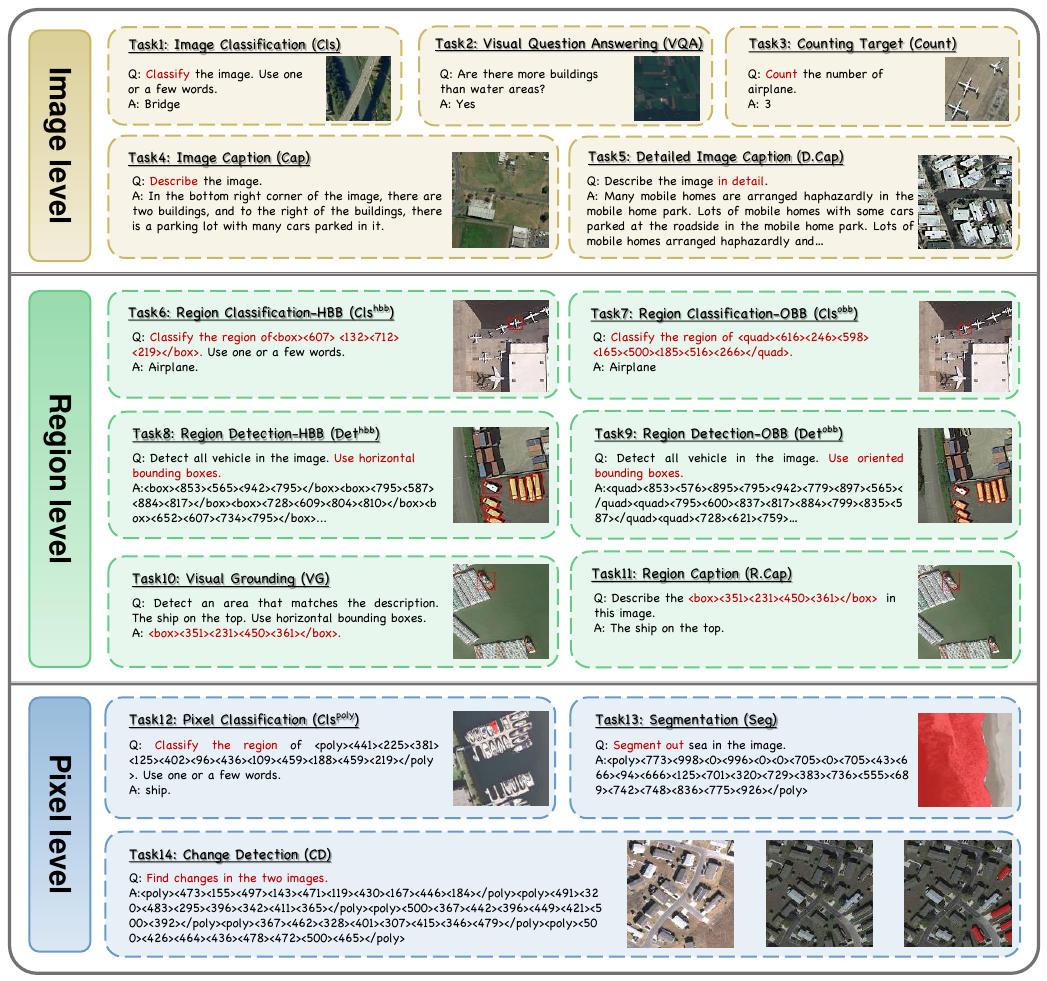}
    \vspace{-2mm}
   \caption{An illustrative example of images, their corresponding instructions, and output format of different tasks in Falcon\_SFT dataset.}
   \vspace{-4mm}
   \label{fig:dataset_vis}
\end{figure*}

Moreover, to facilitate Falcon’s training, we further develop Falcon\_SFT, a large-scale, multi-task instruction-tuning dataset. 
Early remote sensing datasets \cite{dota1,dota2,millionAID} usually focused on a single or a few \textit{vision} tasks. 
Recent studies proposed \textit{mutlimodal} remote sensing datasets suitable for training vision-language models.
However, these datasets often contain a limited number of image-text pairs, making them only useful for training models on specific tasks \cite{zhang2024rs5m, hu2023rsgpt,zhan2024skyeyegpt}.
Therefore, we present Falcon\_SFT, a large-scale multi-task instruction-tuning dataset.
The Falcon\_SFT dataset consists of approximately 78 million high-quality data samples, covering 5.6 million multi-spatial resolution and multi-view remote sensing images.
Specifically, we uniformly standardize each sample in the Falcon\_SFT dataset into a unified format, facilitating the training of our proposed Falcon.
Please see Fig.\;\ref{fig:dataset_vis} for data examples.

\begin{table*}[t] 
	\centering
	\small
    {
    \linespread{1.0}
\setlength\tabcolsep{6pt}
\scriptsize
    {

		\begin{tabular}{l | c |c c c c c | c c c c c c| c c c}
			\toprule

   		& & \multicolumn{5}{c|}{Image level} &  \multicolumn{6}{c|}{Region level} & \multicolumn{3}{c}{Pixel level}\\
            \cmidrule(lr){3-7}
			\cmidrule(lr){8-13}
   		\cmidrule(lr){14-16}
			& Models & Cls & Cap & D. Cap & Count & VQA & Cls$^{hbb}$ & Cls$^{obb}$ & R.Cap & Det$^{hbb}$ & Det$^{obb}$ & VG & Cls$^{poly}$ & Seg & CD \\
			\midrule

     
   
   
     
            
            \multirow{16}{*}{\rotatebox[origin=c]{90}{\centering Remote Sensing VLMs}}
			& GeoChat\cite{kuckreja2024geochat}  & {\cmark} & {\cmark} &  {\cmark} & {\cmark} & {\cmark} & & & {\cmark}& & &{\cmark}  & & &  \\		
			& GeoRSCLIP\cite{zhang2024rs5m} & {\cmark} &  &   &  &  & & & & {\cmark}& {\cmark}& {\cmark} & &{\cmark} &  \\
			& LHRS-Bot\cite{muhtar2024lhrs}& {\cmark} & {\cmark} & {\cmark} & {\cmark} & {\cmark} & & & & & & {\cmark} & &  &  \\
			& RemoteCLIP\cite{liu2024remoteclip} & {\cmark} &  &   & {\cmark} &  &  &  &  &  &  &  &  &  &  \\
   		& SkyCLIP\cite{wang2024skyscript} & {\cmark} &  &   &  &  & & & & &  &  & & &  \\
			& RSGPT\cite{hu2023rsgpt} & {\cmark} & {\cmark} &  {\cmark} & {\cmark} & {\cmark} & & & & & &  & & &  \\
			& GRAFT \cite{mallremote} & {\cmark} & {\cmark} & {\cmark} & {\cmark} & {\cmark} & & & & & &  & & {\cmark} & \\
			& EarthGPT\cite{zhang2024earthgpt} & {\cmark} & {\cmark} &  {\cmark} & {\cmark} & {\cmark} & & & {\cmark} & {\cmark}& {\cmark}& {\cmark}  & & &  \\
			& RS-ChatGPT \cite{rschatgpt} & {\cmark} & {\cmark} & {\cmark} & {\cmark} & {\cmark} & & & & & &  & & {\cmark}&  \\
   		& SkyEyeGPT\cite{zhan2024skyeyegpt} & {\cmark} & {\cmark} &  {\cmark} & {\cmark} & {\cmark} & & & & & & {\cmark} & & &  \\
			& RS-CapRat \cite{silva2024large} &  & {\cmark} & {\cmark} & {\cmark} & {\cmark} & & & & & &  & & &  \\
			& Popeye \cite{zhang2024popeye} &  & {\cmark} & {\cmark} & {\cmark} & {\cmark} & & & & {\cmark} & {\cmark} &  & & {\cmark} &  \\
			& MGIMM \cite{yang2024mgimm} &  & {\cmark} & {\cmark} & {\cmark} & {\cmark} & & & {\cmark} & & &  & & &  \\
			& EarthMarker \cite{zhang2024earthmarker} & {\cmark} & {\cmark} & {\cmark} & {\cmark} & {\cmark} & {\cmark} & & {\cmark} & & &  & {\cmark} & &  \\
		  & \textbf{Falcon(Ours)} & {\cmark} & {\cmark} & {\cmark} & {\cmark} & {\cmark} & {\cmark} & {\cmark} &{\cmark} & {\cmark} & {\cmark} & {\cmark} & {\cmark} & {\cmark} & {\cmark} \\
			\bottomrule
	\end{tabular}
    }
    }
	\caption{Comparisons of capabilities of different remote sensing vision-language models. Several representative models have been included in this table. 
    Notably, Falcon exhibits the most comprehensive understanding and reasoning capabilities, covering image, region, and pixel levels comprehensively.
    For task abbreviations in the second row, please see Fig.\;\ref{fig:dataset_vis} for details.}
    \vspace{-4mm}
    \label{vlms feature}
\end{table*}

In experiments, we conduct a variety of evaluations of our proposed Falcon both qualitatively and quantitatively (see Fig.\;\ref{fig:first_image} for a quick preview).
For qualitative evaluations, we visualize the prediction results of 14 tasks individually and compare with other state-of-the-art methods, in order to evaluate the performance of Falcon.
For quantitative evaluations, we assess the performance of Falcon on each downstream task, along with its zero-shot performance on unseen data samples, highlighting the generalization ability of Falcon.
Beside, we conduct detailed ablation studies for Falcon, showcasing the effectiveness of our training recipes.

Finally, to address the critical absence of a high-performance foundational model for remote sensing in the community, we will fully \textit{open-source} our work with complete dataset, code and model weights, aiming to bridge the gap between foundational models for remote sensing imagery and foundational models for natural imagery.
Despite the substantial financial investment of our proposed Falcon, we hope this effort will foster further research and development in the field, advancing the capabilities of remote sensing models and their real-world applications.

Contributions of this paper can be summarized as follows. 
1) To the best of our knowledge, Falcon is the first remote sensing vision-language model to feature image, region, and pixel-level understanding and reasoning capabilities, supporting 14 tasks within a unified architecture. 
2) As of March 2025, Falcon\_SFT stands as the largest and most comprehensive dataset for training vision-language models in the remote sensing field.
3) We have conducted extensive experiments to demonstrate the superiority of Falcon over previous VLMs, highlighting the effectiveness of Falcon and Falcon\_SFT in the field of remote sensing. The complete dataset, code, and model weights will be fully open-sourced to the community.


\section{Related work}
\label{sec:formatting}

\subsection{Remote sensing datasets}
The development of high-quality remote sensing datasets has attracted increasing attention in recent years. 
Previous studies on this field mainly focused on two perspectives.
Some studies \cite{DIOR,dota1,dota2,2022FAIR1M} focused on image datasets each targeting a single or a few vision tasks. 
Long \textit{et al.} \cite{millionAID} proposed Million-AID, a large-scale image dataset containing 51 categories and a million instances for remote sensing scene classification.
G. Sumbul \textit{et al.} \cite{bigearthnet} introduced BigEarthNet, comprising 590,326 images collected from Sentinel-1 and Sentinel-2 satellites, featuring several resolutions and image sizes.
The DOTA series datasets \cite{dota1,dota2} were mainly sourced from Google Earth, the GF-2 satellite, and aerial images, which have greatly advanced the field of object detection. 
The latest version \cite{dota2} featured 11,268 images, 18 categories, and an extensive set of annotations with oriented bounding boxes. 
Jacob Shermeyer \textit{et al.} \cite{RarePlanes} proposed RarePlanes dataset in order to improve the performance of detecting aircraft and their attributes in satellite imagery.
GID \cite{GID}, UAVid\cite{uavid}, DLRSD\cite{DLRSD} were commonly used datasets for the semantic segmentation task of RGB remote sensing images.

Besides, several studies \cite{LRBEN_HRBEN, RSVG, RSITMD, RSICD} have developed multimodal datasets to support vision-language models in remote sensing.
Dilxat Muhtar \textit{et al.} \cite{muhtar2024lhrs} developed LHRS-Align, which included 0.9K samples for visual reasoning, 4K samples for detailed image descriptions, and 7K samples for conversational tasks. 
However, to use this dataset, users must download the original images from Google Earth imagery.
RSICD \cite{RSICD}, Sydney-Captions \cite{UCMCaptions_SydneyCaptions}, UCM-Captions \cite{UCMCaptions_SydneyCaptions}, NWPU-Captions \cite{nwpu-caption} were datasets specifically created for remote sensing image caption generation tasks, containing 10921, 613, 2000, 31500 images, each accompanied by descriptions of varying lengths.

Despite previous advancements, existing remote sensing datasets remained limited in terms of data scale, task diversity, hierarchical annotation, and annotation quality. 
The field still lacked a large-scale, multi-task dataset suitable for training foundational vision-language models, hindering their progress. 
To address this challenge, we present Falcon\_SFT in this paper, a comprehensive, large-scale, multi-task instruction-tuning dataset for remote sensing.
Specifically, we compiled 67 remote sensing datasets covering a variety of tasks, please refer to supplementary material for details.

\begin{table}[t]
\centering
\small
\resizebox{1\columnwidth}{!}{
    \begin{tabular}{c c c c c c c c c}
    \toprule
    Dataset & Rep.VLM & \#Images & \#Annotations & \#tasks & Spatial hierarchy \\
    \midrule
    RS5M & GeoRSCLIP \cite{zhang2024rs5m} & - & 5M & 1 & Image-level\\
    LHRS-Align & LHRS-Bot \cite{muhtar2024lhrs} & 1.15M & 1.15M & 1 & Image-level\\
    RSICap & RSGPT \cite{hu2023rsgpt} & 2192 & 2192 & 1 & Image-level\\
    MMRS-1M & EarthGPT \cite{zhang2024earthgpt} & - & 1M & 5 & Image \& Region-level\\
    SkyEye-968k & SkyEyeGPT \cite{zhan2024skyeyegpt} & - & 968k & 4 & Image \& Region-level\\
    RSVP-3M &  EarthMarker \cite{zhang2024earthmarker} & - & 3M & 5 & Image \& Region-level \\
    \textbf{Falcon\_SFT(Ours)} & Falcon(Ours) & 5.6M & 78.2M & 14 & Image \& Region \& Pixel-level \\
    \bottomrule
    \end{tabular}
    }
    \caption{Comparisons with VLMs' remote sensing datasets.}
    \vspace{-4mm}
    \label{tab:datasets_comparisons}
    \end{table}


\subsection{Remote Sensing Foundation Models}
Recently, a considerable literature has grown up around the theme of developing remote sensing foundation models.
These pre-trained foundation models can be categorized based on architectural design.
The first category consists of ViT-based vision foundation models \cite{mall2023change, Scale-mae, li2024s2mae, manas2021seasonal}. 
For instance, Sun \textit{et al.} proposed RingMo \cite{sun2022ringmo}, a classic remote sensing vision model fine-tuning on 4 downstream tasks. 
These methods lacked reasoning abilities and cannot be controlled via natural language instructions.
The second category includes CLIP-based vision-language models \cite{wang2024skyscript, zhang2024rs5m, liu2024remoteclip}. 
For instance, Liu \textit{et al.} proposed RemoteCLIP \cite{liu2024remoteclip}, the first vision-language foundation model for remote sensing that aligned text embeddings for downstream application. 
However, these methods cannot perform different tasks without designing additional modules.
The third category comprises LLM-based vision-language models \cite{kuckreja2024geochat,zhang2024earthgpt,muhtar2024lhrs,zhang2024popeye}. 
Zhan \textit{et al.} proposed SkyEyeGPT \cite{zhan2024skyeyegpt}, specifically designed for remote sensing images understanding. 
Kartik Kuckreja \textit{et al.} \cite{kuckreja2024geochat} introduced GeoChat, a versatile LLaVA-based remote sensing vision-language model, but it cannot perform complex pixel-level tasks such as segmentation or change detection.
Similarly, LHRS-Bot \cite{muhtar2024lhrs} also lacked such capabilities.
Furthermore, these methods often exceeded 7 billion parameters, leading to computational bottlenecks and low inference efficiency when deployed on edge devices. 
More importantly, we believe that the LLMs module containing significant number of parameters may not play an essential role in remote sensing, considering that this task still primarily focuses on the visual input.
Therefore, in this paper, we propose a lightweight vision-language model to efficiently handle various remote sensing tasks in a unified paradigm.






\section{Algorithm}
In this section, we aim to delve into the details of Falcon, introducing a simple yet effective way to address challenges of unifying many complex remote sensing tasks in one manner. 
Specifically, we will introduce the design of Falcon's architecture and a multi-task learning paradigm, that enables the unification of various vision-language tasks.

\textbf{Notation:} Let $\mathcal{I}\in\mathbb{R}^{H\times W\times3}$ denote the input remote sensing image, with $H$ and $W$ denoting the height and width of the image.
$\mathcal{T}$ denotes the input textual prompt.
$y$ denotes the prediction target \textit{i.e.}, the formulated visual annotations.
$\mathcal{G}$ denotes the image encoder.
$\mathcal{E}$ denotes the text token embedding function.
$\mathcal{F}$ denotes the standard encoder-decoder network of the transformer architecture.

In Falcon, we employ a sequence-to-sequence framework that is capable of putting all distinct tasks in a uniformed format.
As depicted in Fig.\;\ref{fig:model_arc}, given a remote sensing image $\mathcal{I}$ and a text prompt $\mathcal{T}$, we feed $\mathcal{I}$ to image encoder $\mathcal{G}$ to extract the visual token embedding $\mathcal{V} \in\mathbb{R}^{N_v \times D_v}$, with $N_v$ and $D_v$ respectively represent the number and dimension of vision tokens.
At the same time, we leverage $\mathcal{E}$ to process $\mathcal{T}$ in order to obtain the text token embedding $\mathcal{E(T)} \in \mathbb{R}^{N_t\times D}$.
Next, we combine the vision token embedding and the text token embedding to form a multi-modality embedding $\mathcal{X} = [\mathcal{V}', \mathcal{E(T)}]$, with $\mathcal{V}' \in \mathbb{R}^{N_t\times D}$ is derived from $\mathcal{V}$ through a visual adapter \cite{xiao2024florence}, serving as the task-agnostic input to $\mathcal{F}$.

Unlike the previous studies \cite{xiao2024florence,kuckreja2024geochat}, we propose a dynamic prompt training strategy to eliminate the reliance on task-specific tokens.
Specially, given a prompt $\mathcal{T}$, Falcon will dynamically sample several differently phrased versions $\{\mathcal{T}^\prime_i\}_{i=1}^M$ from a predefined prompt pool to form the $\mathcal{X} = \{[\mathcal{V}', \mathcal{E(T}^\prime_i)]\}_{i=1}^{M}$ to join the training process. 
Note that $\{\mathcal{T}^\prime_i\}_{i=1}^M$ and $\mathcal{T}$ share similar semantic meanings.
This design further enhances Falcon's understanding ability of natural language.

To ensure the input and output of distinct tasks in a unified format, we treat each task as a sequence-to-sequence translation task.
As shown in Fig.\;\ref{fig:dataset_vis}, we regard images, prompts, annotations as special languages.
For example, an instruction of unified format for the region caption is as follows: "Describe the \texttt{<}$region$\texttt{>} in the image.", where \texttt{<}$region$\texttt{>} is 
\texttt{<}$box$\texttt{>} \texttt{<}$x1$\texttt{>} \texttt{<}$y1$\texttt{>} \texttt{<}$x2$\texttt{>} \texttt{<}$y2$\texttt{>} \texttt{<}$/box$\texttt{>}  representing location tokens. 
The location tokens are the coordinates of the bounding box. 
We add location tokens to the tokenizer’s vocabulary list, representing quantized coordinates. We create 1000 bins which represent regions using formats tailored to task requirements.

\noindent \textbf{Loss function.} We utilize the cross-entropy loss to optimize the Falcon for 14 tasks like normal large language models.
\begin{equation}
    \mathcal{L}=-\sum_{i=1}^{|y|}\sum_{x\in \mathcal{X}}logP_\theta(y_i|y_{<i},x),
\end{equation}
where $x\in \mathcal{X}$ is the input vector consisting of the image embedding output by the image encoder and the prompt embedding; $y$ is the prediction target; $|y|$ is the number of target tokens, $\theta$ is the Falcon's parameter.


\section{Dataset}
To equip Falcon with powerful image, region, and pixel-level understanding and reasoning capabilities, we introduce Falcon\_SFT, the first large-scale, multi-task remote sensing instruction-tuning dataset. 
It contains 78 million high-quality samples covering 5.6 million multi-resolution, multi-view remote sensing images. This section details its creation process, including data collection, preprocessing, and instruction generation.

\subsection{Data Collection and Preprocessing}
Currently, no existing dataset can fully meet the training requirements of Falcon.
To address this, we devised a simple and straightforward approach, \textit{i.e.} curating and combining various open-source datasets in remote sensing filed.

We collected 90 annotated task-specific RGB image datasets, such as Million-AID \cite{millionAID}, RSICD \cite{RSICD}, and DOTA \cite{dota1, dota2}, encompassing nearly all publicly available datasets originating from satellites, airplanes, drones, etc. 
After manual screening, we refined the selection to 67 relevant datasets. 
The complete list is available in Sec.\;\ref{supp:sec_A} of the supplementary material. 
Notably, we provide download links and metadata (image size, spatial resolution, and quantity) to help reduce data collection efforts for researchers.

Next, we integrate the 67 collected remote sensing datasets, by establishing a unified and consistent annotation format.
This standardization is necessary because different datasets use varying annotation formats (e.g., polygons vs. mask images), which can complicate data integration.
Besides, to broaden application scenarios, we repurpose existing data structures to generate additional annotations, expanding the number of supported tasks to 14. 
These tasks are categorized into three levels, namely, \textit{Image-level}: Image Classification, Image VQA, Counting, Image Captioning, and Image Detailed Captioning; \textit{Region-level}: Region Classification-HBB, Region Classification-OBB, Region Detection-HBB, Region Detection-OBB, Visual Grounding, and Region Captioning; \textit{Pixel-level}: Pixel Classification, Pixel Segmentation, and Change Detection.
This categorization aligns with prior discussions in \cite{OMGLLaVA, wu2024visionllm2}.
For more detailed data collection and preprocessing procedures, please see Sec.\;\ref{supp:sec_A} of the supplementary material.

\begin{table*}[t]
\centering
\small
\resizebox{2.1\columnwidth}{!}{
\begin{tabular}{ccccccccccccccccccccccccccccccc}
\hline
 &  & \multicolumn{29}{c}{Accuracy} \\
\cmidrule(lr){3-31}
Models &\#params & \rotatebox{85}{BHP Watertanks} & \rotatebox{85}{CLRS} & \rotatebox{85}{DIOR} & \rotatebox{85}{DOTA2.0} & \rotatebox{85}{FAIR1M1.0} & \rotatebox{85}{GEONRW} & \rotatebox{85}{Globe230k} & \rotatebox{85}{Hefei} & \rotatebox{85}{Hurricane\_Damage} & \rotatebox{85}{LoveDA} & \rotatebox{85}{MultiScene} & \rotatebox{85}{NWPU\_RESISC45} & \rotatebox{85}{NaSC\_TG2} & \rotatebox{85}{OPTIMAL31} & \rotatebox{85}{AiRound} & \rotatebox{85}{PatternNet} & \rotatebox{85}{RSD46\_WHU} & \rotatebox{85}{RSITMD} & \rotatebox{85}{RSI\_CB} & \rotatebox{85}{RSOD} & \rotatebox{85}{RSSCN7} & \rotatebox{85}{RS\_C11} & \rotatebox{85}{SIRI\_WHU} & \rotatebox{85}{SODA-A} & \rotatebox{85}{UCAS-AOD} & \rotatebox{85}{WHU\_GID} & \rotatebox{85}{iSAID} & \rotatebox{85}{million-AID} & \rotatebox{85}{xView} \\ \hline
MiniCPM-V\cite{hu2024minicpm} &3B  & 0.06 & 0.14 & 0.08 & 0.10 & 0.12 & 0.03 & 0.07 & 0.13 & 0.14 & 0.06 & 0.03 & 0.15 & 0.12 & 0.17 & 0.11 & 0.14 & 0.10 & 0.14 & 0.09 & 0.25 & 0.16 & 0.15 & 0.10 & 0.08 & 0.36 & 0.09 & 0.12 & 0.11 & 0.02 \\
MiniGPT-v2\cite{zhu2023minigpt} &7B  & 0.03 & 0.26 & 0.16 & 0.11 & 0.06 & 0.03 & 0.05 & 0.11 & 0.12 & 0.06 & 0.06 & 0.31 & 0.08 & 0.36 & 0.43 & 0.34 & 0.16 & 0.30 & 0.19 & 0.03 & 0.27 & 0.18 & 0.12 & 0.05 & 0.02 & 0.32 & 0.11 & 0.25 & 0.12 \\
LLaVA-1.5\cite{llava} &7B  & 0.28 & 0.44 & 0.35 & 0.19 & 0.33 & 0.17 & 0.31 & 0.29 & 0.41 & 0.15 & 0.15 & 0.46 & 0.31 & 0.56 & 0.54 & 0.47 & 0.36 & 0.47 & 0.31 & 0.44 & 0.50 & 0.48 & 0.36 & 0.34 & 0.52 & 0.59 & 0.24 & 0.30 & 0.24 \\
Qwen-VL-Chat\cite{2023Qwen-VL} &7B  & 0.11 & 0.33 & 0.17 & 0.10 & 0.12 & 0.09 & 0.23 & 0.18 & 0.20 & 0.05 & 0.07 & 0.32 & 0.17 & 0.40 & 0.41 & 0.34 & 0.20 & 0.30 & 0.15 & 0.27 & 0.39 & 0.41 & 0.12 & 0.11 & 0.42 & 0.53 & 0.13 & 0.27 & 0.09 \\
Sphinx\cite{2023SPHINX} &7B  & 0.21 & 0.22 & 0.18 & 0.13 & 0.34 & 0.08 & 0.18 & 0.23 & 0.33 & 0.11 & 0.09 & 0.24 & 0.24 & 0.29 & 0.27 & 0.26 & 0.13 & 0.21 & 0.14 & 0.81 & 0.26 & 0.33 & 0.24 & 0.21 & 0.69 & 0.37 & 0.19 & 0.08 & 0.06 \\
RemoteCLIP \cite{liu2024remoteclip} &304M  & 0.40 & 0.64 & 0.59 & 0.46 & 0.54 & 0.36 & 0.37 & 0.37 & 0.40 & \textbf{0.60} & 0.30 & 0.68 & 0.57 & 0.78 & 0.60 & 0.60 & 0.44 & \textbf{0.81} & 0.46 & 0.98 & 0.67 & 0.69 & 0.57 & 0.56 & 0.99 & 0.87 & 0.59 & 0.44 & 0.38 \\
GeoChat\cite{kuckreja2024geochat} &7B  & 0.56 & 0.46 & 0.65 & 0.60 & 0.70 & 0.12 & 0.20 & 0.21 & 0.38 & 0.11 & 0.10 & 0.58 & 0.36 & 0.62 & 0.59 & 0.48 & 0.29 & 0.46 & 0.29 & 0.94 & 0.36 & 0.51 & 0.32 & 0.60 & 0.91 & 0.53 & 0.54 & 0.38 & 0.29 \\
LHRS-Bot\cite{muhtar2024lhrs} &7B  & 0.28 & 0.58 & 0.35 & 0.20 & 0.33 & 0.16 & 0.25 & 0.21 & 0.17 & 0.16 & 0.14 & 0.73 & 0.49 & 0.87 & 0.56 & 0.59 & 0.38 & 0.74 & 0.39 & 0.54 & 0.55 & 0.72 & 0.44 & 0.28 & 0.55 & 0.76 & 0.23 & 0.37 & 0.23 \\
\textbf{Falcon(Ours)} &0.7B  & \textbf{0.98} & \textbf{0.91} & \textbf{0.87} & \textbf{0.95} & \textbf{0.98} & \textbf{0.90} & \textbf{0.71} & \textbf{0.79} & \textbf{0.99} & 0.56 & \textbf{0.57} & \textbf{0.94} & \textbf{0.99} & \textbf{0.97} & \textbf{0.85} & \textbf{0.99} & \textbf{0.56} & 0.50 & \textbf{0.99} & \textbf{0.99} & \textbf{0.94} & \textbf{0.92} & \textbf{0.96} & \textbf{0.88} & \textbf{1.00} & \textbf{0.95} & \textbf{0.89} & \textbf{0.93} & \textbf{0.85}
 \\ \hline
\end{tabular}
}
\vspace{-3mm}
\caption{A comparison of image classification performance on several datasets with 8 generic and remote sensing VLMs.}
\label{tab:image_classification_performance}
\end{table*}


\begin{table*}[t]
\centering
\small
\resizebox{2.1\columnwidth}{!}{
\begin{tabular}{ccccccccccccccccc}
\hline
 &  & \multicolumn{15}{c}{Accuracy} \\
\cmidrule(lr){3-17}
Models & \#params & \rotatebox{85}{ASD} & \rotatebox{85}{DIOR} & \rotatebox{85}{DOTA2.0} & \rotatebox{85}{FAIR1M1.0} & \rotatebox{85}{RSVQA HR} & \rotatebox{85}{HRSC2016} & \rotatebox{85}{RSOD} & \rotatebox{85}{S2-SHIPS} & \rotatebox{85}{SODA-A} & \rotatebox{85}{ShipRS} & \rotatebox{85}{UCAS-AOD} & \rotatebox{85}{VHRShips} & \rotatebox{85}{airplane\_det} & \rotatebox{85}{ship\_det} & \rotatebox{85}{xView} \\ \hline
MiniCPM-V\cite{hu2024minicpm} & 3B & 0.662 & 0.426 & 0.260 & 0.295 & 0.281 & 0.599 & 0.302 & 0.031 & 0.161  & 0.555 & 0.321 & 0.777 & 0.360 & 0.707 & 0.132 \\
MiniGPT-v2\cite{zhu2023minigpt} & 7B & 0.637 & 0.429 & 0.248 & 0.336 & 0.275 & 0.595 & 0.295 & 0.063 & 0.148  & 0.527 & 0.293 & 0.768 & 0.306 & 0.659 & 0.152 \\
LLaVA-1.5\cite{llava} & 7B & 0.681 & 0.249 & 0.221 & 0.268 & 0.000 & 0.478 & 0.326 & 0.125 & 0.175 & 0.371 & 0.400 & 0.683 & 0.453 & 0.561 & 0.101 \\
Sphinx\cite{2023SPHINX} &7B  & 0.480 & 0.430 & 0.257 & 0.365 & 0.181 & 0.669 & 0.403 & 0.031 & 0.151 & 0.516 & 0.388 & 0.741 & 0.384 & 0.195 & 0.146 \\
GeoChat\cite{kuckreja2024geochat} & 7B & 0.738 & 0.453 & 0.240 & 0.377 & 0.240 & 0.588 & 0.302 & \textbf{0.156} & 0.165 & 0.545 & 0.308 & 0.774 & 0.466 & 0.683 & 0.171 \\
LHRS-Bot\cite{muhtar2024lhrs} & 7B & 0.678 & 0.455 & 0.244 & 0.357 & 0.681 & 0.714 & 0.319 & 0.125 & 0.171 & 0.611 & 0.362 & 0.773 & 0.371 & 0.732 & 0.170 \\
\textbf{Falcon(Ours)} & 0.7B & \textbf{0.952} & \textbf{0.770} & \textbf{0.619} & \textbf{0.816} & \textbf{0.718} & \textbf{0.838} & \textbf{0.821} & \textbf{0.156} & \textbf{0.391} & \textbf{0.782} & \textbf{0.857} & \textbf{0.916} & \textbf{0.808} & \textbf{0.854} & \textbf{0.278} \\ \hline
\end{tabular}
}
\vspace{-2mm}
\caption{A comparison of object number counting performance on several datasets with both generic and remote sensing VLMs.}
\vspace{-4mm}
\label{tab:counting_performance}
\end{table*}

\subsection{Unified Instruction Generation}
Next, we transform our integrated dataset into a multi-task instruction-tuning dataset for vision-language model training.
We take the steps as follows.

\noindent \textbf{Define Instruction Templates}.
To facilitate the understanding and execution of specific tasks by VLMs, we design standardized instruction templates based on different remote sensing tasks.
For examples, for the Object Detection Task, ``Detect \texttt{<}$class$\texttt{>} in the image.
Use Rotated bounding boxes." is given.
The rotated bounding box is represented as 
\texttt{<}$quad$\texttt{>} \texttt{<}$x_1$\texttt{>} \texttt{<}$y_1$\texttt{>} \texttt{<}$x_2$\texttt{>} \texttt{<}$y_2$\texttt{>} 
\texttt{<}$x_3$\texttt{>} \texttt{<}$y_3$\texttt{>}
\texttt{<}$x_4$\texttt{>} \texttt{<}$y_4$\texttt{>}
\texttt{<}$/quad$\texttt{>}, specifying the coordinates of the four vertices, each expressed in thousandths.
Please see Fig.\;\ref{fig:dataset_vis} for instruction examples of all 14 tasks.
\noindent \textbf{Generate Image Instruction Pairs}.
To create image instruction pairs based on the defined templates, we first iterate over the dataset and generate specific instruction for each image based on its task type (\emph{e.g.}, detection, segmentation).
We then combine the generated instruction with corresponding image and annotations into a structured pair. 
This enables the model to learn diverse task responses using different instruction-based prompts.

\noindent \textbf{Generate the Multi-instruction Pool}.
To enhance language understanding and reduce reliance on task-specific tokens, we diversify instruction patterns for each task using an LLM \cite{bai2023qwentechnicalreport}. It generates multiple variations of the same instruction with different complexity levels. For instance, ``Describe the image." is expanded into ``Describe the contents of this image.", ``Analyze the image and explain its visual content.", and ``Can you identify what this image shows?". This approach enriches textual diversity in training data, helping VLMs to improve performance across various tasks. 
Please see Sec. \ref{supp:sec_B} of the supplementary material for multi-instruction examples.



\subsection{Falcon\_SFT Dataset}
Following the above data processing steps, we finally constructed the large-scale remote sensing instruction-tuning dataset Falcon\_SFT.
We compare Falcon\_SFT with various datasets used for remote sensing vision-language models in Tab.\;\ref{tab:datasets_comparisons}.
The Falcon\_SFT dataset features the largest number of samples (78 million) and images (5.6 million), supporting the highest number of tasks (14).
It is also more comprehensive by covering image, region, and pixel-level spatial hierarchies.
For detailed statistics of Falcon\_SFT dataset, please see Tab.\;\ref{tab:datasets_analysis} in Sec.\;\ref{supp:sec_A} of the supplementary material.


\label{Dataset}
\section{Experiments}

\begin{table}[t]
\centering
\small
{
\linespread{1.0}
\setlength\tabcolsep{10pt}
\scriptsize
{
\begin{tabular}{cccc}
\hline
\multirow{2}{*}{Models} & \multirow{2}{*}{\#params} & \multicolumn{2}{c}{RSVQA HR(Accuracy)} \\
\cmidrule(lr){3-4}
 &  & \multicolumn{1}{c}{Compare} & \multicolumn{1}{c}{presence} \\ \hline
MiniCPM-V\cite{hu2024minicpm} & 3B & 0.734 & 0.646 \\
MiniGPT-v2\cite{zhu2023minigpt} & 7B & 0.647 & 0.668 \\
Qwen-VL-Chat\cite{2023Qwen-VL} & 7B & 0.668 & 0.643 \\
Florence-2-L\cite{xiao2024florence} & 0.7B & 0.396 & 0.650 \\
Sphinx\cite{2023SPHINX} & 7B & 0.556 & 0.514 \\
GeoChat\cite{kuckreja2024geochat} & 7B & 0.778 & 0.688 \\
LHRS-Bot\cite{muhtar2024lhrs} & 7B & 0.922 & 0.928 \\
\textbf{Falcon(Ours)} & 0.7B & \textbf{0.927} & \textbf{0.931} \\\hline
\end{tabular}
}
}
\vspace{-2mm}
\caption{A comparison of VQA performance on several datasets with 7 generic and remote sensing VLMs.}
\vspace{-2mm}
\label{tab:VQA_performance}
\end{table}

\begin{table}[t]
\small
\centering
\resizebox{1\columnwidth}{!}{
\begin{tabular}{@{}lcccccccccccc@{}}
\toprule
\multirow{2}{*}{Rank}     & \multicolumn{4}{c}{Detail} & \multicolumn{4}{c}{Position} & \multicolumn{4}{c}{Hallucination} \\ \cmidrule(lr){2-5} \cmidrule(lr){6-9} \cmidrule(lr){10-13}
         & GeoChat & Qwen & Sphinx & Falcon & GeoChat & Qwen & Sphinx & Falcon & GeoChat & Qwen & Sphinx & Falcon \\ \midrule
A=4      & 7       & 45   & 92     & 196    & 15      & 38   & 75     & 148    & 47      & 290  & 161    & 280    \\
B=3      & 101     & 170  & 240    & 250    & 106     & 110  & 163    & 207    & 95      & 137  & 197    & 159    \\
C=2      & 295     & 218  & 154    & 49     & 328     & 188  & 221    & 136    & 170     & 53   & 121    & 55     \\
D=1      & 97      & 67   & 14     & 5      & 51      & 164  & 41     & 9      & 188     & 20   & 21     & 6      \\
Average  & 2.036   & 2.386 & 2.82   & \textbf{3.274}  & 2.17    & 2.044 & 2.544  & \textbf{2.988}  & 2.002   & 3.39 & 2.996  & \textbf{3.426}  \\ \bottomrule
\end{tabular}
}
\vspace{-2mm}
\caption{A comparison of human evaluations for image captioning. Each value in 3-6 rows represents the number of captions marked as A/B/C/D by 10 volunteers. We calcuated the average score in the last row, by quantifying the A-D ratings as 4 to 1 points.}
\vspace{-4mm}
\label{tab:ranked_comparison}
\end{table}


In this section, we present the experimental setup and results to evaluate Falcon's performance, including: 1) both qualitative and quantitative performance evaluations on all $14$ complex remote sensing tasks; 
2) zero-shot performance of Falcon compared with previous methods. 
The results demonstrate Falcon's ability to handle complex vision-language tasks and highlight its strengths in image, region, and pixel-level understanding and reasoning.
To point out, due to the page limit, we provide additional experimental results in the supplementary material, including qualitative performance evaluations of all 14 tasks in Sec.\;\ref{supp:sec_E}, quantitative performance evaluations for for tasks not covered in the main paper in Sec.\;\ref{supp:sec_F}, qualitative performance evaluations on diversified instructions in Sec.\;\ref{supp:sec_G}, human evaluations on image captioning performance in Sec.\;\ref{supp:sec_H}, more ablation studies in Sec.\;\ref{supp:sec_I} and the details of evaluation metrics for each task in Sec.\;\ref{supp:sec_J}.

\noindent \textbf{Implementation Details}.
Falcon consists of an image encoder and a transformer-based encoder-decoder, with a total of 0.7B parameters. 
The detailed architecture is illustrated in Fig.\;\ref{fig:model_arc}. 
We initialized the model's parameters using the pre-trained weights provided by \cite{xiao2024florence}. Unlike \cite{xiao2024florence}, we increased the output token length to 4096 in order to obtain more detailed representations. The training batch size for Falcon was 640, the learning rate was set to $1e^{-5}$, and the image size is 448 × 448. We trained the model for 4 days using 160 Nvidia A100 GPUs.

\begin{figure}[t]
  \centering
   \includegraphics[width=1\linewidth]{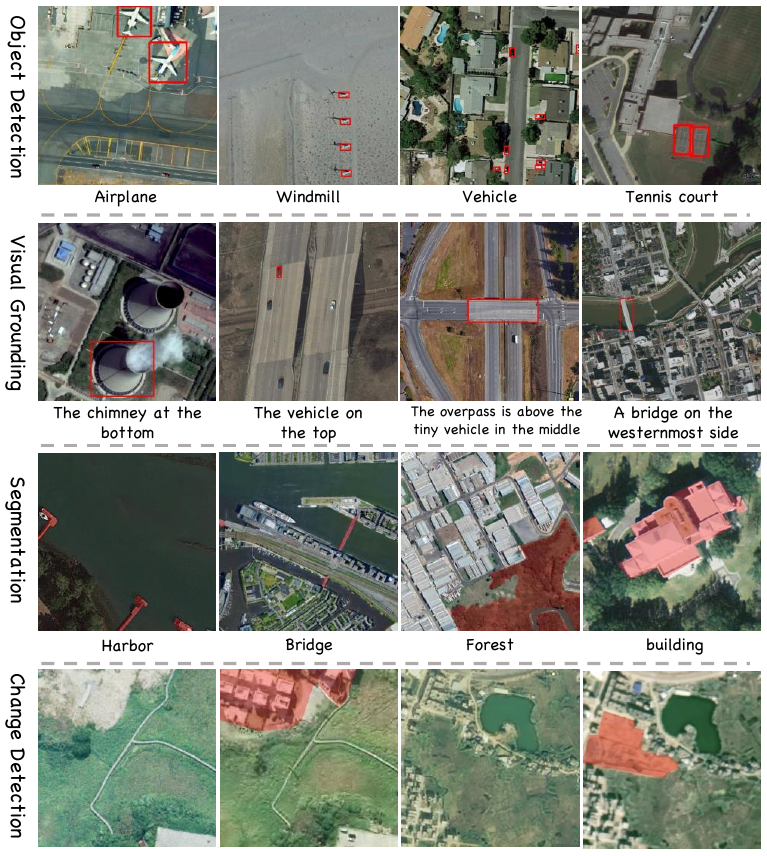}
    \vspace{-6mm}
   \caption{Visualization of Falcon's output on tasks of object detection, visual grounding, segmentation, and change detection.}
   \vspace{-4mm}
   \label{fig:vis_four_task}
\end{figure}

\begin{table*}[t]
\centering
\small
\resizebox{2.1\columnwidth}{!}{
\begin{tabular}{cccccccccccccc}
\hline
\multirow{2}{*}{Models} & \multirow{2}{*}{\#params} & \multicolumn{12}{c}{AP@IoU=0.5(\%)} \\
\cmidrule(lr){3-14}
 &  & BHP Watertanks & DIOR & DOTA2.0 & GEONRW & Globe230k & HRSC2016 & LoveDA & RSOD & UCAS-AOD & VHRShips & iSAID & xView \\ \hline
MiniGPTv2\cite{zhu2023minigpt} & 7B & 7.228 & 9.430 & 1.624 & 2.085 & 14.456 & 51.156 & 15.769 & 24.564 & 34.944 & 42.896 & 2.535 & 0.722 \\
Florence-2-L\cite{xiao2024florence} & 0.7B & 5.810 & 26.975 & 12.245 & 0.410 & 8.971 & 67.159 & 8.484 & 62.750 & 78.792 & 66.883 & 16.673 & 3.259 \\
Qwen-VL-Chat\cite{2023Qwen-VL} & 7B & 9.795 & 15.807 & 2.970 & 2.672 & 12.106 & 58.547 & 18.989 & 38.906 & 53.929 & 53.393 & 4.292 & 1.561 \\
Sphinx\cite{2023SPHINX} &7B & 0.068 & 0.469 & 0.054 & 0.572 & 5.780 & 3.866 & 4.481 & 0.292 & 0.021 & 0.537 & 0.111 & 0.014 \\
\textbf{Falcon(Ours)} & 0.7B & \textbf{81.896} & \textbf{56.652} & \textbf{27.043} & \textbf{30.410} & \textbf{30.458} & \textbf{93.750} & \textbf{47.214} & \textbf{85.249} & \textbf{93.838} & \textbf{89.543} & \textbf{33.846} & \textbf{27.165} \\ \hline
\end{tabular}
}
\vspace{-3mm}
\caption{A comparison with generic and remote sensing VLMs on object detection with horizontal bounding box.}
\vspace{-3mm}
\label{tab:object_detection_performance}
\end{table*}

\begin{table*}[t]
\centering
\small
\resizebox{2.1\columnwidth}{!}{
\begin{tabular}{cccccccccccc}
\hline
\multirow{4}{*}{Models} & \multirow{4}{*}{\#params} & \multicolumn{4}{c}{Image   level}                       & \multicolumn{3}{c}{Region   level}          & \multicolumn{3}{c}{Pixel   level}       \\
\cmidrule(lr){3-6} \cmidrule(lr){7-9} \cmidrule(lr){10-12}
                        &                           & Cap          & D.Cap        & \multicolumn{2}{c}{Count(\%)} & \multicolumn{2}{c}{Det$^{hbb}$(\%)} & Cls$^{hbb}(\%)$   & Seg        & \multicolumn{2}{c}{CD}    \\
                        &                           & (CIDEr)      & (CIDEr)      & \multicolumn{2}{c}{(Acc)} & \multicolumn{2}{c}{(AP@IoU=0.5)}    & (Acc)       & (mIoU)     & \multicolumn{2}{c}{(mIoU)} \\
\cmidrule(lr){3-3} \cmidrule(lr){4-4} \cmidrule(lr){5-6} \cmidrule(lr){7-8} \cmidrule(lr){9-9} \cmidrule(lr){10-10} \cmidrule(lr){11-12}
                        &                           & UCM-Captions & UCM-Captions & MAR20    & NWPU-VHR-10    & MAR20       & NWPU-VHR-10     & NWPU-VHR-10 & GID15      & CCD          & WHU-CD      \\ \hline
MiniCPM-V \cite{hu2024minicpm}                 & 3B                        & 0.000            & 0.000            & 39.1    & 48.1          & -           & -               & -           & -          & -            & -           \\
MiniGPT-v2 \cite{zhu2023minigpt}               & 7B                        & 16.282       & 0.166        & 35.4    & 49.9          & 53.078           & 22.828               & 76.5       & -          & -            & -           \\
LLaVA-1.5 \cite{llava}        & 7B                        & 0.004        & 0.010         & 47.5        & 34.4          & -           & -               & -           & -          & -            & -           \\
Qwen-VL-Chat \cite{2023Qwen-VL}              & 7B                        & 12.992       & 1.912        & -    & -          & 79.286      & 32.099          & -           & -          & -            & -           \\
Florence-2-L \cite{xiao2024florence}             & 0.7B                      & 13.844       & 1.568        & -        & -              & 88.843      & 38.905          & 52.9       & -          & -            & -           \\
Sphinx \cite{2023SPHINX}                 & 7B                        & 0.000            & 0.056        & 45.8    & 50.1          & 0.161       & 0.185           & 35.9       & -          & -            & -           \\
GeoChat \cite{kuckreja2024geochat}                & 7B                        & 0.288        & 0.092        & 42.3    & 43.6          & -           & -               & -           & -          & -            & -           \\
LHRS-Bot \cite{muhtar2024lhrs}               & 7B                        & 8.365        & 20.180        & 40.6    & 48.7          & -           & -               & -           & -          & -            & -           \\
\textbf{Falcon(Ours)}            & 0.7B                      & \textbf{30.481}       & \textbf{23.553}       & \textbf{87.6}    & \textbf{81.7}          & \textbf{94.025}      & \textbf{81.847}          & \textbf{98.8}       & \textbf{0.389}      & \textbf{0.427}        & \textbf{0.531}      \\ \hline
\end{tabular}
}
\vspace{-3mm}
\caption{A comparison of zero-shot performance on various tasks with 8 generic and remote sensing VLMs.}
\vspace{-3mm}
\label{tab:zeroshot_performance}
\end{table*}

\begin{table*}[!ht]
\centering
\small
\resizebox{2.0\columnwidth}{!}{
\begin{tabular}{cccccccccccc}
\hline
\multirow{3}{*}{Data scale} & \multicolumn{3}{c}{Task granularity} & \multirow{3}{*}{\#params} & \multicolumn{3}{c}{Image level} & \multicolumn{2}{c}{Region level}   & \multicolumn{2}{c}{Pixel level}       \\ \cmidrule(lr){2-4} \cmidrule(lr){6-8} \cmidrule(lr){9-11} \cmidrule(lr){11-12}
 &  Image  & Region & Pixel &             & Cap          & D.Cap        & Count(\%) & Det$^{hbb}$(\%) & Cls$^{hbb}(\%)$   & Seg        & CD    \\
 &  level  & level  & level   &             & (CIDEr)      & (CIDEr)      & (Acc) & (AP@IoU=0.5)    & (Acc)       & (mIoU)     & (mIoU) \\ \hline
10\%  & \cmark & \cmark & \cmark & 0.7B & 74.3  & 15.1  & 58.2 &  38.4 &  98.4 &  0.488 & 0.387 \\
50\%  & \cmark & \cmark & \cmark & 0.7B & 94.7  & 26.3  & 60.7 &  42.6 &  93.7 &  0.524 & 0.514 \\ 
- & \cmark & \xmark & \xmark & 0.7B & 96.7  & 25.9  & 61.6 &  0.0  &  75.0 &  0.042 & 0.000 \\ 
- & \cmark & \cmark & \xmark & 0.7B & 97.2  & 24.8  & 63.2 &  36.0 &  99.3 &  0.042 & 0.000 \\ 
100\% & \cmark & \cmark & \cmark & 0.3B & 107.6 & 25.6  & 64.4 &  42.8 &  97.7 &  0.529 & 0.542 \\ 
100\% & \cmark & \cmark & \cmark & 0.7B & 111.4 & 27.9  & 65.2 &  43.3 &  99.2 &  0.544 & 0.536 \\ \hline
\end{tabular}
}
\vspace{-2mm}
\caption{Ablation studies on the effects of data scale, task granularity, and model size for Falcon.}
\vspace{-4mm}
\label{tab:ablation_studies}
\end{table*}

\subsection{Performance Evaluation across 14 tasks}
\label{sec: 5.1}
\noindent \textbf{Image-level Tasks}.
In this section, we presented the performance of Falcon over image classification tasks (\textit{c.f.} Tab.\;\ref{tab:image_classification_performance}), counting tasks (\textit{c.f.} Tab.\;\ref{tab:counting_performance}) and VQA tasks (\textit{c.f.} Tab.\;\ref{tab:VQA_performance}).
As shown in Tab.\;\ref{tab:image_classification_performance}, generic VLMs, such as MiniGPTv2 \cite{zhu2023minigpt} and Qwen\_chat \cite{2023Qwen-VL} encountered obstacles in performing effectively on remote sensing data, since they usually lacked the expert knowledge of this domain.
Meanwhile, compared with VLMs specialized in remote sensing \cite{liu2024remoteclip,kuckreja2024geochat,muhtar2024lhrs}, Falcon achieved better performance in all related datasets, with only 0.7B parameters.
Besides, we also provided detailed performance comparison of counting targets in Tab.\;\ref{tab:counting_performance}.
Such a task requires compositional perception and reasoning capabilities, presenting significant challenges to state-of-the-art VLMs.
To this end, Falcon achieved superior performance in targets counting, showcasing its sophisticated capabilities.
Finally, we compared Falcon with previous VLMs in VQA tasks, which these models usually excelled in.
As shown in Tab.\;\ref{tab:VQA_performance}, Falcon still surpassed previous VLMs with less model parameters, indicating its strong instruction following capabilities.

For image captioning tasks, we conduct human evaluations for Falcon's responses.
Specifically, captions were evaluated across three dimensions: detail, position, and hallucination, using a four-level rating system (\emph{i.e.}, A, B, C, D quantified as 4 to 1 points, where a higher point represents a better caption). The results in Tab. \ref{tab:ranked_comparison} showed that Falcon achieved the highest average scores across all three dimensions, compared with other VLMs. 
Please see Sec.\;\ref{supp:sec_H} of the supplementary material for detailed experimental setup.

\noindent \textbf{Region-level Tasks}.
Beyond image-level tasks, our Falcon also support fine-grained region-level tasks.
To this end, we present the performance of Falcon on object detection (horizontal bounding box) in Tab.\;\ref{tab:object_detection_performance}.
It is noticeable that previous VLMs demonstrated limited performance in this task, exposing their limitations in localization capabilities.
In contrast, Falcon outperformed previous methods, highlighting its ability to handle complex remote sensing tasks. 

\noindent \textbf{Pixel-level Tasks}.
Besides, we also present the evaluation results of Falcon on pixel-level tasks. 
To the best of our knowledge, Falcon is the first VLM capable of showing satisfactory performance on pixel-level tasks, such as segmentation and change detection.
The qualitative results of Falcon are shown in Fig.\;\ref{fig:vis_four_task}.
Falcon successfully segmented designated complex targets in images based on prompts and also identified changes between two similar images.


\subsection{Zero-shot Evaluation}
Finally, we evaluate the capabilities of Falcon in terms of zero-shot evaluations.
We present the detailed performance comparison in Tab.\;\ref{tab:zeroshot_performance}, where these evaluation datasets were not used during training. 
Compared with previous VLMs, Falcon achieved performance improvements over all three levels of tasks.
For image-level tasks, Falcon established a new record on many datasets, such as UCM-Captions and MAR20 for image captioning and image counting.
For region-level tasks and pixel-level tasks, Falcon demonstrated exceptional performance on many datasets, which required comprehensive localization and reasoning capabilities.
In contrast, such capabilities were commonly missing or even not supported in prior VLMs.

\subsection{Ablation experiments}
This section presents the ablation studies to analyze the effects of data scale, task granularity, and model size on performance, as summarized in Tab.\;\ref{tab:ablation_studies}. The results demonstrate a consistent performance improvement as the training data scale increases — for instance, from 10\% training samples to 50\% training samples and ultimately to 100\% training samples. Furthermore, as the task granularity becomes more refined, the model not only handles more complex tasks effectively but also enhances performance on simpler ones. A comparison between the 0.3B and 0.7B parameter models reveals that a larger parameter count leads to better generalization performance. 
More ablation studies can be found in Sec.\;\ref{supp:sec_I} of the supplementary material.



\section{Conclusion}
This paper develops Falcon, a holistic vision-language foundation model tailored for remote sensing with comprehensive perception and reasoning capabilities. 
To facilitate the training of Falcon, we further create Falcon\_SFT dataset which consists of approximately 78M high-quality data samples, covering 5.6M remote sensing images.
Various qualitative and quantitative experiments have demonstrated that Falcon showcased remarkable zero-shot and in-dataset performance across 14 remote sensing vision-language tasks and more than 100 test datasets.
We will release the complete dataset, code, and model weights, hoping to help further advance this research field.



{
    \small
    \bibliographystyle{ieeenat_fullname}
    \bibliography{main}
}

\clearpage
\setcounter{page}{1}
\renewcommand\thealgorithm{\Roman{algorithm}}
\renewcommand\thetable{\Roman{table}}
\renewcommand\thefigure{\Roman{figure}}
\renewcommand\theequation{S\arabic{equation}}
\setcounter{algorithm}{0}
\setcounter{table}{0}
\setcounter{figure}{0}
\setcounter{equation}{0}

\onecolumn

{
   \newpage
        \centering
        \Large
        \textbf{\thetitle}\\
        \vspace{0.5em}Supplementary Material \\
        \vspace{1.0em}
   }

In the supplementary material, we will introduce the following content to complement the details of our study.

\addtocontents{toc}{\protect\setcounter{tocdepth}{3}} 
\tableofcontents

\clearpage 
\renewcommand{\thesection}{\Alph{section}}
\setcounter{section}{0}
\section{67 collected remote sensing datasets for construction of Falcon\_SFT}
\label{supp:sec_A}

Here we provide more details for the 67 collected remote sensing datasets which are used for the construction of our Falcon\_SFT datasets. It should be noted that only images with annotations are considered in the construction of Falcon\_SFT. 
Moreover, we provide the direct download link of each dataset in Table \ref{tab:all_datasets_list} for easy accessibility.

\begin{table}[htbp]
\small
\centering
\resizebox{1\columnwidth}{!}{
\begin{tabular}{cccccccc}
\hline
\multirow{2}{*}{Dataset}     & \multirow{2}{*}{Download Link}    & Dataset Size  & Number of Images & \multirow{2}{*}{Image Size}     & \multirow{2}{*}{Image Modalities}  & \multirow{2}{*}{Number of Class} & \multirow{2}{*}{Supported Tasks}      \\
    &           & (Unzipped) & Used in Our Experiments &            &   &  &  \\
\hline
AID\cite{AID}                                & \href{https://captain-whu.github.io/AID/}{Link}                                                                                                                  & 2.5GB        & 10000                                      & 600$\times$600               & RGB                                                                        & 30              & Image Classification                                         \\
ASD \cite{Airbus_Ship_Detection_Challenge} & \href{https://www.kaggle.com/c/airbus-ship-detection}{Link}                                                                                                      & 30GB         & 42556                                      & 768$\times$768               & RGB                                                                           & 1               & Object Detection                                             \\
AiRound\cite{AiRound}                   & \href{http://patreo.dcc.ufmg.br/2020/07/22/multi-view-datasets/}{Link}                                                                                           & 5.9GB       & 1165                                       & 500$\times$500               & RGB, multi-spectral                                                           & 11              & Image Classification                                         \\
airplane\_det\cite{airplane_detection}                & \href{https://aistudio.baidu.com/datasetdetail/131179}{Link}                                                                                                      & 8.8GB        & 430                                        & 4096$\times$4096             & RGB                                                                           & 1               & Object Detection (Airplane)                                  \\
BHP Watertanks\cite{BH_pool}     & \href{http://patreo.dcc.ufmg.br/2020/07/29/bh-pools-watertanks-datasets/}{Link}                                                                                  & 2.0GB        & 325                                        & 3840$\times$2160             & RGB                                                                           & 2               & Semantic Segmentation                                        \\
CDD\cite{CDD}                                & \href{https://drive.google.com/file/d/1GX656JqqOyBi\_Ef0w65kDGVto-nHrNs9/edit}{Link}                                                                             & 4.6GB        & 338                                        & 1900$\times$1000             & RGB                                                                           & -               & Change Detection                                             \\
CLRS\cite{2020CLRS}                               & \href{https://huggingface.co/datasets/jonathan-roberts1/CLRS}{Link}                                                                                              & 2.8GB        & 15000                                      & 256$\times$256               & RGB                                                                           & 25              & Image Classification                                         \\
DIOR\cite{DIOR}                               & \href{https://gcheng-nwpu.github.io/\#Datasets}{Link}                                                                                                            & 11.7GB         & 23463                                      & 800$\times$800               & RGB                                                                           & 20              & Object Detection                                             \\
DIOR-RSVG\cite{RSVG}                          & \href{https://drive.google.com/drive/folders/1hTqtYsC6B-m4ED2ewx5oKuYZV13EoJp\_}{Link}                                                                           & 10.5GB         & 17402                                      & 800$\times$800               & RGB                                                                           & 20              & Visual Grounding                                             \\
DLRSD\cite{DLRSD}                              & \href{https://sites.google.com/view/zhouwx/dataset\#h.p\_hQS2jYeaFpV0}{Link}                                                                                     & 844MB        & 2100                                       & 256$\times$256               & RGB                                                                           & 17              & Semantic Segmentation, Scene Classification, Image Retrieval \\
DOTA2.0\cite{dota2}                            & \href{https://captain-whu.github.io/DOTA/dataset.html}{Link}                                                                                                     & 23GB         & 2423                                       & 800$\times$800-20000 $\times$ 20000 & RGB                                                                           & 18              & Object Detection                                             \\
DSIFN\cite{DSIFN}                             & \href{https://github.com/GeoZcx/A-deeply-supervised-image-fusion-network-for-change-detection-in-remote-sensing-images/tree/master/dataset}{Link}                & 531MB        & 7604                                       & 512$\times$512               & RGB                                                                           & -               & Change Detection                                             \\
EGY\_BCD\cite{EGY-BCD}                           & \href{https://github.com/oshholail/EGY-BCD}{Link}                                                                                                                & 1.1GB        & 4366                                       & 256$\times$256               & RGB                                                                           & -               & Change Detection                                             \\
EuroSAT\cite{2017EuroSAT}                            & \href{https://zenodo.org/records/7711810}{Link}                                                                                                                  & 3.0GB        & 27000                                      & 64$\times$64                 & RGB, multi-spectral                                                           & 10              & Image Classification                                         \\
FAIR1M1.0\cite{2022FAIR1M}                          & \href{https://gaofen-challenge.com/benchmark}{Link}                                                                                                              & 74GB         & 16488                                      & 500$\times$500 -1200$\times$5000    & RGB                                                                           & 36              & Object Detection                                             \\
GEONRW\cite{GEONRW}                             & \href{https://ieee-dataport.org/open-access/geonrw}{Link}                                                                                                        & 30.15GB      & 7549                                       & 1000$\times$1000             & RGB, SAR                                                                      & 11              & Semantic Segmentation                                        \\
GID15\cite{GID-15}                              & \href{https://captain-whu.github.io/GID15/}{Link}                                                                                                                & 18GB         & 110                                        & 7200$\times$6800             & multi-spectral                                                                & 15              & Semantic Segmentation                                        \\
Globe230k\cite{shi2023globe230k}                          & \href{https://zenodo.org/records/8429200}{Link}                                                                                                                  & 12GB         & 232819                                     & 512$\times$512               & RGB                                                                           & 10              & Semantic Segmentation                                        \\
Hefei\cite{Heifei}                              & \href{https://aistudio.baidu.com/datasetdetail/88597}{Link}                                                                                                      & 11MB         & 533                                        & 256$\times$256               & RGB                                                                           & 5               & Image Classification                                         \\
RSVQA HR\cite{LRBEN_HRBEN}                              & \href{https://zenodo.org/records/6344367}{Link}                                                                                                                  & 14GB         & 9505                                       & 512$\times$512               & RGB                                                                           & 89              & VQA                                                          \\
HRSC2016\cite{HRSC2016}                  & \href{https://www.kaggle.com/datasets/guofeng/hrsc2016}{Link}                                                                                                    & 7.0GB        & 1070                                       & 1000$\times$600              & RGB                                                                           & 1               & Object Detection                                             \\
HRSCD\cite{HRSCD}                              & \href{https://ieee-dataport.org/open-access/hrscd-high-resolution-semantic-change-detection-dataset}{Link}                                                       & 12GB         & 554                                        & 10000$\times$10000           & RGB                                                                           & 2/6             & Change Detection, Land Cover Classification                            \\
Hurricane\_Damage\cite{HurricaneDamage}                  & \href{https://huggingface.co/datasets/jonathan-roberts1/Satellite-Images-of-Hurricane-Damage}{Link}                                                              & 71MB         & 23000                                      & 128$\times$128               & RGB                                                                           & 2               & Image Classification                                         \\
Inria\cite{maggiori2017can}                              & \href{https://project.inria.fr/aerialimagelabeling/}{Link}                                                                                                       & 26GB         & 180                                        & 5000$\times$5000             & RGB                                                                           & 2               & Semantic Segmentation                                        \\
iSAID\cite{waqas2019isaid}                              & \href{https://captain-whu.github.io/BED4RS/\#}{Link}                                                                                                             & 20.9GB         & 1869                                       & 500$\times$500 -1200$\times$5000    & RGB                                                                           & 15              & Instance Segmentation                                        \\
LEVIR-CD\cite{LEVIR-CD}                           & \href{http://chenhao.in/LEVIR/}{Link}                                                                                                                            & 2.4GB        & 1098                                       & 1024$\times$1024             & RGB                                                                           & -               & Change Detection                                             \\
LEVIR-CD+\cite{LEVIR-CD+}                          & \href{https://github.com/S2Looking/Dataset}{Link}                                                                                                                & 7.1GB        & 1766                                       & 1024$\times$1024             & RGB                                                                           & -               & Change Detection                                             \\
LoveDA\cite{wang2021loveda}                             & \href{https://github.com/Junjue-Wang/LoveDA}{Link}                                                                                                               & 9.0GB        & 4191                                       & 1024$\times$1024             & RGB                                                                           & 8               & Semantic Segmentation                                        \\
RSVQA LR\cite{LRBEN_HRBEN}                              & \href{https://zenodo.org/records/6344367}{Link}                                                                                                                  & 199MB        & 772                                        & 256$\times$256               & RGB                                                                           & 89              & VQA                                                          \\
MAR20\cite{Yu2023MAR20AB}                              & \href{https://gcheng-nwpu.github.io/\#Datasets}{Link}                                                                                                            & 2.3GB        & 3842                                       & 800$\times$800               & RGB                                                                           & 20              & Object Detection                                             \\
million-AID\cite{millionAID}                        & \href{https://captain-whu.github.io/DiRS/}{Link}                                                                                                                 & 136G         & 10000                                      & 256 $\times$ 256, 512 $\times$ 512  & RGB                                                                           & 51              & Image Classification                                         \\
MSBC\cite{MSBC_MSOSCD}                               & \href{https://github.com/Lihy256/MSCDUnet}{Link}                                                                                                                 & 3.2GB        & 7402                                       & 256$\times$256               & multi-spectral                                                                & 2               & Change Detection                                             \\
MSOSCD\cite{MSBC_MSOSCD}                             & \href{https://github.com/Lihy256/MSCDUnet}{Link}                                                                                                                 & 2.4GB        & 10144                                      & 256$\times$256               & multi-spectral                                                                & 2               & Change Detection                                             \\
MultiScene\cite{hua2021multiscene}                         & \href{https://multiscene.github.io/}{Link}                                                                                                                       & 7.6GB        & 100000                                     & 512$\times$512               & RGB                                                                           & 36              & Multilable Image Classification                              \\
NaSC\_TG2\cite{Zhou2021NaSCTG2NS}                          & \href{https://captain-whu.github.io/BED4RS/\#}{Link}                                                                                                             & 8.8GB        & 20000                                      & 128$\times$128               & multi-spectral                                                                & 10              & Image Classification                                         \\
NJDS\cite{NJDS}                               & \href{https://drive.google.com/file/d/1cQRWORIgW-X2BaeRo1hvFj7vlQtwnmne/view?userstoinvite=infinitemabel.wq@gmail.com\&ts=636c5f76\&actionButton=1\&pli=1}{Link} & 1.2GB        & 2                                          & 14231$\times$11381           & RGB                                                                           & -               & Change Detection, Semantic Segmentation                      \\
NWPU\_RESISC45\cite{NWPU-RESISC45}                     & \href{https://gcheng-nwpu.github.io/\#Datasets}{Link}                                                                                                            & 414MB        & 31500                                      & 256$\times$256               & RGB                                                                           & 45              & Image Classification                                         \\
NWPU-VHR-10\cite{NWPU_VHR-10}                        & \href{https://gcheng-nwpu.github.io/\#Datasets}{Link}                                                                                                            & 74MB         & 650                                        & 900$\times$500               & RGB, Infrared                                                                 & 10              & Image Classification, Object Detection                       \\
OPTIMAL31\cite{OPTIMAL31}                          & \href{https://huggingface.co/datasets/jonathan-roberts1/Optimal-31}{Link}                                                                                        & 25MB         & 1860                                       & 256$\times$256               & RGB                                                                           & 31              & Image Classification                                         \\
PatternNet\cite{Zhou2017PatternNetAB}                         & \href{https://sites.google.com/view/zhouwx/dataset\#h.p\_hQS2jYeaFpV0}{Link}                                                                                     & 1.4GB        & 30400                                      & 256$\times$256               & RGB                                                                           & 38              & Image Classification                                         \\
RS\_C11\cite{RS_C11}                            & \href{https://huggingface.co/datasets/jonathan-roberts1/RS\_C11}{Link}                                                                                           & 925MB        & 1232                                       & 512$\times$512               & RGB                                                                           & 11              & Image Classification                                         \\
RSD46\_WHU\cite{RSD46-WHU}                         & \href{https://huggingface.co/datasets/jonathan-roberts1/RSD46-WHU}{Link}                                                                                         & 9.7GB        & 100131                                     & 256$\times$256               & RGB                                                                           & 4               & Image Classification                                         \\
RSI\_CB\cite{RSI-CB}                            & \href{https://github.com/lehaifeng/RSI-CB}{Link}                                                                                                                 & 6.3GB        & 61454                                      & 256 $\times$ 256, 128 $\times$ 128  & multi-spectral                                                                & 35              & Image Classification                                         \\
RSICap\cite{hu2023rsgpt}                             & \href{https://github.com/Lavender105/RSGPT}{Link}                                                                                                                & 1.3GB        & 2685                                       & 512$\times$512               & RGB                                                                           & 16              & Image Caption                                                \\
RSICD\cite{RSICD}                              & \href{https://github.com/201528014227051/RSICD\_optimal}{Link}                                                                                                   & 1.1GB        & 10734                                      & 224$\times$224               & RGB                                                                           & 30              & Image Caption                                                \\
RSITMD\cite{RSITMD}                             & \href{https://github.com/xiaoyuan1996/AMFMN/tree/master/RSITMD}{Link}                                                                                            & 911MB        & 4743                                       & 256$\times$256               & RGB                                                                           & 32              & Image Caption, Image Classification                          \\
RSOD\cite{RSOD}                               & \href{https://github.com/RSIA-LIESMARS-WHU/RSOD-Dataset-}{Link}                                                                                                  & 318MB        & 936                                        & 227$\times$227               & RGB                                                                           & 4               & Object Detection                                             \\
RSSCN7\cite{RSSCN7}                             & \href{https://github.com/palewithout/RSSCN7}{Link}                                                                                                               & 352MB        & 2800                                       & 400$\times$400               & RGB                                                                           & 7               & Image Classification                                         \\
RSVG\cite{sun2022}                               & \href{https://sunyuxi.github.io/publication/GeoVG}{Link}                                                                                                         & 1.5GB        & 4239                                       & 1024$\times$1024             & RGB                                                                           & 10              & Object Retrieval                                             \\
S2Looking\cite{S2Looking}                          & \href{https://github.com/S2Looking/Dataset}{Link}                                                                                                                & 14GB         & 1970                                       & 1024$\times$1024             & multi-spectral                                                                & -               & Change Detection                                             \\
S2-SHIPS\cite{S2-SHIPS}                           & \href{https://drive.google.com/file/d/1zDgz6wr5kxikPR7o9nJ2IjMcaqwtiLLu/view}{Link}                                                                              & 5.2GB        & 16                                         & 1783$\times$938              & multi-spectral                                                                & 1               & Object Detection (Ship)                                      \\
SAMRS\cite{wang2024samrs}                               & \href{https://github.com/ViTAE-Transformer/SAMRS}{Link}                                                                                                          & 99.2GB        & 93352                                      & 600$\times$600-1024$\times$1024     & RGB                                                                           & 18, 20, 37      & Semantic Segmentation                                        \\
ship\_det\cite{rsaicp2021}                     & \href{https://aistudio.baidu.com/datasetdetail/134218}{Link}                                                                                                      & 2.6GB        & 25                                         & 20000$\times$20000           & SAR                                                                           & 1               & Object Detection (Ship)                                      \\
ShipRSImagerNet\cite{zhang2021shiprsimagenet}                    & \href{https://drive.google.com/file/d/1wApkaSoa9mXRfXQiq6lTtlVrv4cSc6vv/view}{Link}                                                                              & 8.4GB        & 2748                                       & 930$\times$930               & RGB                                                                           & 50              & Object Detection (Ship)                                      \\
SIRI\_WHU\cite{SIRI-WHU}                          & \href{https://huggingface.co/datasets/jonathan-roberts1/SIRI-WHU}{Link}                                                                                          & 1.1GB        & 2400                                       & 200$\times$200               & high resolution                                                               & 12              & Image Classification                                         \\
SODA-A\cite{SODA-A}                             & \href{https://shaunyuan22.github.io/SODA/}{Link}                                                                                                                 & 12GB         & 2513                                       & 4800 $\times$ 2700           & RGB                                                                           & 9               & Object Detection                                             \\
Sydney\_Captions\cite{UCMCaptions_SydneyCaptions}                   & \href{https://github.com/isaaccorley/torchrs\#sydney-captions}{Link}                                                                                             & 441MB        & 530                                        & 500$\times$500               & RGB                                                                           & -               & Image Caption                                                \\
SYSU-CD\cite{SYSU-CD}                            & \href{https://github.com/liumency/DSAMNet}{Link}                                                                                                                 & 11GB         & 40000                                      & 256$\times$256               & RGB                                                                           & 3               & Change Detection                                             \\
SZTAKI\cite{SZTAKI2009}                             & \href{http://web.eee.sztaki.hu/remotesensing/building\_benchmark.html}{Link}                                                                                     & 4.6MB        & 9                                          & 930$\times$930               & RGB                                                                           & 2               & Object Detection                                             \\
UCAS-AOD\cite{UCAS-AOD}                           & \href{https://github.com/ming71/UCAS-AOD-benchmark}{Link}                                                                                                        & 3.3GB        & 1510                                       & 1280$\times$659              & RGB                                                                           & 3               & Object Detection                                             \\
UCM-Captions\cite{UCMCaptions_SydneyCaptions}                       & \href{https://github.com/isaaccorley/torchrs\#uc-merced-ucm-captions}{Link}                                                                                      & 819MB        & 2100                                       & 256$\times$256               & RGB                                                                           & 21              & Image Caption                                                \\
UCM-Classification\cite{UCM-classification}                 & \href{https://vision.ucmerced.edu/datasets/}{Link}                                                                                                               & 819MB        & 2100                                       & 256$\times$256               & RGB                                                                           & 21              & Image Classification                                         \\
VHRShips\cite{VHRShip}                           & \href{https://github.com/radres333/VHRShips}{Link}                                                                                                               & 3.4GB        & 5275                                       & 2272$\times$1270,1280$\times$720    & RGB                                                                           & 34              & Object Detection                                             \\
WHU\_GID\cite{GID}                           & \href{https://captain-whu.github.io/GID15/}{Link}                                                                                                                & 6.1G         & 30000                                      & 112$\times$112, 56$\times$56        & multi-spectral                                                                & 15              & Semantic Segmentation                                        \\
WHU-CD\cite{WHU-CD}                              & \href{http://gpcv.whu.edu.cn/data/building\_dataset.html}{Link}                                                                                                  & 2.1GB        & 3900                                       & 512$\times$512               & RGB                                                                           & -               & Change Detection                                             \\
WHU\_RS19\cite{WHU-RS19}                          & \href{https://captain-whu.github.io/BED4RS/\#}{Link}                                                                                                             & 114M         & 1005                                       & 600$\times$600               & RGB                                                                           & 19              & Image Classification                                         \\
xView\cite{xView}                              & \href{https://challenge.xviewdataset.org/download-links}{Link}                                                                                                   & 24G          & 846                                        & 2500$\times$2500-3200$\times$5000   & RGB                                                                           & 60              & Object Detection   \\
\hline
\end{tabular}
}
\caption{67 collected remote sensing datasets for the construction of Falcon\_SFT.}

\label{tab:all_datasets_list}
\end{table}

\noindent \textbf{Integrating Remote Sensing Datasets.}
After finishing dataset collections, we further integrate the 67 collected remote sensing datasets, a process that presents several challenges, including:

\noindent $\bullet$ \textit{Inconsistent Annotation Formats}.
There are differences in annotation formats across various datasets. For example, different datasets may follow distinct annotation standards or conventions for segmentation masks, such as polygons or mask images, making data integration and unified processing more complex.
To address this, we propose to establish a unified annotation standard and convert all datasets to this format. 
Automated scripts are developed to transform different data formats into the specified standard, reducing the complexity of data integration and ensuring consistent processing.
Please see Section \ref{supp:sec_B} of the supplementary material for examples of the unified annotation format in our proposed Falcon\_SFT dataset.

\noindent  $\bullet$ \textit{Inconsistent Category Naming}.
There are inconsistencies in category naming, namely, the same target objects are labeled with different category names across different datasets. 
For example, ``car" is labeled as ``car" in some datasets \cite{DLRSD} and as ``vehicle" in others \cite{DIOR}, leading to inconsistencies in category labeling.
To solve this issue, we propose to create a unified category naming dictionary to map different labels for the same objects to a standardized category name. 
This can be achieved through a combination of automated mapping rules and manual interventions, ensuring consistent category naming across datasets.
Please see Section \ref{supp:sec_C} of the supplementary material for the unified category naming dictionary.



\noindent \textbf{Data Repurposing and Task Expansion.} 
The collected raw datasets cover 7 tasks, including image classification, object detection, image segmentation, image caption, visual question answering, visual grounding, and change detection.
To further extend the application scenarios of our dataset, we propose to repurpose existing data structure to generate more annotations for additional tasks, enabling support for 14 tasks in total.
Please see Section \ref{supp:sec_D} of the supplementary material for the data conversion of 14 new tasks in our proposed Falcon\_SFT.
Specifically, we divide 14 tasks into image-level, region-level and pixel-level. 
At the image level, tasks involve Image Classification, Image VQA, Counting, Image Captioning and Image Detailed Captioning.
At the region level, tasks involve Region Classification-HBB, Region Classification-OBB, Region Detection-HBB, Region Detection-OBB, Visual Grounding and Region Captioning.
At the pixel level,  tasks involve Pixel Classification, Pixel Segmentation, Change Detection.
This division was also discussed in \cite{OMGLLaVA,wu2024visionllm2}.

\begin{table}[t]
\begin{center}
{
\linespread{1.0}
\setlength\tabcolsep{10pt}
\scriptsize
{
\begin{tabular}{c l c c}
\toprule
Spatial hierarchy & Tasks & \#images & \#Annotations  \\
\midrule
\multirow{5}{*}{\rotatebox[origin=c]{0}{\centering Image Level}}
& Cls & 1.35M & 9.43M \\
& Cap & 24.7K & 343.5K \\
& D.Cap & 15.9K & 89.6K \\
& Count & 316.2K & 5.67M \\
& VQA & 17.2K & 4.36M  \\
\midrule
\multirow{6}{*}{\rotatebox[origin=c]{0}{\centering Region Level}}
& Cls$^{hbb}$ & 121K & 3.15M \\
& Cls$^{obb}$ & 241.5K & 6.26M \\
& R.Cap & 15.3K & 231.2K  \\
& Det$^{hbb}$ & 885.3K & 10.78M  \\
& Det$^{obb}$ & 1.01M & 11.97M \\
& VG & 15.3K & 231.2K\\
\midrule
\multirow{2}{*}{\rotatebox[origin=c]{0}{\centering Pixel Level}}& Cls$^{poly}$ & 764.3K & 15.87M \\
& Seg & 764.3K & 9.31M \\
& CD & 76.5K & 528.8K  \\
\bottomrule
\end{tabular}
}
}
\end{center}
\vspace{-4mm}
\caption{Image and annotation statistics of the Falcon\_SFT dataset.}
\vspace{-4mm}
\label{tab:datasets_analysis}
\end{table}

\clearpage
\section{Unified annotation example and multi-instruction conversation example}
\label{supp:sec_B}

\includegraphics[width=0.95\linewidth]{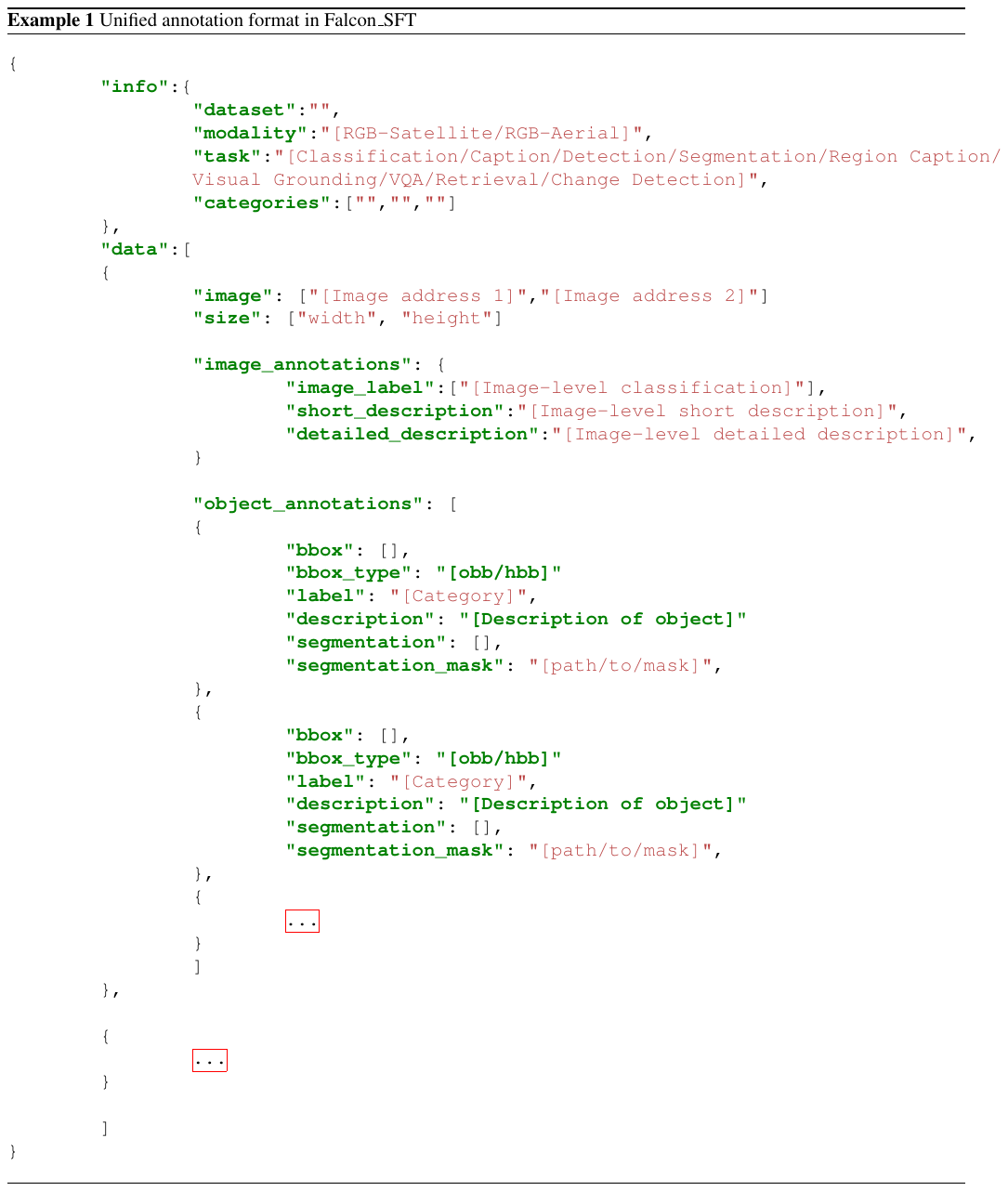}

\includegraphics[width=0.95\linewidth]{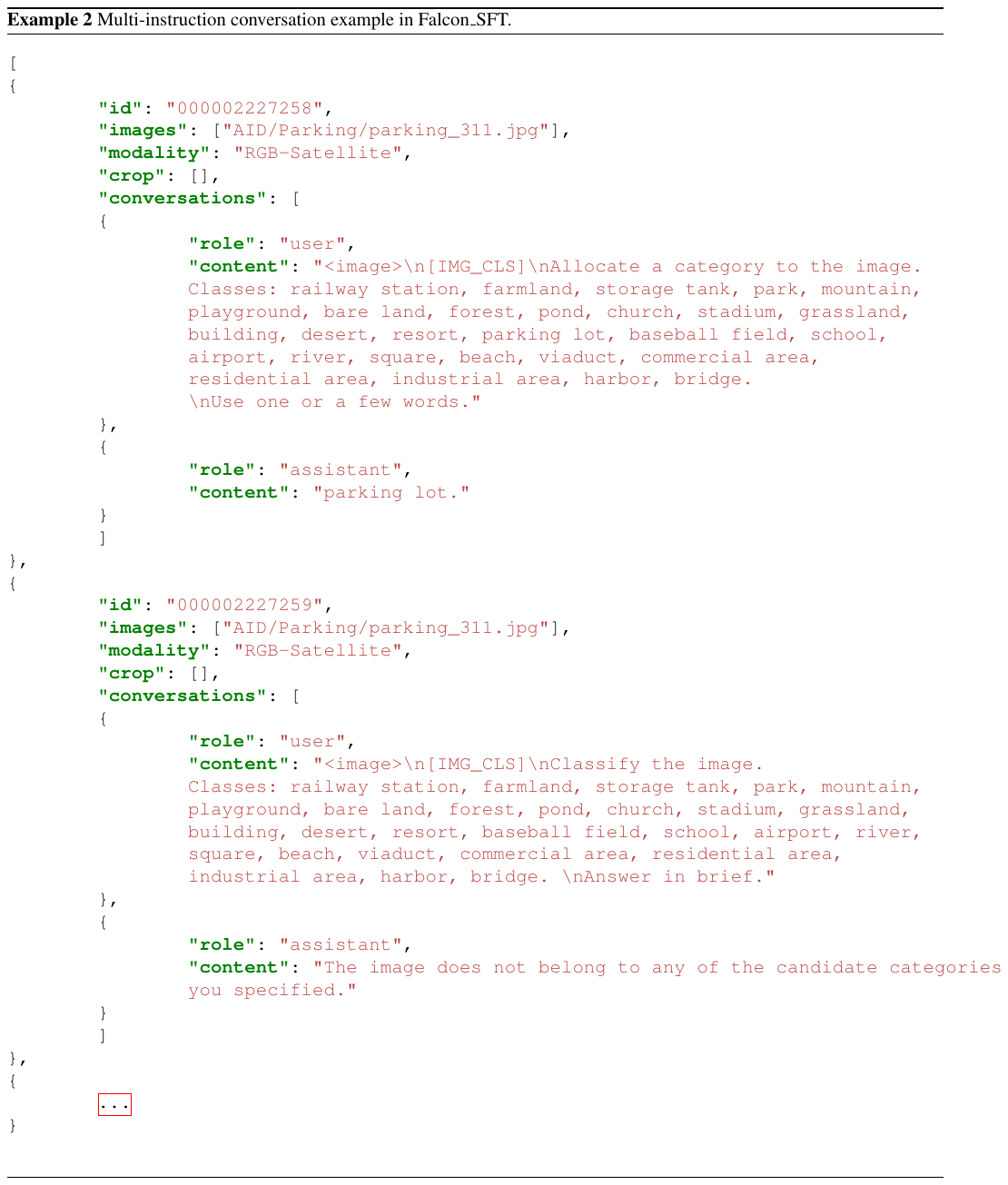}

\clearpage
\section{Mapped category dictionary}
\label{supp:sec_C}
\begin{table}[!h]
\centering
    \begin{tabular}{cc||cc||cc||cc}
    \toprule
    \textbf{Num.} & \textbf{Category} & \textbf{Num.} & \textbf{Category} & \textbf{Num.} & \textbf{Category}& \textbf{Num.} & \textbf{Category}\\
    \midrule
    1 & agriculture area & 2 & aircraft hangar &3 & airplane &4 & airport\\
    5 & airport runway &6 & apron &7 & aquaculture & 8 & avenue \\
    9 & bare land & 10 & baseball field & 11 & basketball court & 12 & beach \\
    13 & bridge & 14 & building &15 & cement mixer& 16 & cemetery \\
    17 & chimney & 18 & church &19 & cloud & 20 & coastline\\
    21 & commercial area & 22 & construction site &23 & container & 24 & crane \\
    25 & crosswalk & 26 & dam & 27 & damaged building &28 & desert \\
    29 & excavator & 30 & expressway service area& 31 & expressway toll station & 32 & factory area\\ 
    33 & farmland & 34 & field & 35 & football field & 36 & footbridge \\
    37 & forest & 38 & fork road & 39 & freeway & 40&garden\\
    41 & golf field & 42 & graff &43 & grassland  & 44 & green house \\
    45 & greenbelt & 46 & ground track field & 47 & harbor & 48 & helicopter\\
    49 & helipad & 50 & highway & 51 & hirst & 52 & ice \\
    53 & ice land & 54 & impervious surface &55 & industrial area& 56 & intersection \\
    57 & irrigated area & 58 & island &59 & lake & 60 & lakeshore\\
    61 & locomotive & 62 & mine &63 & mountain &64 & oil gas field\\
    65 & oil well &66 & orchard &67 & overpass & 68 & palace \\
    69 & park & 70 & parking lot & 71 & pasture & 72 & pavement \\
    73 & pipeline & 74 & playground &75 & pond& 76 & power station \\
    77 & pylon & 78 & quarry &79 & railway & 80 & railway station\\
    81 & refinery & 82 & residential area &83 & resort &84 & river\\
    85 & road &86 & rock land &87 & roundabout & 88 & runway \\
    89 & rural residential area & 90 & school & 91 & sea & 92 & sewage \\
    93 & shed & 94 & ship &95 & shipping yard& 96 & shrub land \\
    97 & snowberg & 98 &soccer ball field &99 & solar panel & 100 & solar power station\\
    101 & square & 102 &stadium &83 & statue &84 & steelsmelter\\
    105 & storage land &106 & storage tank &107 & stream & 108 & substation \\
    109 & swimming pool & 110 & tennis court & 111 & tent & 112 & terrace \\
    113 & terraced field & 114 & thermal power station &115 & tower& 116 & town \\
    117 & train carriage & 118 & transformer station &119 & tree & 120 & tundra\\
    121 & turning circle & 122 & urban residential area &123 & vehicle &124 & viaduct\\
    125 & wastewater plant &126 & water area &127 & wetland & 128 & wind turbine \\
    129& windmill\\
    \bottomrule
    \end{tabular}
    
    \caption{Mapped object category exploited in Falcon\_SFT.}
    \label{tab:Mapped category}
\end{table}

\clearpage
\section{Data conversion of 14 new tasks in Falcon\_SFT}
\label{supp:sec_D}

\begin{table}[htbp]
\centering
\small
\resizebox{1\columnwidth}{!}{
    \begin{tabular}{lll}
    \toprule
    Raw Task & New Task & Explanation \\
    \midrule
    Image Classification & Image Classification & Set the category of the image as the answer. \\
    VQA & VQA & No changes. \\
    Image Caption & Image Caption & Set a description with less than 4 sentences / 35 words as the answer. \\
    Image Caption & Detailed Image Caption & Set a description with more than or equal to 4 sentences / 35 words as the answer. \\
    Visual Grounding & Visual Grounding & Set a description as the question, and the corresponding bounding box as the answer. \\
    Visual Grounding & Region Caption & Set a bounding box as the question, and the corresponding description as the answer. \\
    Object Detection-OBB\&HBB & Image Classification & Set the category of all objects contained in the image as the answer. \\
    Object Detection-OBB\&HBB & Counting Target & Set a category as the question and the total number of the corresponding boxes as the answer. \\
    Object Detection-OBB & Region Classification-OBB & Set a bounding box as the question, and set the corresponding category as the answer. \\
    Object Detection-OBB & Region Detection-OBB & Set a category as the question and set all the corresponding bounding boxes as the answer. \\
    Object Detection-HBB & Region Classification-HBB & Set a bounding box as the question, and set the corresponding category as the answer. \\
    Object Detection-HBB & Region Detection-HBB & Set a category as the question, and set all the corresponding bounding boxes as the answer. \\
    Semantic Segmentation & Image Classification & Set the category of all objects contained in the image as the answer. \\
    Semantic Segmentation & Pixel Classification & Set a polygon as the question, and set the corresponding category as the answer. \\
    Semantic Segmentation & Semantic Segmentation & Set a category as the question, and set all the corresponding polygons as the answer \\
    Semantic Segmentation & Region Detection-OBB & Replace the polygons with its horizontal enclosing rectangles as the answer. \\
    Semantic Segmentation & Region Detection-HBB & Replace the polygons with its minimum enclosing rectangles as the answer \\
    Change Detection & Change Detection & Set the changing polygon area as the answer. \\
    \bottomrule
    \end{tabular}
    }
    \caption{We convert 7 raw tasks to 14 new tasks for the construction of Falcon\_SFT. 7 raw tasks are: Image Classification, VQA, Image Caption, Visual Grounding, Object Detection, Semantic Segmentation and Change Detection. 14 new tasks are: Image Classification, VQA, Image Caption, Detailed Image Caption, Visual Grounding, Region Caption, Counting Target, Region Classification-OBB, Region Detection-OBB, Region Classification-HBB, Region Detection-HBB, Pixel Classification, Semantic Segmentation and Change Detection.}
    \label{tab:task7to14}
\end{table}

\clearpage
\section{Qualitative comparisons of 14 tasks with state-of-the-art models}
\label{supp:sec_E}
In this section, we visualized the prediction results for each task and conducted a qualitative comparison between Falcon and other advanced remote sensing VLMs. 
The results highlight the strength and efficiency of Falcon.
\begin{figure*}[ht]
  \centering
   \includegraphics[width=0.9\linewidth]{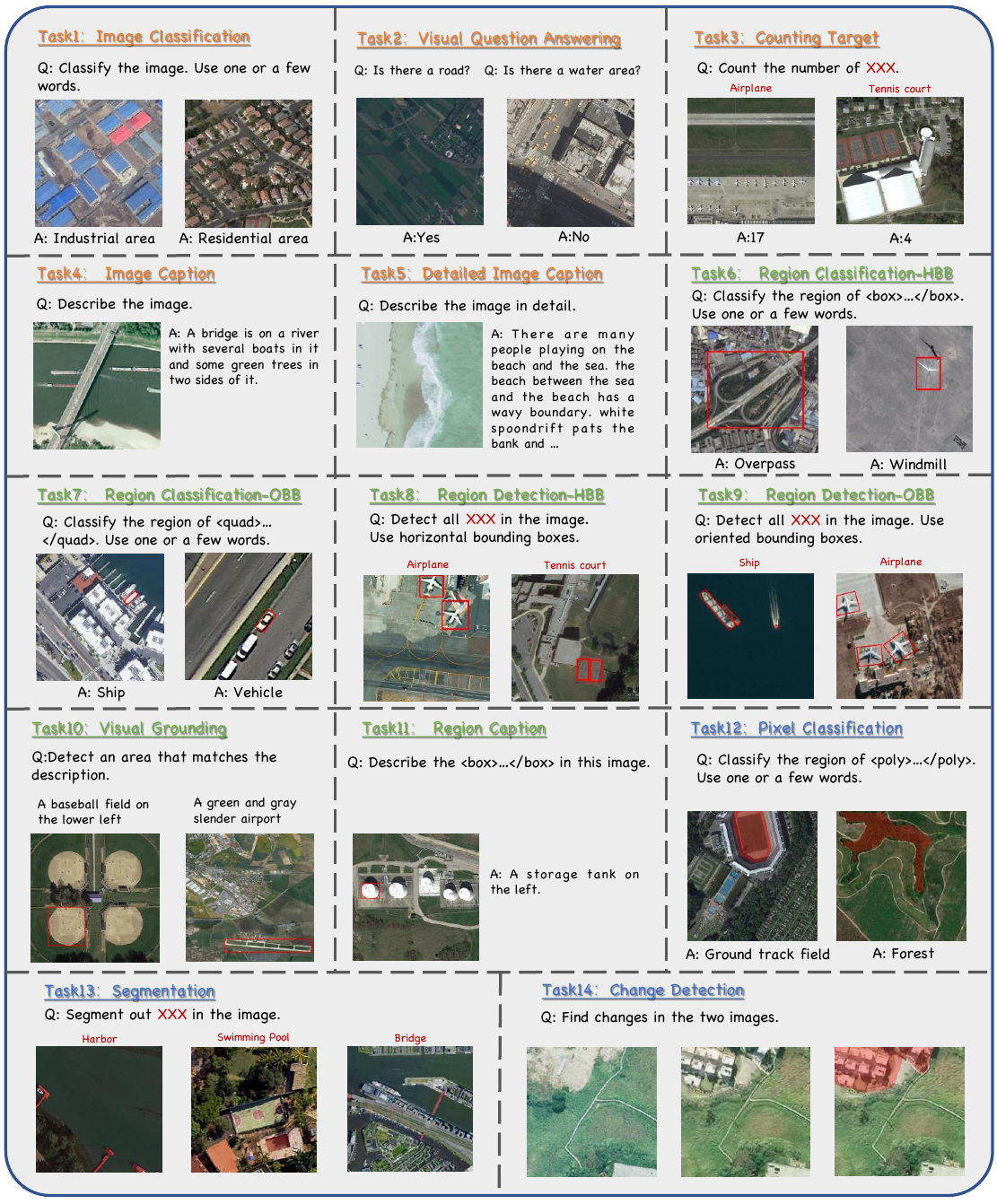}
   \caption{Overview on the qualitative results of Falcon in 14 tasks.}
   \label{fig:visualization of 14 tasks}
\end{figure*}

\clearpage
\subsection{Task1: Image Classification}
\begin{figure*}[ht]
  \centering
   \includegraphics[width=0.9\linewidth]{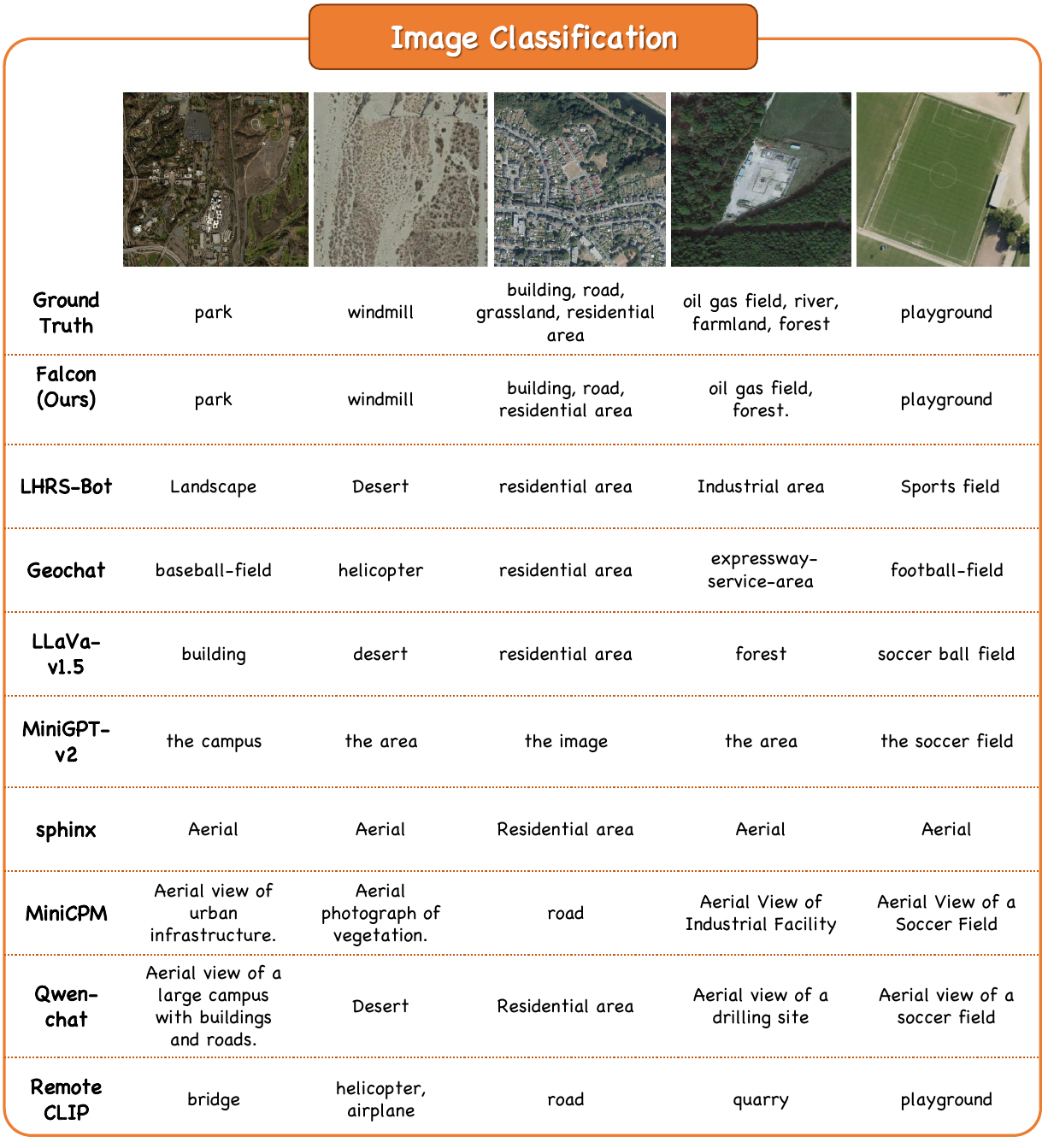}
   \caption{Qualitative comparisons in the task of image classification.}
   \label{task1_fig}
\end{figure*}

\clearpage
\subsection{Task2: Visual Question Answering}
\begin{figure*}[ht]
  \centering
   \includegraphics[width=0.8\linewidth]{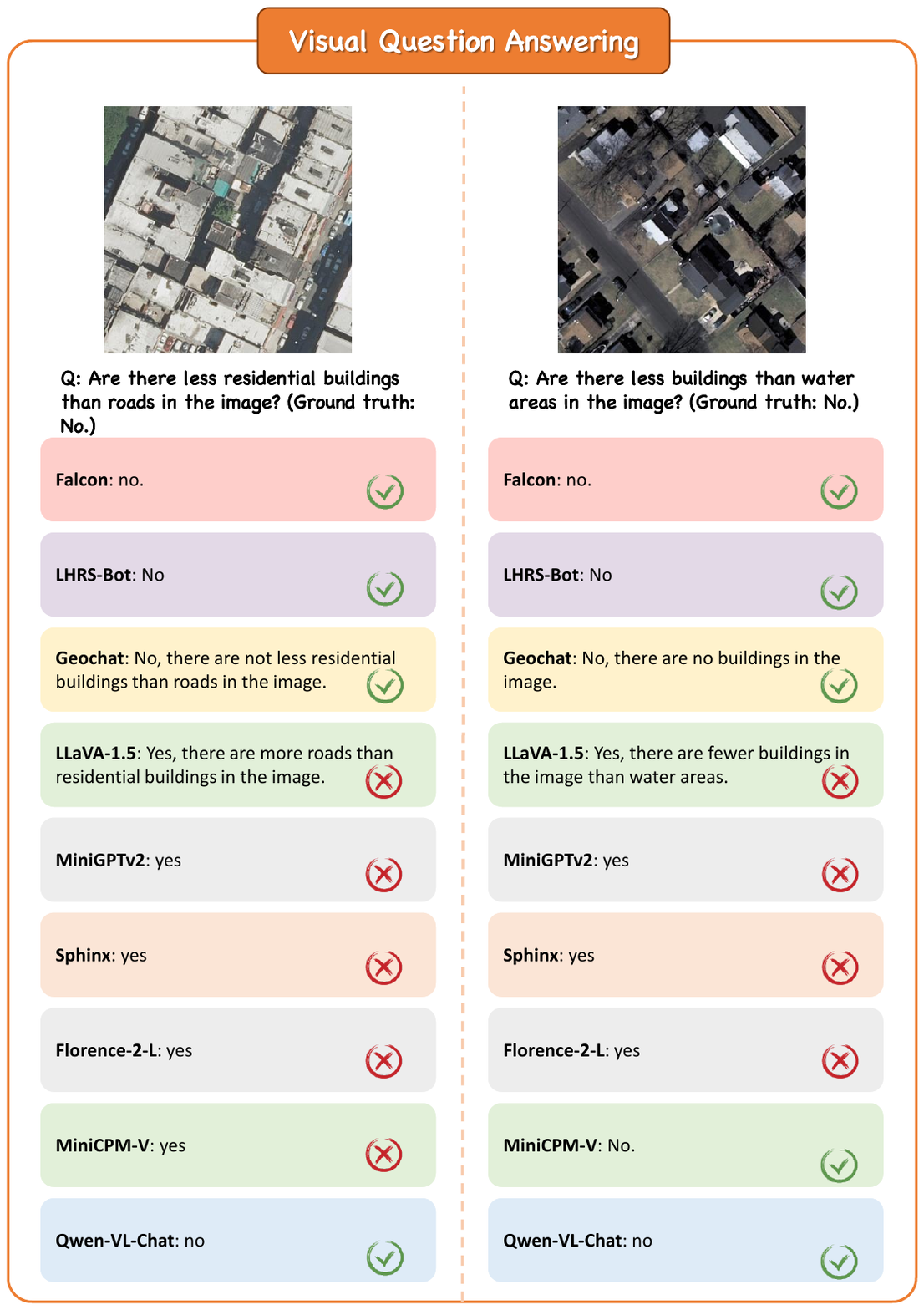}
   \caption{Qualitative comparisons in the task of VQA.}
   \label{task2_fig}
\end{figure*}

\begin{figure*}[ht]
  \centering
   \includegraphics[width=0.8\linewidth]{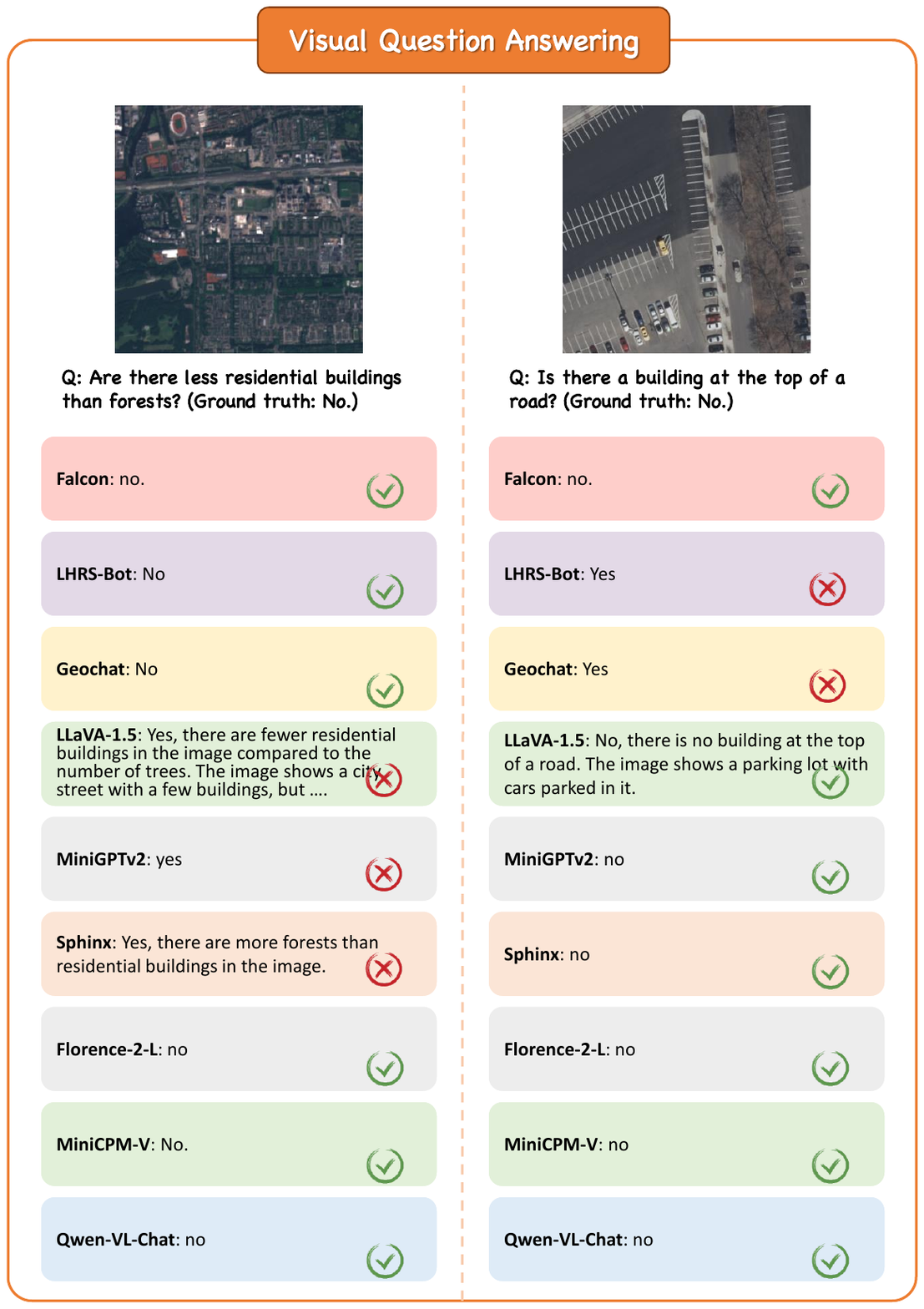}
   \caption{Qualitative comparisons in the task of VQA.}
   \label{task2_fig_2}
\end{figure*}


\clearpage
\subsection{Task3: Counting Target}

\begin{figure*}[ht]
  \centering
   \includegraphics[width=0.8\linewidth]{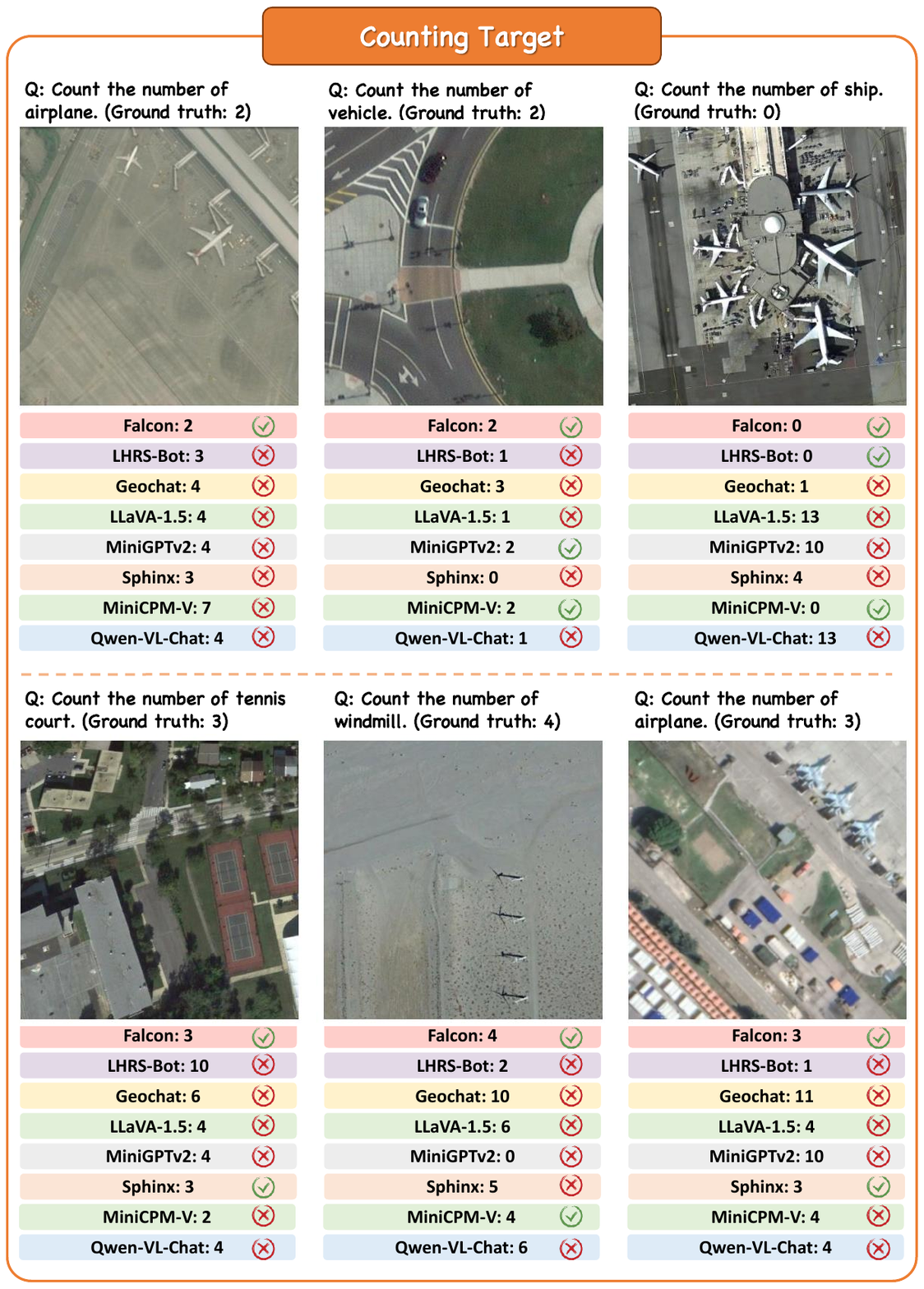}
   \caption{Qualitative comparisons in the task of counting.}
   \label{task3_fig}
\end{figure*}

\clearpage
\subsection{Task4: Image Captioning}

\begin{figure*}[ht]
  \centering
   \includegraphics[width=0.8\linewidth]{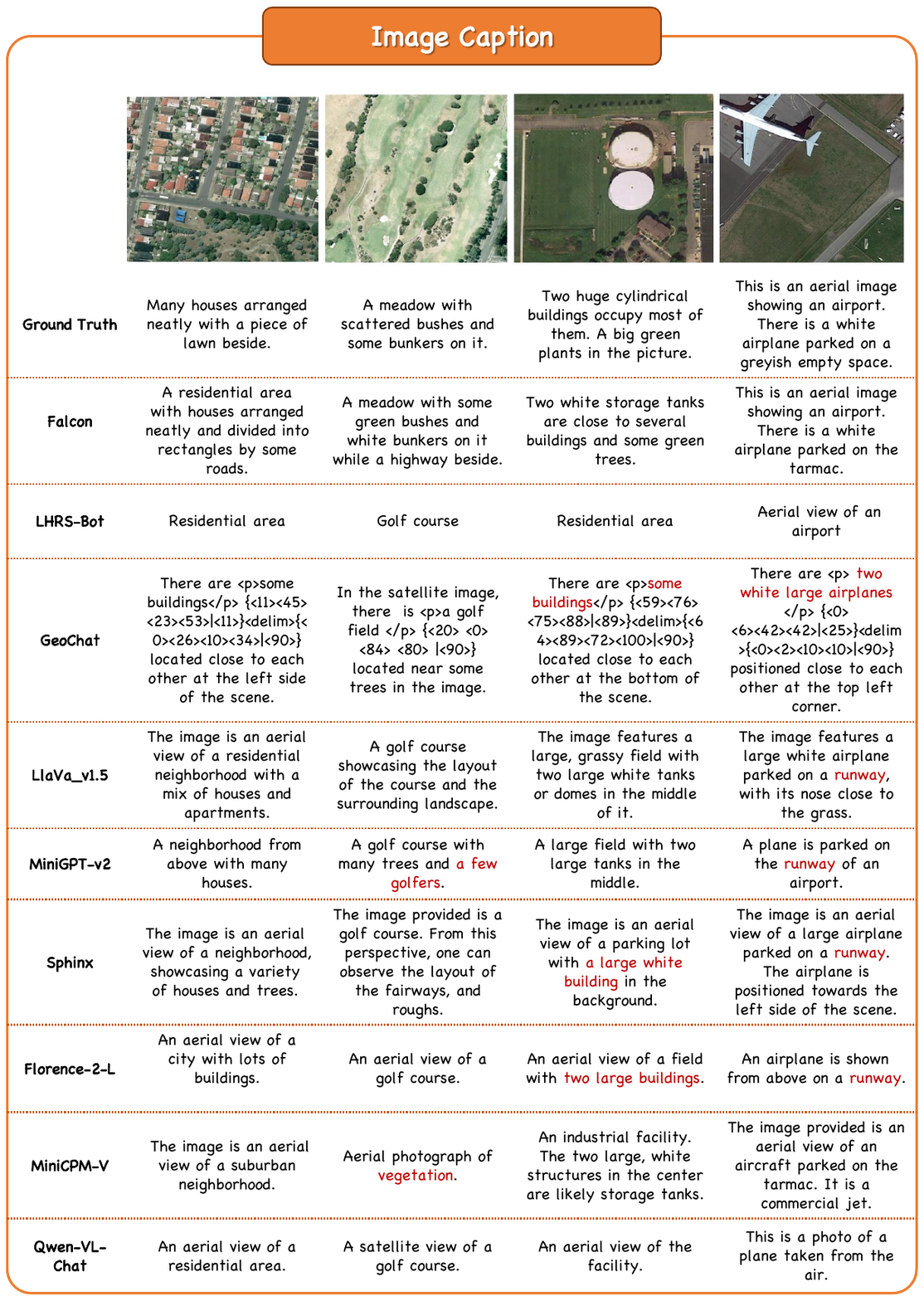}
   \caption{Qualitative comparisons in the task of image captioning. The hallucinations of other VLMs are highlighted in red.}
   \label{task4_fig}
\end{figure*}

\clearpage
\subsection{Task5: Detailed Image Captioning}


\begin{figure*}[ht]
  \centering
   \includegraphics[width=0.9\linewidth]{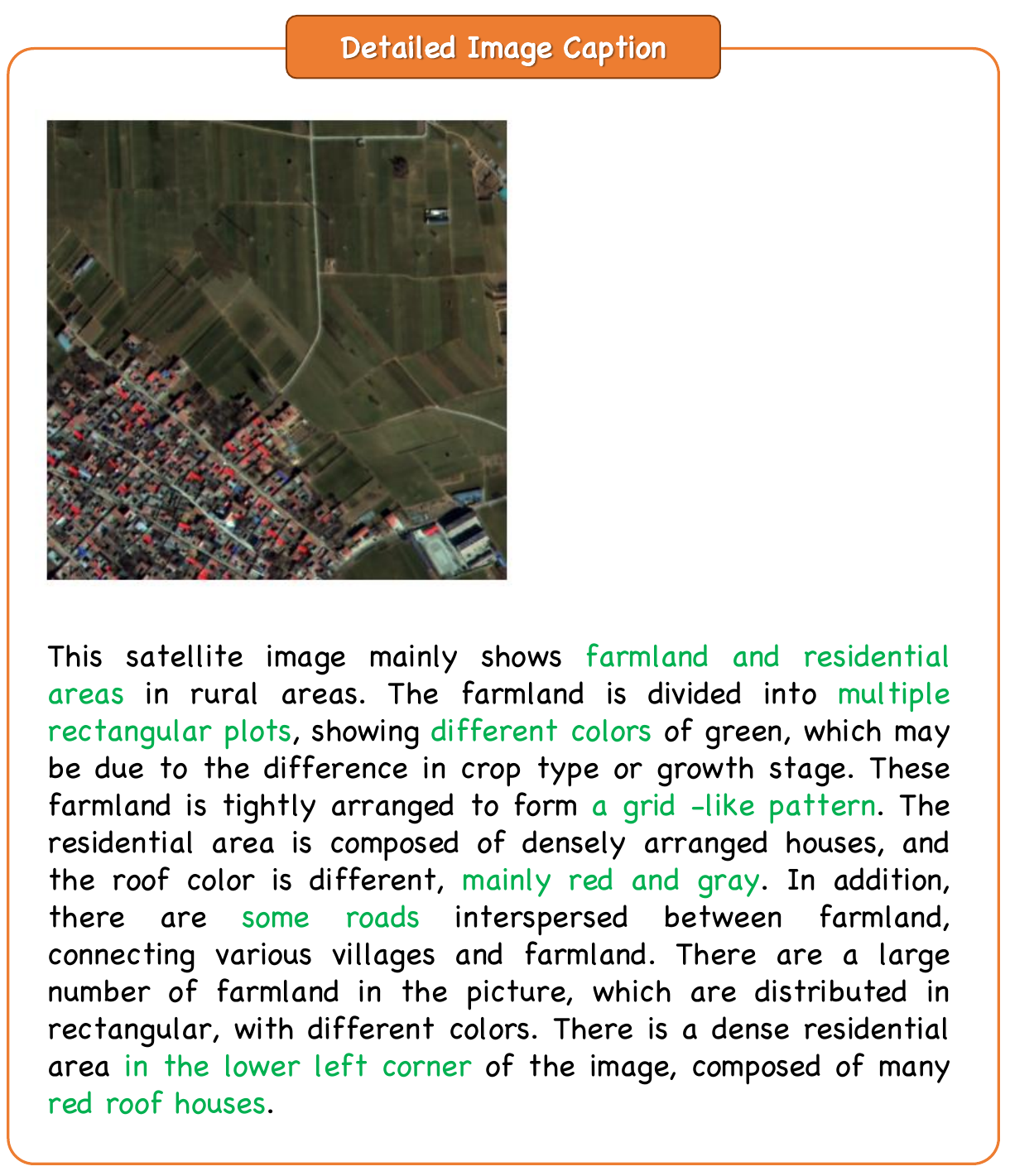}
   \caption{Qualitative results in the task of detailed image captioning from Falcon. Some key correct information is highlighted in green.}
   \label{task5_fig_1}
\end{figure*}

\begin{figure*}[ht]
  \centering
   \includegraphics[width=0.9\linewidth]{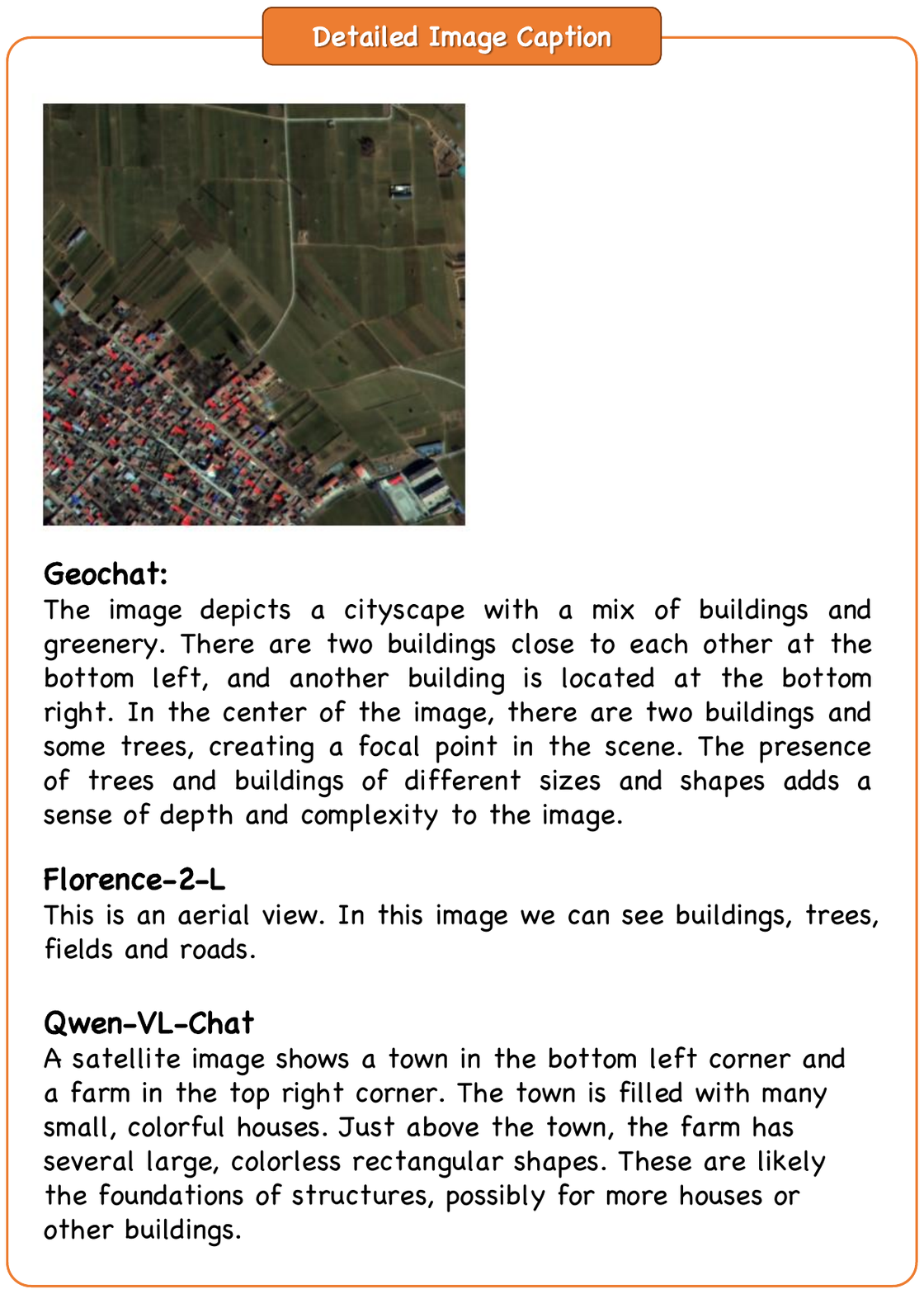}
   \caption{Qualitative results in the task of detailed image captioning from other VLMs. In comparison, Falcon provided more detailed and accurate information.}
   \label{task5_fig_1.5}
\end{figure*}

\begin{figure*}[ht]
  \centering
   \includegraphics[width=0.9\linewidth]{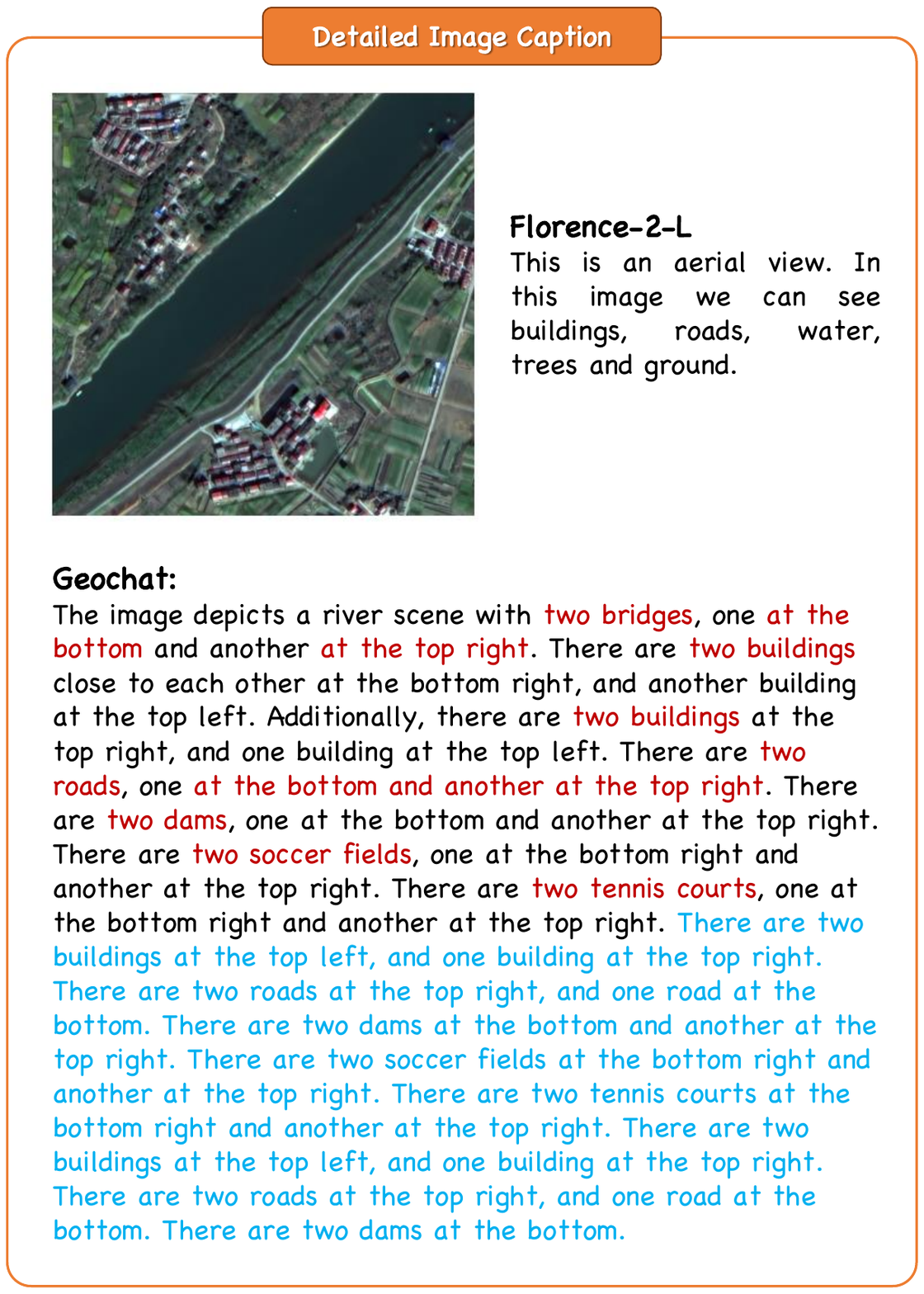}
   \caption{Qualitative results in the task of detailed image captioning from other VLMs. The hallucinations of other VLMs are highlighted in red and some repetitive sentences are marked in blue. In comparison, Falcon provided more detailed and accurate information.}
   \label{task5_fig_1.55}
\end{figure*}

\begin{figure*}[ht]
  \centering
   \includegraphics[width=0.9\linewidth]{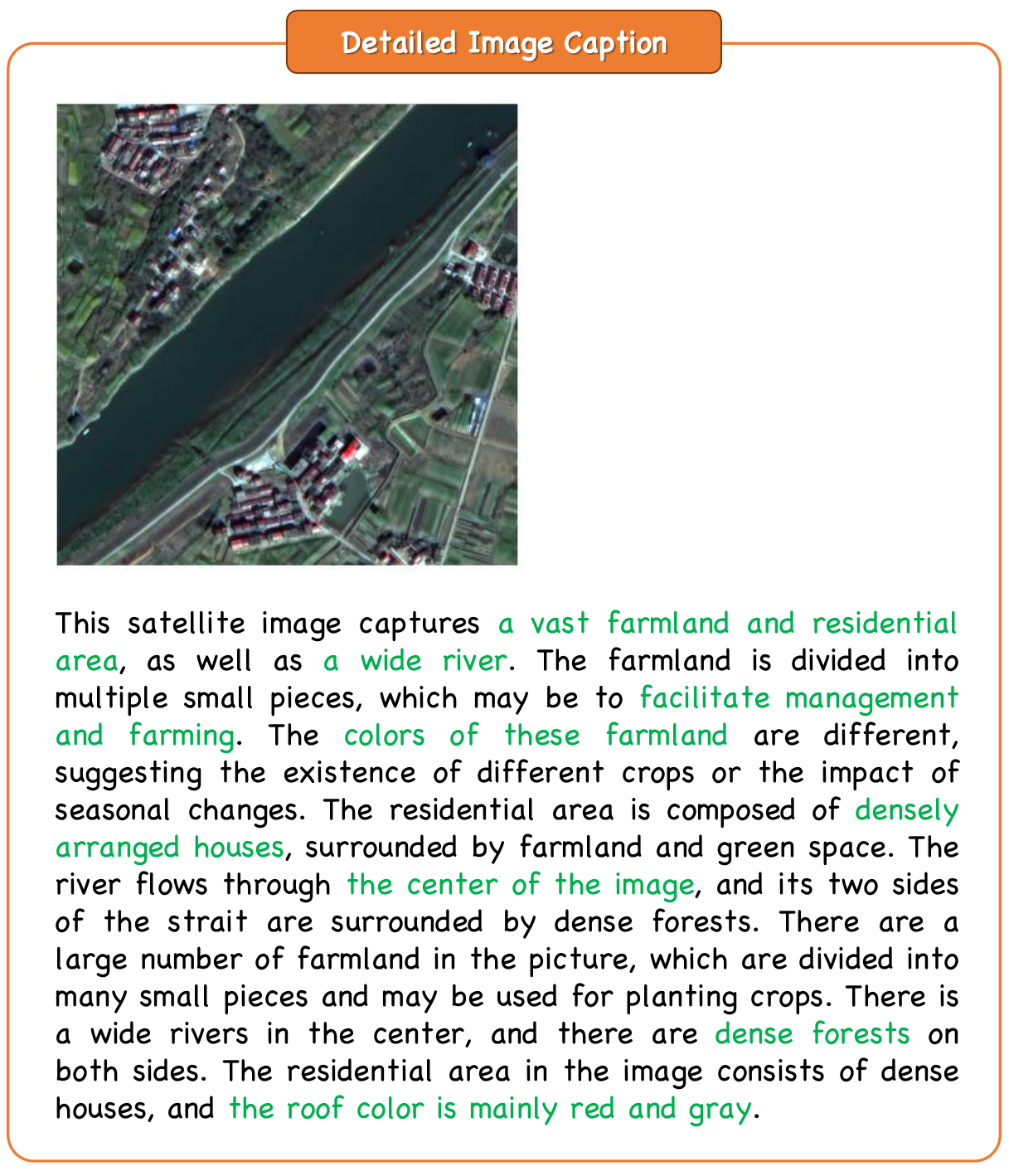}
   \caption{Qualitative results in the task of detailed image captioning from Falcon. Some key correct information is highlighted in green.}
   \label{task5_fig_2}
\end{figure*}

\clearpage
\subsection{Task6: Region Classification-HBB}

\begin{figure*}[ht]
  \centering
   \includegraphics[width=0.8\linewidth]{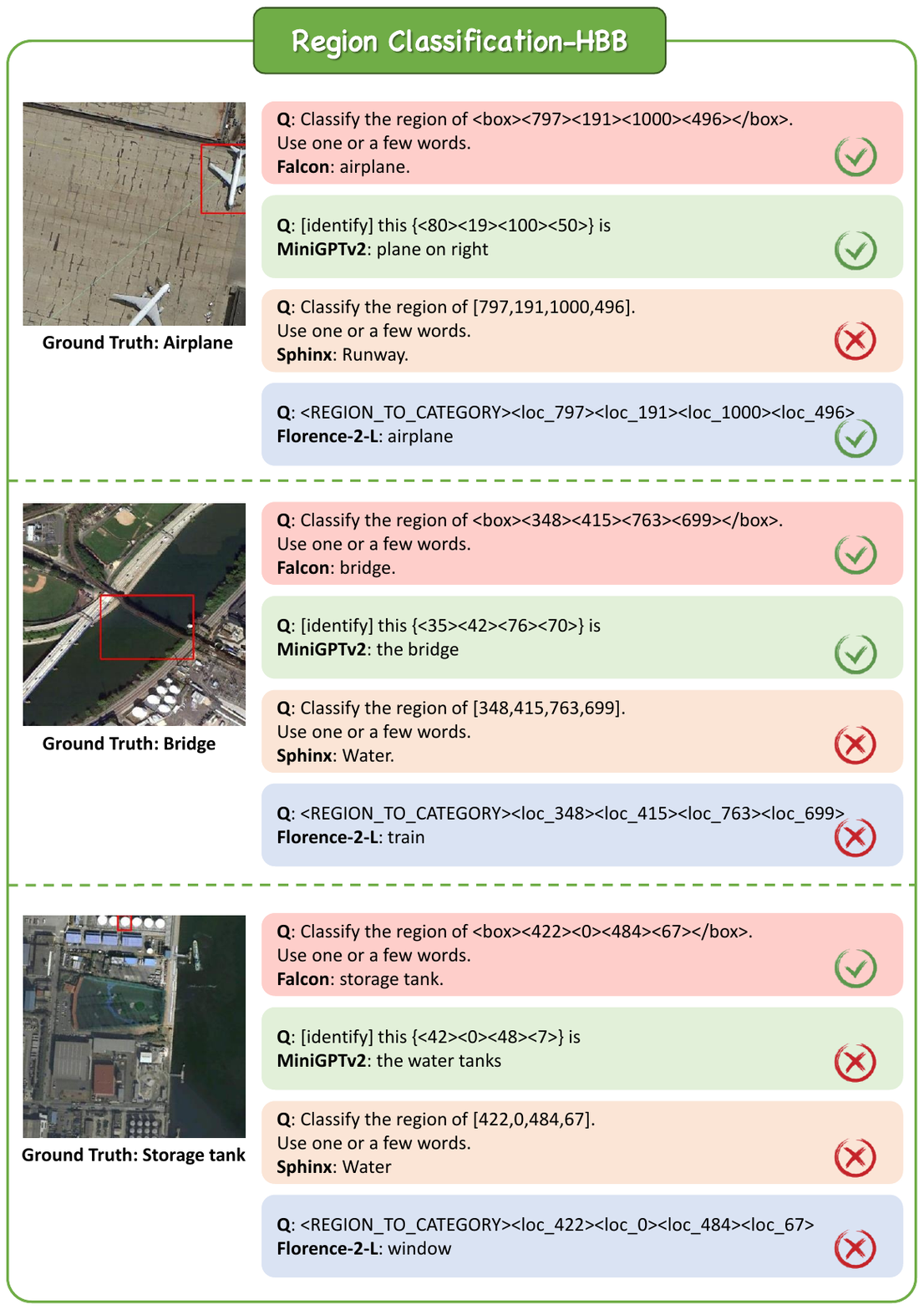}
   \caption{Qualitative comparisons in the task of region classification-HBB. The red bounding box in each image is for visualization only.}
   \label{task6_fig}
\end{figure*}

\clearpage
\subsection{Task7: Region Classfication-OBB}

\begin{figure*}[ht]
  \centering
   \includegraphics[width=0.9\linewidth]{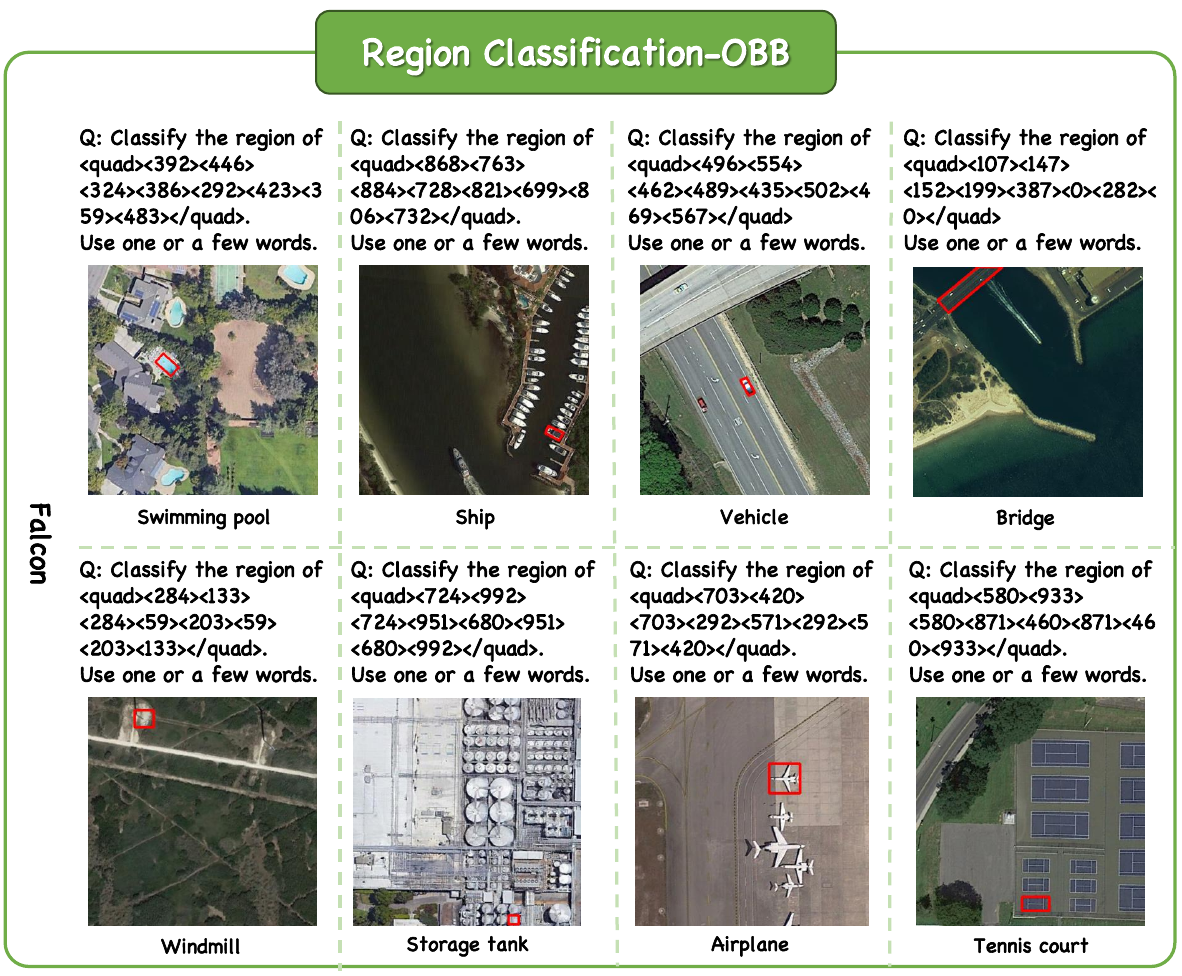}
   \caption{Qualitative results in the task of region classification-OBB. The red bounding box in each image is for visualization only.}
   \label{task7_fig}
\end{figure*}

\clearpage
\subsection{Task8: Region Detection-HBB}

\begin{figure*}[ht]
  \centering
   \includegraphics[width=0.80\linewidth]{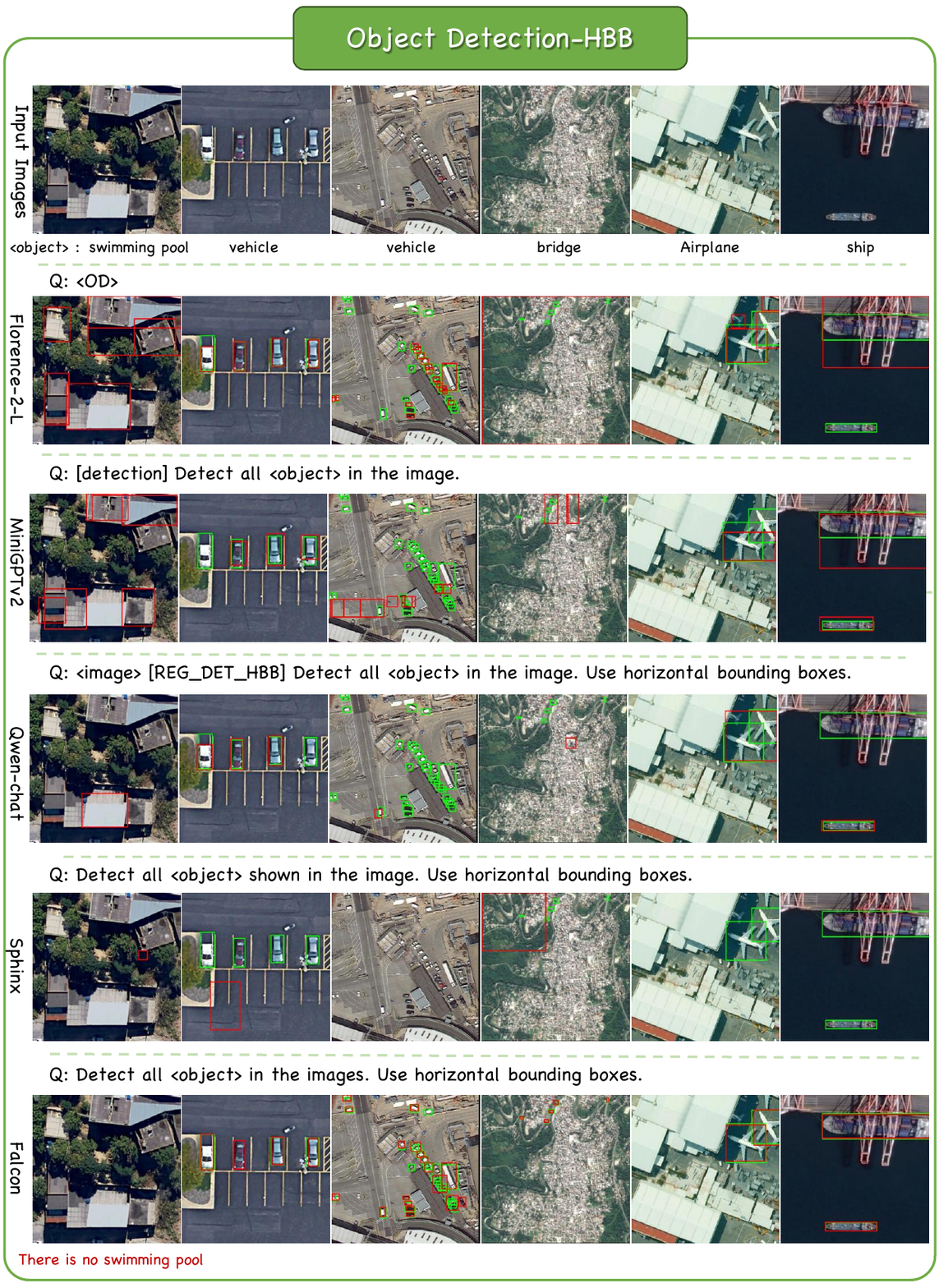}
   \caption{Qualitative comparisons in the task of region detection-HBB. The red bounding boxes are the prediction results, while the green bounding boxes are the ground truth. Falcon successfully detected objects with tiny sizes.}
   \label{task8_fig}
\end{figure*}

\clearpage
\subsection{Task9: Region Detection-OBB}

\begin{figure*}[ht]
  \centering
   \includegraphics[width=0.80\linewidth]{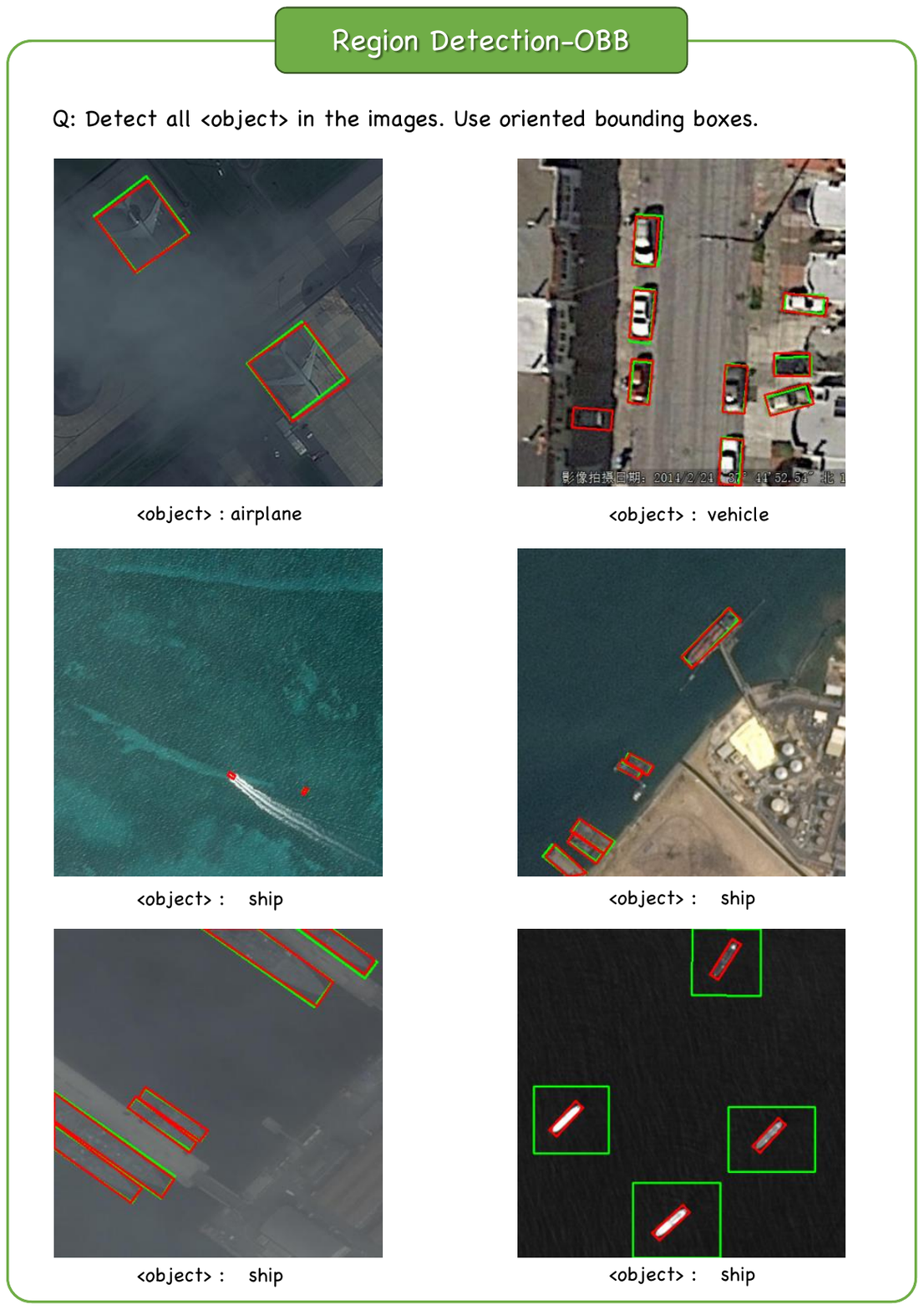}
   \caption{Qualitative results in the task of region detection-OBB. The red bounding boxes are the prediction results, while the green bounding boxes are the ground truth. Falcon successfully detected objects with occlusions. To point out, Falcon also provided more accurate detections than the original annotations in the last image.}
   \label{task9_fig}
\end{figure*}


\clearpage
\subsection{Task10: Visual Grounding}
\begin{figure*}[ht]
  \centering
   \includegraphics[width=0.8\linewidth]{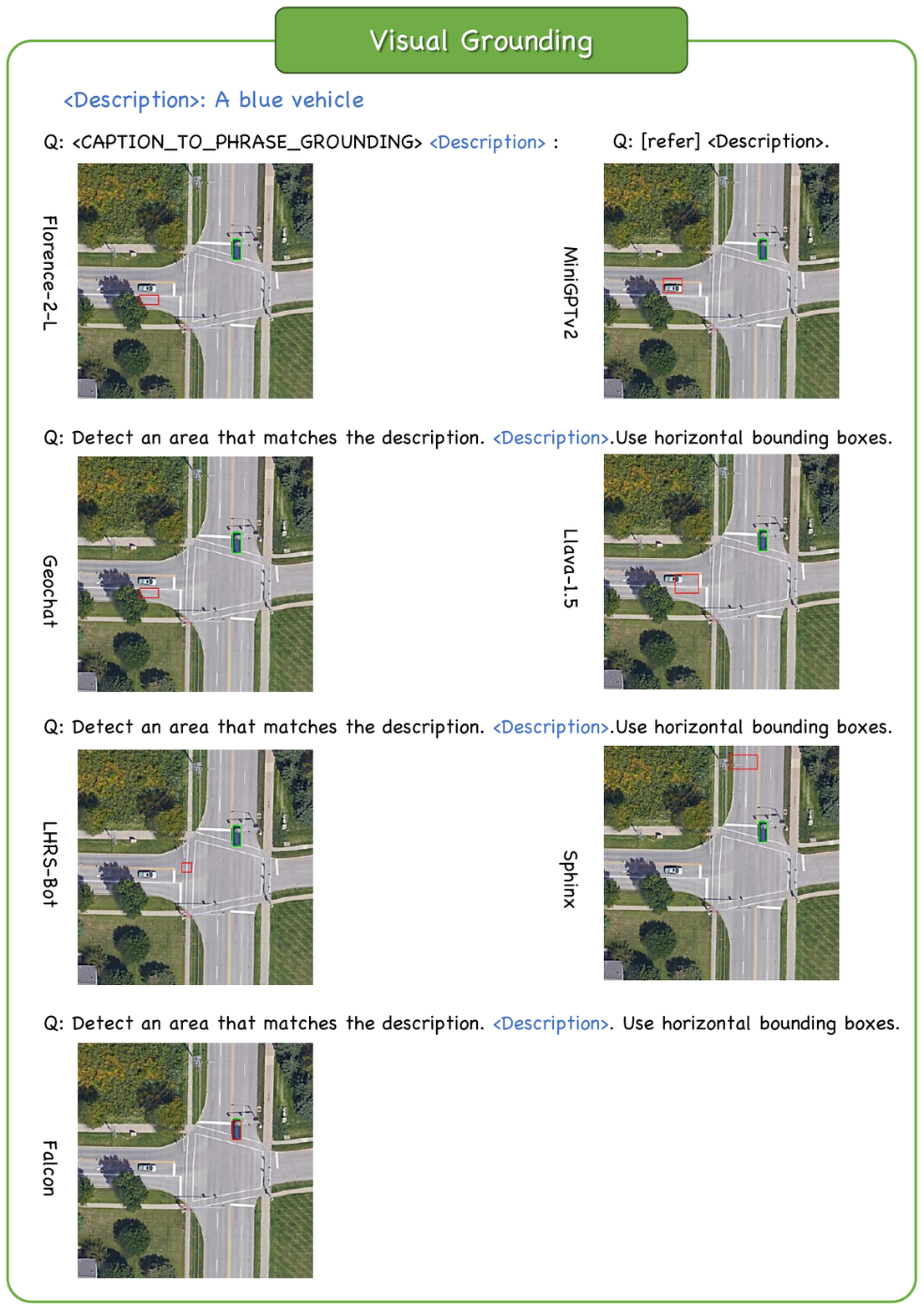}
   \caption{Qualitative comparisons in the task of visual grounding.}
   \label{task10_fig_1}
\end{figure*}

\begin{figure*}[ht]
  \centering
   \includegraphics[width=0.8\linewidth]{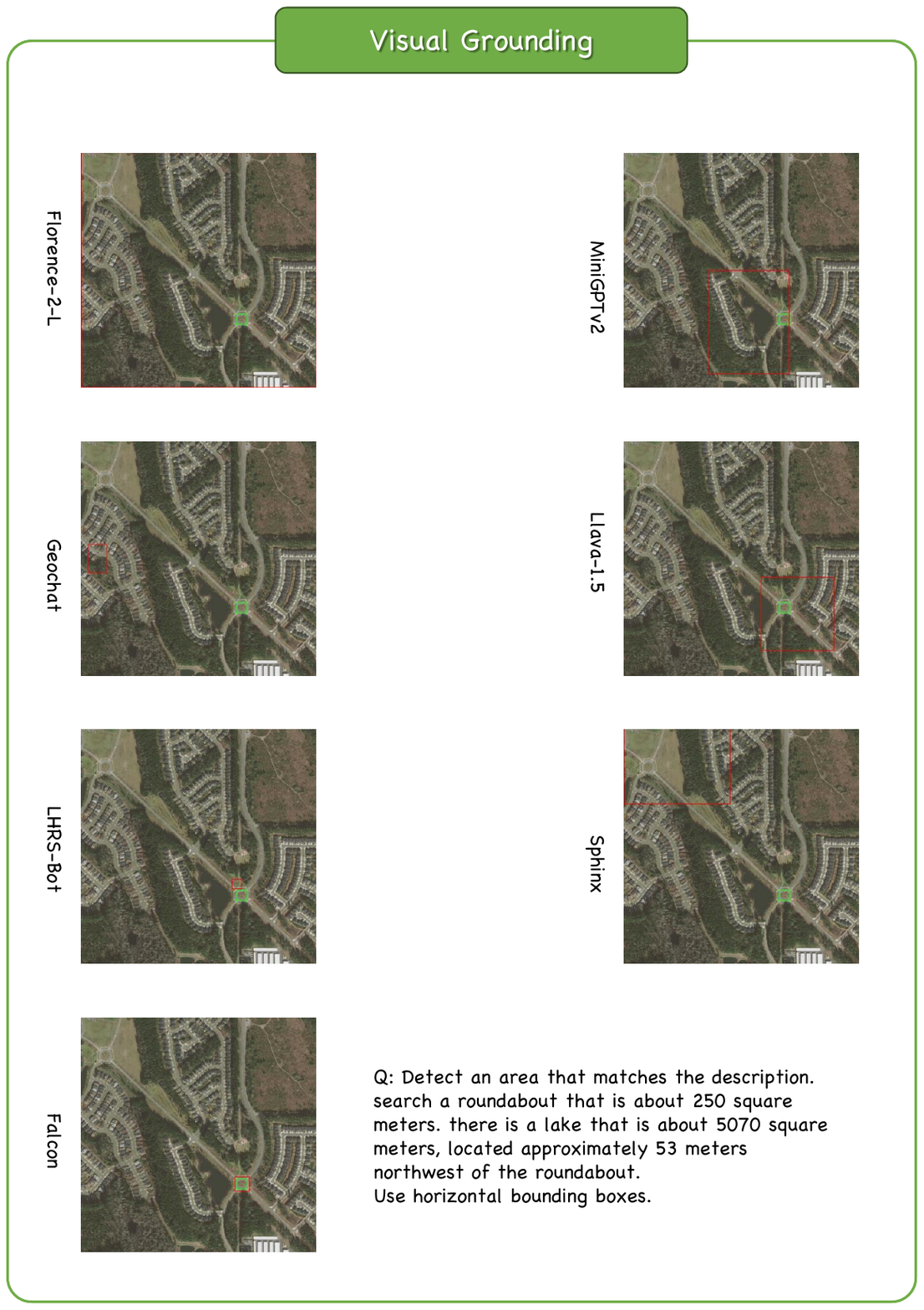}
   \caption{Qualitative comparisons in the task of visual grounding.}
   \label{task10_fig_2}
\end{figure*}

\clearpage
\subsection{Task11: Region Captioning}

\begin{figure*}[ht]
  \centering
   \includegraphics[width=0.95\linewidth]{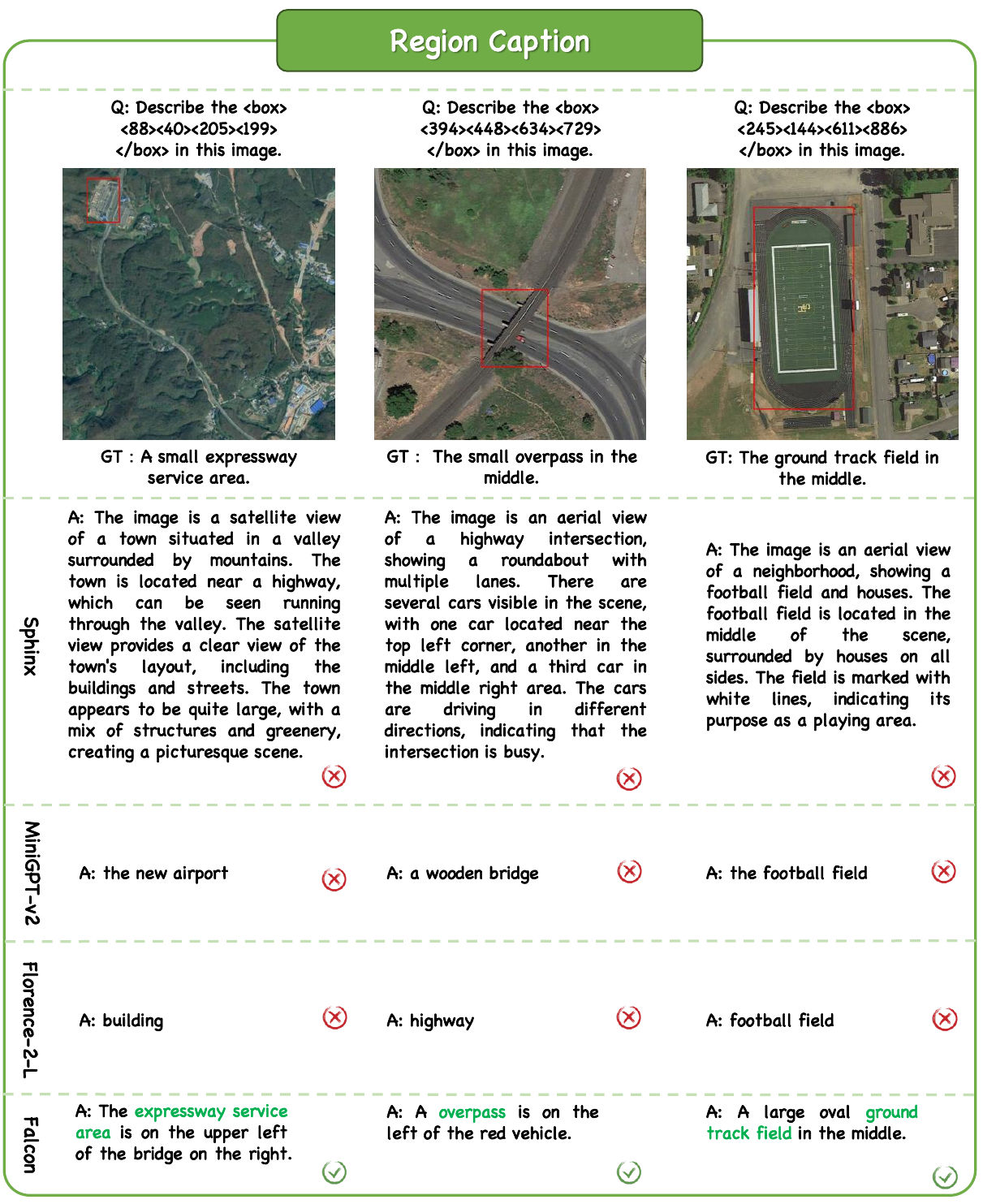}
   \caption{Qualitative comparisons in the task of region captioning.}
   \label{task11_fig}
\end{figure*}

\clearpage
\subsection{Task12: Pixel Classification}
\begin{figure*}[ht]
  \centering
   \includegraphics[width=0.8\linewidth]{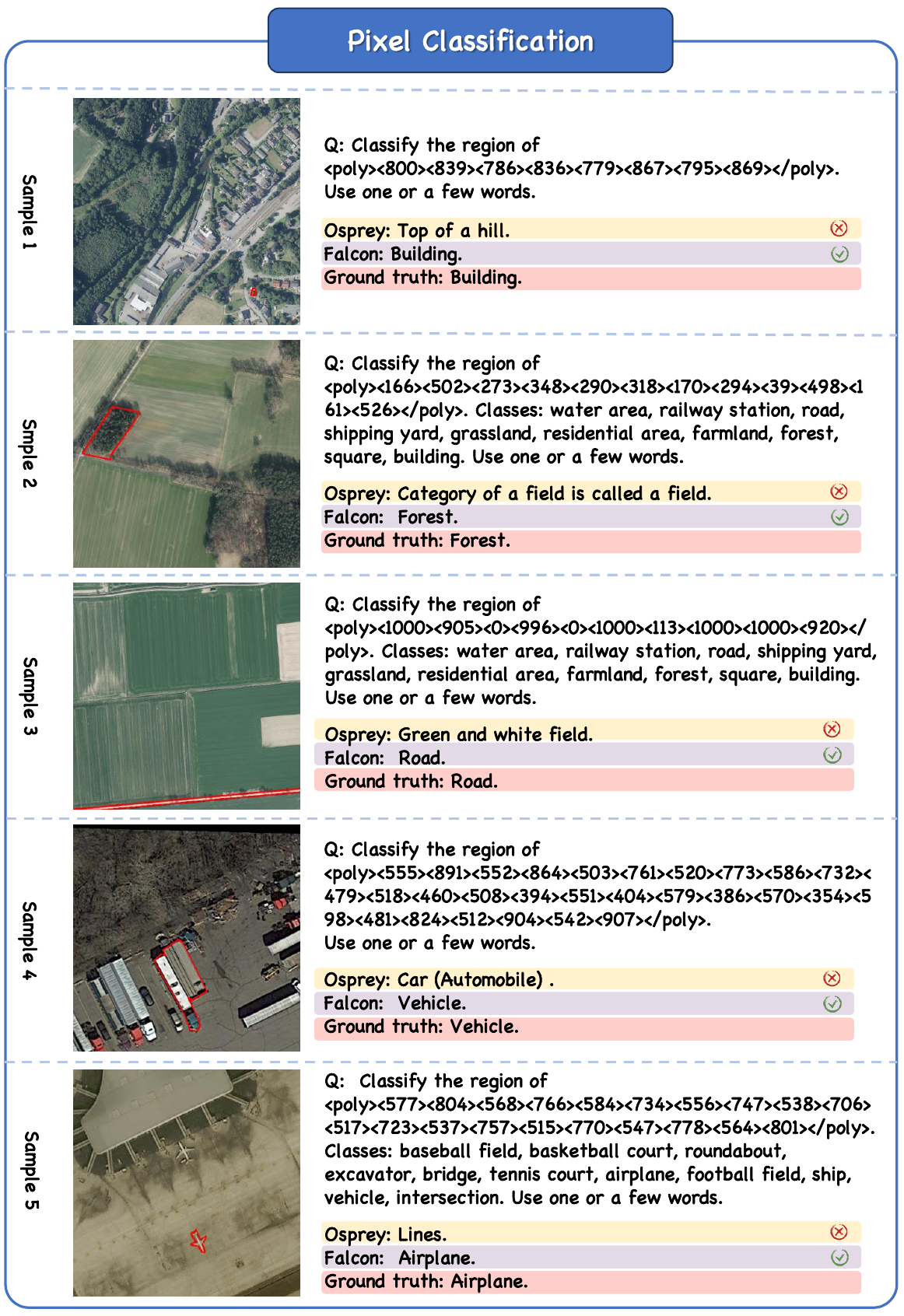}
   \caption{Qualitative comparisons in the task of pixel classification.}
   \label{task12_fig}
\end{figure*}

\clearpage
\subsection{Task13: Segmentation}

\begin{figure*}[ht]
  \centering
   \includegraphics[width=0.8\linewidth]{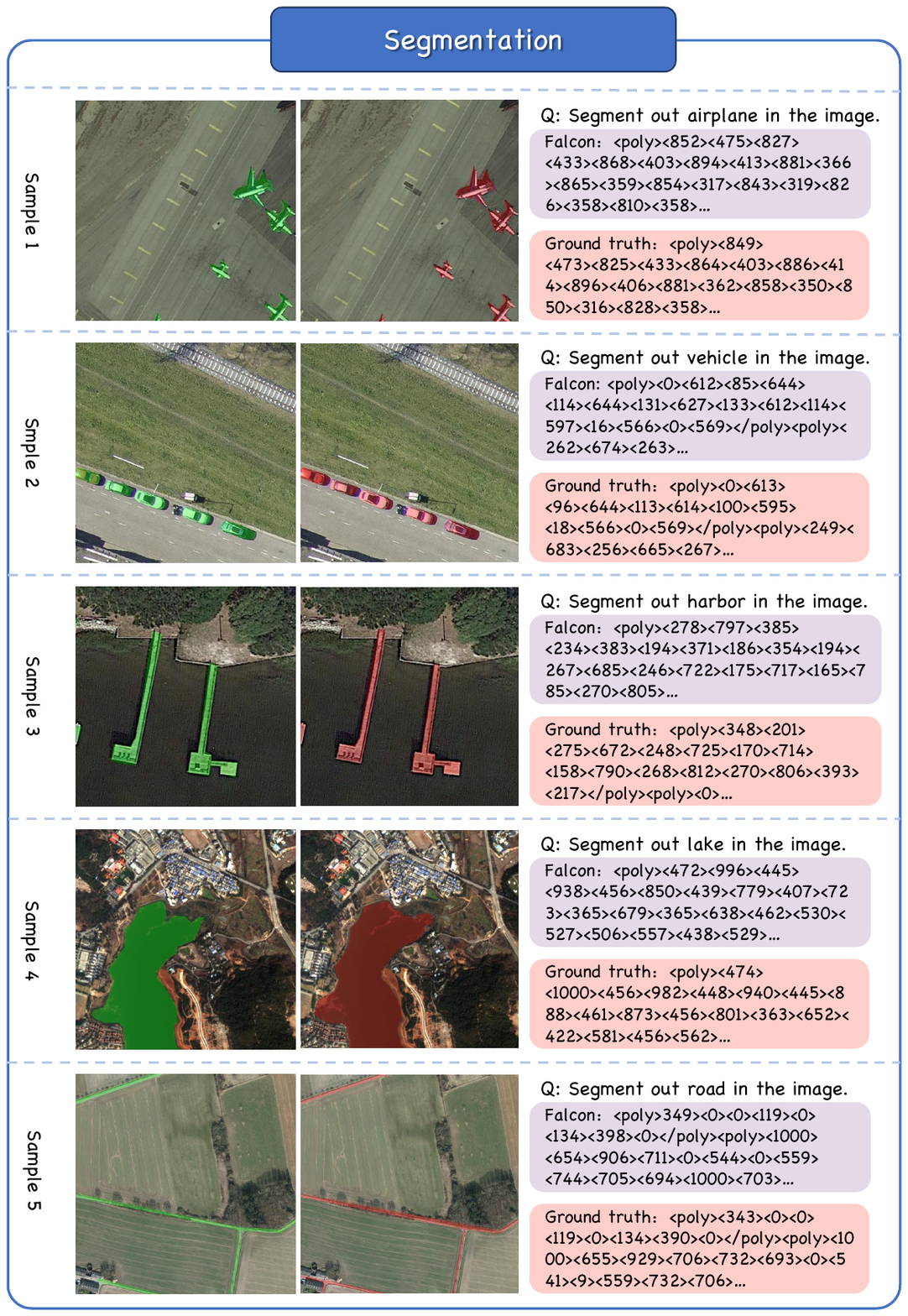}
   \caption{Qualitative results from Falcon in the task of segmentation.}
   \label{task13_fig}
\end{figure*}

\clearpage
\subsection{Task14: Change Detection}

\begin{figure*}[ht]
  \centering
   \includegraphics[width=0.8\linewidth]{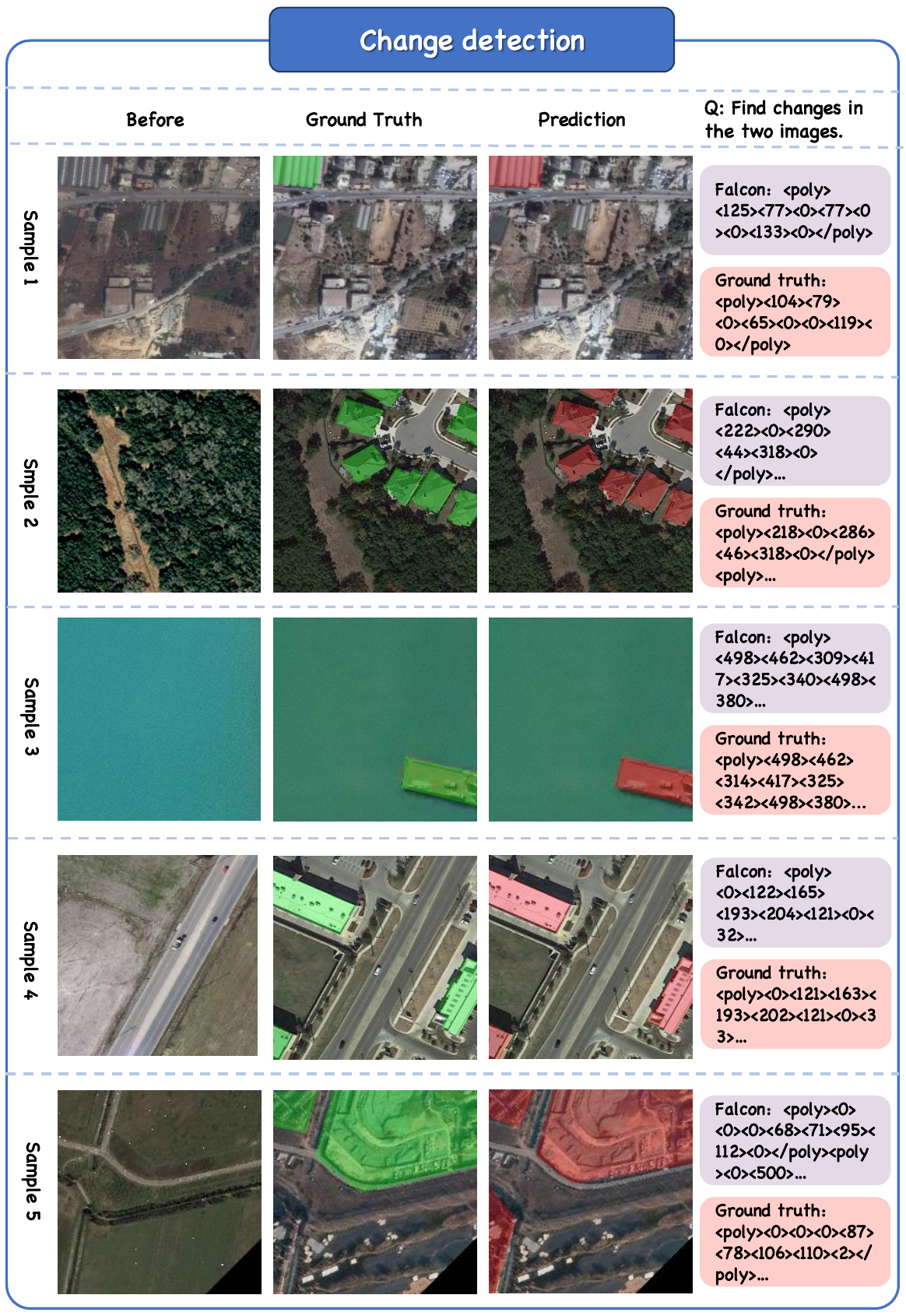}
   \caption{Qualitative results from Falcon in the task of change detection.}
   \label{task14_fig}
\end{figure*}

\clearpage
\section{Quantitative comparison results for remaining tasks}
\label{supp:sec_F}
In this section, we first presented the performance of Falcon over image captioning and detailed image captioning tasks (\textit{c.f.} Tab.\;\ref{tab:cap_dcap_performance}), region classification tasks (\textit{c.f.} Tab.\;\ref{tab:region_classification_performance}), visual grounding  tasks (\textit{c.f.} Tab.\;\ref{tab:visual_grounding_performance}) and image region caption tasks (\textit{c.f.} Tab.\;\ref{tab:region_caption_performance}). 
As shown in Tab.\;\ref{tab:cap_dcap_performance} to Tab.\;\ref{tab:region_caption_performance}, general VLMs, such as MiniGPTv2 \cite{zhu2023minigpt} and Qwen-VL-Chat \cite{2023Qwen-VL} encountered obstacles in performing effectively on remote sensing data, since they usually lacked the expert knowledge of this domain.
Meanwhile, compared with VLMs specialized in remote sensing \cite{kuckreja2024geochat,muhtar2024lhrs}, Falcon achieved better performance in all related datasets, with only 0.7B parameters.
Besides, we also provided detailed task performance of region classification with oriented bounding box, object detection with oriented bounding box, semantic segmentation and change detection in Tab.\;\ref{tab:obb_region_classification_performance}, Tab.\;\ref{tab:obb_detection_performance} , Tab.\;\ref{tab:segmentation_performance} and Tab.\;\ref{tab:cd_performance}.

\begin{table}[htbp]
\centering
\small
\begin{tabular}{ccccccccc}
\hline
\multirow{3}{*}{Models} & \multirow{3}{*}{\#params} & \multicolumn{4}{c}{Image caption} & \multicolumn{3}{c}{Detailed image caption} \\
 &  & \multicolumn{4}{c}{(CIDEr)} & \multicolumn{3}{c}{(CIDEr)} \\
 \cmidrule(lr){3-6} \cmidrule(lr){7-9}
 &  & RSICD & RSICap & RSITMD & Sydney\_Captions & RSICD & RSICap & RSITMD \\ \hline
MiniCPM-V\cite{hu2024minicpm} & 3B & 0.000 & 0.000 & 0.000 & 0.000 & 0.001 & 0.547 & 0.005 \\
MiniGPT-v2\cite{zhu2023minigpt} & 7B & 12.212 & 5.763 & 9.990 & 9.971 & 0.010 & 0.003 & 0.112 \\
Florence-2-L\cite{xiao2024florence} & 0.7B & 10.107 & 4.733 & 8.262 & 6.948 & 0.562 & 0.801 & 1.849 \\
LLaVA-1.5\cite{llava} & 7B & 0.001 & 0.000 & 0.000 & 0.000 & 0.512 & 2.120 & 0.037 \\
Qwen-VL-Chat\cite{2023Qwen-VL} & 7B & 7.603 & 8.275 & 8.660 & 8.871 & 1.070 & 3.118 & 3.896 \\
Sphinx\cite{2023SPHINX} & 7B & 0.001 & 0.000 & 0.000 & 0.000 & 0.773 & 1.368 & 0.723 \\
GeoChat\cite{kuckreja2024geochat} & 7B & 0.342 & 1.653 & 0.418 & 0.415 & 2.243 & 5.191 & 3.695 \\
LHRS-Bot\cite{muhtar2024lhrs} & 7B & 4.492 & 5.222 & 3.588 & 12.810 & 16.195 & 6.119 & 16.956 \\
\textbf{Falcon(Ours)} & 0.7B & \textbf{107.070} & \textbf{58.111} & \textbf{32.323} & \textbf{227.564} & \textbf{39.819} & \textbf{26.009} & \textbf{41.905} \\ \hline
\end{tabular}
\caption{A comparison of image captioning performance and detailed image captioning performance on several datasets with 7B+ generic and remote sensing VLMs.}
\label{tab:cap_dcap_performance}
\end{table}

\begin{table*}[htbp]
\centering
\small
\resizebox{1\columnwidth}{!}{
\begin{tabular}{ccccccccccccccc}
\hline
\multirow{2}{*}{Models} & \multirow{2}{*}{\#params} & \multicolumn{7}{c}{Region   level(with horizontal bounding box)} & \multicolumn{6}{c}{Pixel   level} \\
\cmidrule(lr){3-9} \cmidrule(lr){10-15}
 &  & DIOR & DOTA2.0 & HRSC2016 & RSOD & UCAS-AOD & VHRShips & xView & BHP Watertanks & GEONRW & Globe230k & LoveDA & SAMRS & iSAID \\ \hline
MiniGPT-v2\cite{zhu2023minigpt} & 7B & 0.579 & 0.647 & 0.821 & 0.702 & 0.873 & 0.825 & 0.465 & - & - & - & - & - & - \\
Florence-2-L\cite{xiao2024florence} & 0.7B & 0.433 & 0.731 & 0.461 & 0.712 & 0.939 & 0.369 & 0.811 & - & - & - & - & - & - \\
Sphinx\cite{2023SPHINX} &7B  & 0.174 & 0.106 & 0.549 & 0.505 & 0.389 & 0.482 & 0.021 & - & - & - & - & - & - \\
Osprey\cite{2023Osprey} & 7B & - & - & - & - & - & - & - & 0.009 & 0.060 & 0.123 & 0.076 & 0.258 & 0.276 \\
\textbf{Falcon(Ours)} & 0.7B & \textbf{0.982} & \textbf{0.998} & \textbf{0.999} & \textbf{0.998} & \textbf{0.999} & \textbf{0.999} & \textbf{0.972} & \textbf{0.999} & \textbf{0.908} & \textbf{0.872} & \textbf{0.813} & \textbf{0.834} & \textbf{0.973} \\ \hline
\end{tabular}
}
\caption{A comparison of region classification performance (Accuracy) on several datasets with 7B+ generic and remote sensing VLMs.}
\label{tab:region_classification_performance}
\end{table*}

\begin{table}[htbp]
\centering
\small
\begin{tabular}{cccc}
\hline
\multirow{2}{*}{Models} & \multirow{2}{*}{\#params} & \multicolumn{2}{c}{AP@IoU=0.5(\%)} \\
\cmidrule(lr){3-4}
 &  & DIOR-RSVG & RSVG \\ \hline
MiniGPT-v2\cite{zhu2023minigpt} & 7B & 29.892 & 1.771 \\
Florence-2-L\cite{xiao2024florence} & 0.7B & 16.929 & 1.320 \\
LLaVA-1.5\cite{llava} & 7B & 12.085 & 0.165 \\
Qwen-VL-Chat\cite{2023Qwen-VL} & 7B & 31.528 & 3.627 \\
Sphinx\cite{2023SPHINX} &7B  & 0.939 & 0.000 \\
GeoChat\cite{kuckreja2024geochat} & 7B & 21.024 & 0.741 \\
LHRS-Bot\cite{muhtar2024lhrs} & 7B & 11.826 & 1.318 \\
\textbf{Falcon(Ours)} & 0.7B & \textbf{87.539} & \textbf{56.878} \\ \hline
\end{tabular}
\caption{A comparison of visual grounding performance with horizontal bounding box on several datasets with 7B+ generic and remote sensing VLMs.}
\label{tab:visual_grounding_performance}
\end{table}

\begin{table}[htbp]
\centering
\small
\begin{tabular}{cccccccccc}
\hline
\multirow{2}{*}{Models} & \multirow{2}{*}{\#params} & \multicolumn{4}{c}{DIOR-RSVG} & \multicolumn{4}{c}{RSVG} \\ \cmidrule(lr){3-6}  \cmidrule(lr){7-10} 
 &  & Bleu-4 & Meteor & Rouge\_L & CIDEr & Bleu-4 & Meteor & Rouge\_L & CIDEr \\ \hline
MiniGPT-v2\cite{zhu2023minigpt} & 7B & 1.583 & 0.105 & 21.358 & 17.480 & 0.000 & 0.037 & 10.454 & 1.588 \\
Florence-2-L\cite{xiao2024florence} & 0.7B & 0.000 & 0.033 & 8.956 & 5.459 & 0.000 & 0.003 & 1.008 & 0.349 \\
Sphinx\cite{2023SPHINX} & 7B & 0.000 & 0.163 & 7.535 & 0.329 & 0.382 & 0.166 & 11.781 & 0.220 \\
\textbf{Falcon(Ours)} & 0.7B & \textbf{45.294} & \textbf{0.675} & \textbf{62.932} & \textbf{440.809} & \textbf{24.891} & \textbf{0.477} & \textbf{51.242} & \textbf{99.485} \\ \hline
\end{tabular}
\caption{A comparison of image region caption performance with horizontal bounding box on several datasets with 7B+ generic and remote sensing VLMs.}
\label{tab:region_caption_performance}
\end{table}

\begin{table}[htbp]
\centering
\small
\begin{tabular}{ccccccc}
\hline
\multirow{2}{*}{Models} & \multirow{2}{*}{\#params} & \multicolumn{5}{c}{Accuracy} \\
\cmidrule(lr){3-7}
 &  & DIOR & DOTA2.0 & FAIR1M1.0 & SODA-A & UCAS-AOD \\ \hline
\textbf{Falcon(Ours)} & 0.7B & 0.981 & 0.997 & 0.999 & 0.974 & 0.999 \\
\hline
\end{tabular}
\caption{Region classification performance with oriented bounding box of Falcon.}
\label{tab:obb_region_classification_performance}
\end{table}

\begin{table*}[htbp]
\centering
\small
\resizebox{1\columnwidth}{!}{
\begin{tabular}{cccccccccccccccccc}
\hline
\multirow{2}{*}{Models} & \multirow{2}{*}{\#params} & \multicolumn{16}{c}{AP@IoU=0.5(\%)}                                                                  \\
\cmidrule(lr){3-18}
                         &                           & \rotatebox{85}{ASD} & \rotatebox{85}{BHP Watertanks} & \rotatebox{85}{DIOR}   & \rotatebox{85}{DOTA2.0}  & \rotatebox{85}{FAIR1M1.0} & \rotatebox{85}{GEONRW} & \rotatebox{85}{Globe230k} & \rotatebox{85}{LoveDA} & \rotatebox{85}{S2-SHIPS} & \rotatebox{85}{SODA-A} & \rotatebox{85}{SZTAKI} & \rotatebox{85}{ShipRS} & \rotatebox{85}{UCAS-AOD} & \rotatebox{85}{airplane\_det} & \rotatebox{85}{iSAID}  & \rotatebox{85}{ship\_det}   \\ \hline
\textbf{Falcon(Ours)} & 0.7B & 89.383 & 78.563 & 55.299 & 23.293 & 60.751 & 21.720 & 22.930 & 36.498 & 20.758 & 7.013 & 20.869 & 59.936 & 88.219 & 83.088 & 28.832 & 21.667 \\ \hline
\end{tabular}
}
\caption{Object detection performance with oriented bounding box of Falcon.}
\label{tab:obb_detection_performance}
\end{table*}

\begin{table*}[htbp]
\centering
\small
\begin{tabular}{cccccccc}
\hline
\multirow{2}{*}{Models} & \multirow{2}{*}{\#params} & \multicolumn{6}{c}{mIoU} \\
\cmidrule(lr){3-8}
 &  & BHP Watertanks & GEONRW & Globe230k & LoveDA & SAMRS & iSAID \\ \hline
\textbf{Falcon(Ours)} & 0.7B & 0.684 & 0.473 & 0.521 & 0.435 & 0.754 & 0.517 \\ \hline
\end{tabular}
\caption{Semantic segmentation performance of Falcon.}
\label{tab:segmentation_performance}
\end{table*}

\begin{table}[htbp]
\centering
\small
\begin{tabular}{ccccccccccc}
\hline
\multirow{2}{*}{Models} & \multirow{2}{*}{\#params} & \multicolumn{9}{c}{mIoU} \\ \cline{3-11} 
 &  & DSFIN & EGY\_BCD & HRSCD & LEVIR-CD+ & LEVIR-CD & MSBC & MSOSCD & S2Looking & SYSU-CD \\ \hline
\textbf{Falcon(Ours)} & 0.7B & 0.575 & 0.554 & 0.341 & 0.570 & 0.699 & 0.384 & 0.474 & 0.570 & 0.561 \\ \hline
\end{tabular}
\caption{Change detection performance of Falcon.}
\label{tab:cd_performance}
\end{table}

\clearpage
\section{Qualitative comparisons using diversified instructions}
\label{supp:sec_G}
\begin{figure*}[ht]
  \centering
   \includegraphics[width=0.8\linewidth]{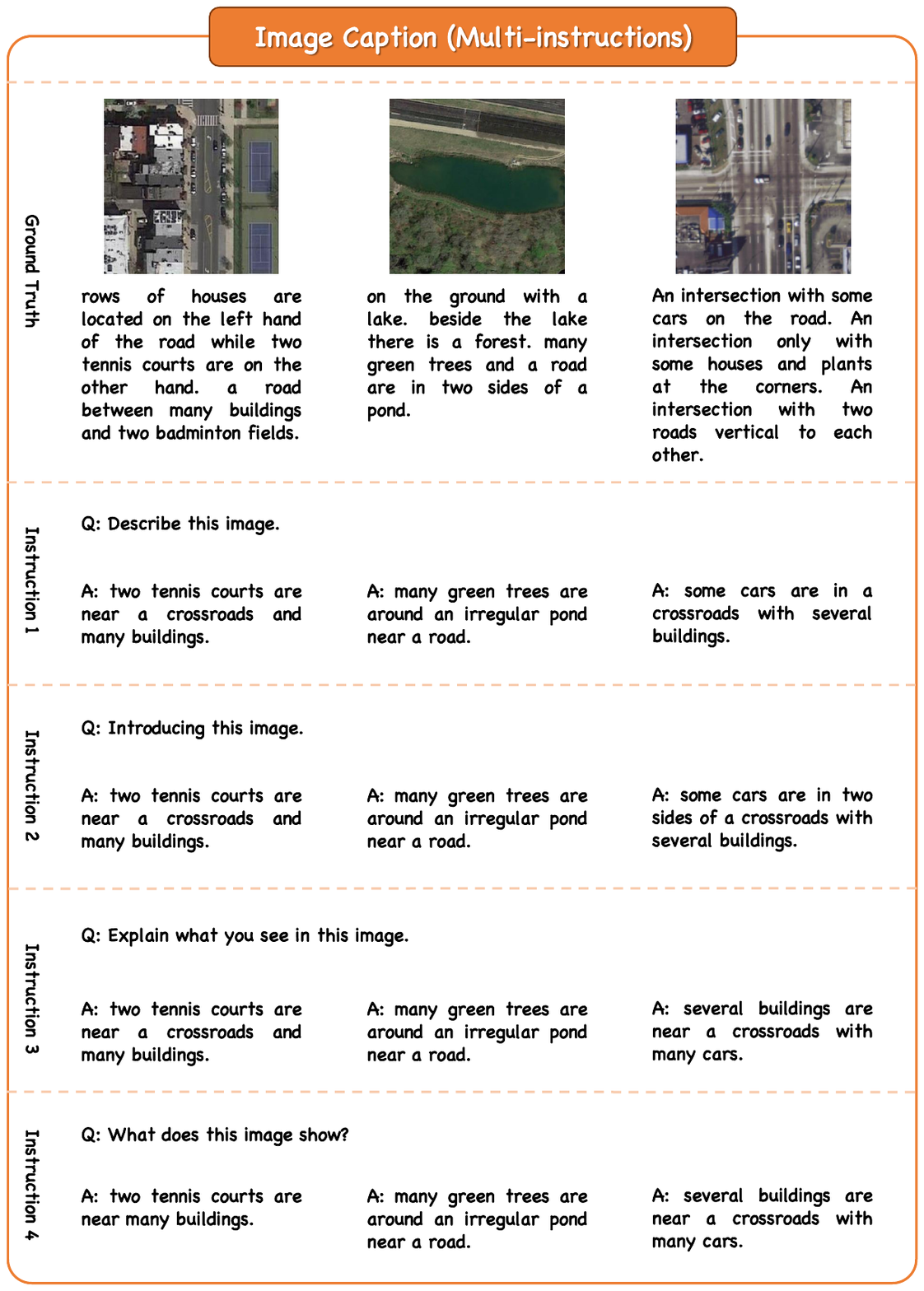}
   \caption{Qualitative comparisons using diversified instructions in the task of image captioning.}
   \label{fig:multi_ins1}
\end{figure*}

\begin{figure*}[ht]
  \centering
   \includegraphics[width=0.8\linewidth]{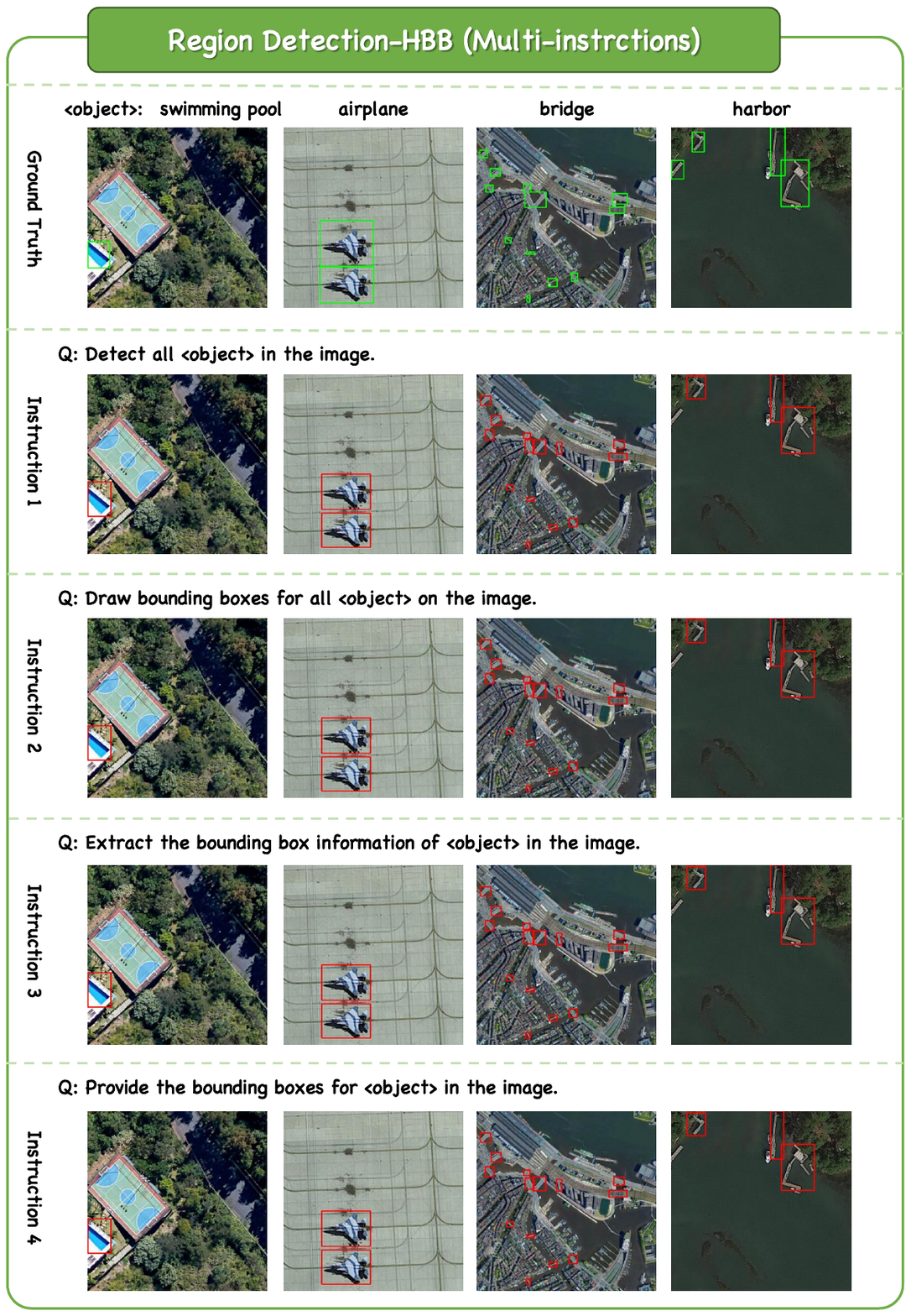}
   \caption{Qualitative comparisons using diversified instructions in the task of Region Detection-HBB.}
   \label{fig:multi_ins2}
\end{figure*}

In this section, we show that Falcon can understand diversified instructions to perform each task, highlighting its instruction-following capabilities.
The results are in Figure \ref{fig:multi_ins1} and Figure \ref{fig:multi_ins2}.

\clearpage
\section{Experiment setup for human evaluation}
\label{supp:sec_H}
Here we provide more details for the human evaluation described in Sec \ref{sec: 5.1}.  We selected 50 images of various types from Falcon\_SFT dataset, including diverse scenes such as urban, rural, and industrial areas, covering amount of labels such as roads, grasslands, buildings, ponds, and farmlands, etc. To ensure the accuracy and reliability of the evaluation, we invited ten volunteers to assess the models' image captioning performance.  They were presented with the images, instruction and model responses, with all the model information anonymized. Following \cite{hu2023rsgpt}, the generated image captions were scored from three dimensions, i.e. detail, position and hallucination description. Each dimension was rated with a four-level rating system as A, B, C, and D. The specific criteria for each level are shown in Tab.\;\ref{tab:caption hunman evaluation}. 

\begin{figure}[ht]
  \centering
   \includegraphics[width=0.95\linewidth]{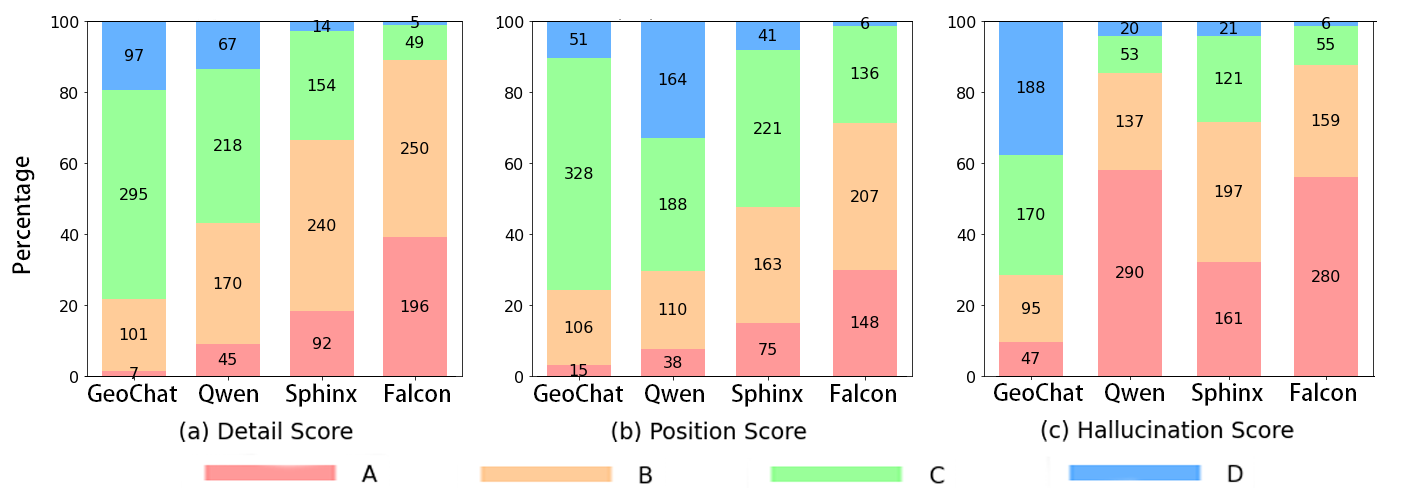}
   \caption{Human evaluation among generic and remote sensing VLMs on image captioning task. }
   \label{fig:human evaluation}
\end{figure}

\begin{figure*}[ht]
  \centering
  \includegraphics[width=0.9\textwidth, trim=0 0 0 0, clip]{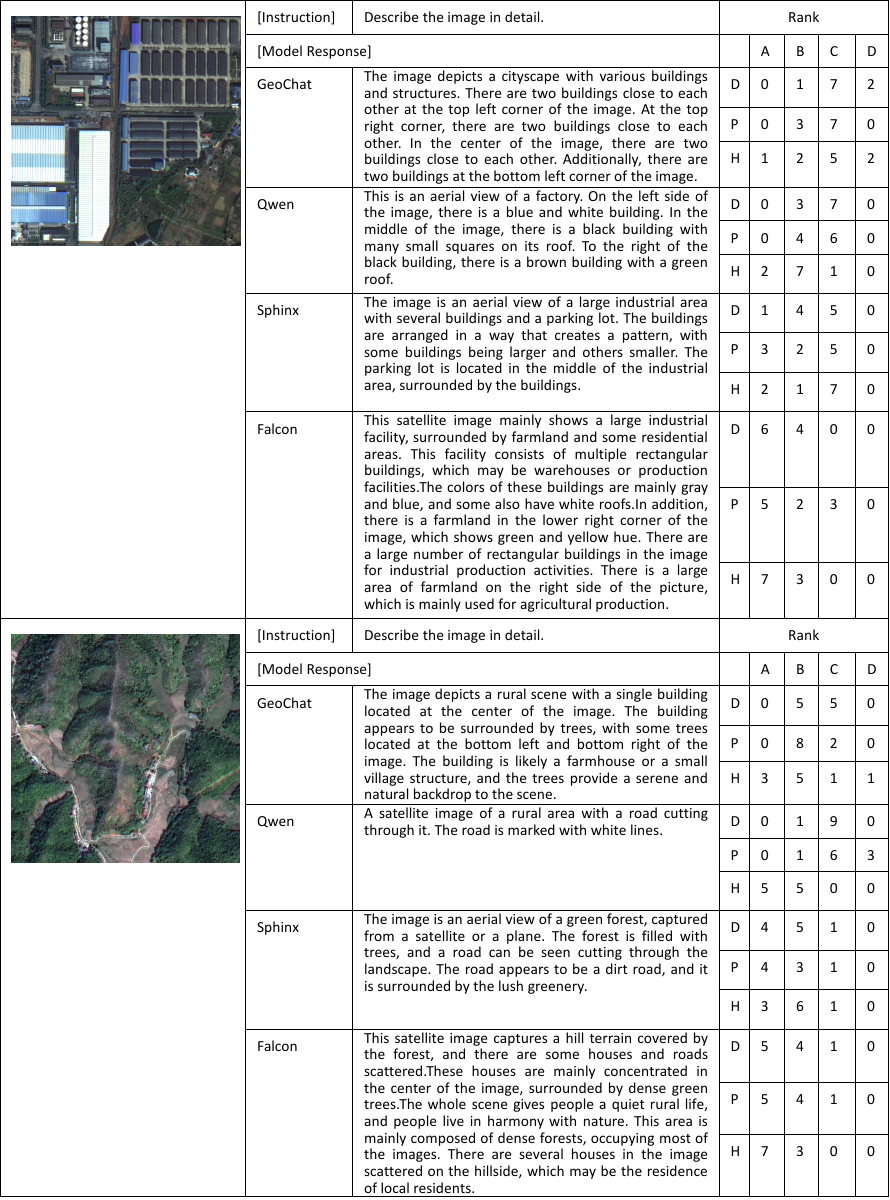}
   \caption{The qualitative comparison among GeoChat, Qwen, Sphinx, and Falcon on the proposed image captioning test set. The number of ratings for detail (D), position (P), and hallucination (H) description is shown in the `Rank' columns.}
   \label{fig:caption}
\end{figure*}

The detailed scoring results are shown in Tab.\;\ref{tab:ranked_comparison} and Fig.\;\ref{fig:human evaluation}. By quantifying the A-D ratings as 4 to 1 points, our model Falcon achieved the highest average scores across all three dimensions, with scores of 3.274, 3.426, and 2.988, respectively. As shown in the Fig.\;\ref{fig:human evaluation}, Falcon received the fewest C and D ratings and the most A and B ratings across all three dimensions. Although Falcon received fewer A ratings than Qwen in the hallucination dimension, examples in Fig.\;\ref{fig:caption} reveal that the outputs of Qwen are overly simplistic. While they avoid hallucination issues, they lack detailed descriptions and accurate positional information. Overall, Falcon consistently outperformed other models, receiving the highest scores in the quantitative analysis and providing the most detailed and accurate descriptions in the qualitative comparison. 

\begin{table}[ht]
\centering
\begin{tabular}{@{}llp{0.75\textwidth}@{}}
\toprule
\textbf{Dimension} & \textbf{Level} & \textbf{Description} \\ \midrule
\multirow{4}{*}{Detail} &
  A &
  The caption has comprehensive and rich details, describing almost all types of objects in the ground truth. \\
& B &
  The caption has rich details, describing most types of objects and their attribute information. \\
& C &
  The caption has only a small amount of details, describing a few types of objects and their attribute information. \\
& D &
  The caption has no detail descriptions. \\ \midrule
\multirow{4}{*}{Position} &
  A &
  The caption has rich position descriptions for objects and all are correct. \\
& B &
  The caption has rich position descriptions for objects with an accuracy higher than 50\%. \\
& C &
  The caption has few position descriptions, or rich position descriptions but with an accuracy less than 50\%. \\
& D &
  The caption has no position descriptions. \\ \midrule
\multirow{4}{*}{Hallucination} &
  A &
  The caption has no hallucination description. \\
& B &
  The caption has hallucination description, and it accounts for less than 50\% \\
& C &
  The caption has a large proportion of hallucination description, more than 50\% \\
& D &
  The caption is entirely hallucination description. \\ \bottomrule
\end{tabular}
\caption{Caption Evaluation Criteria}
\label{tab:caption hunman evaluation}
\end{table}


\clearpage
\section{More ablation experiments}
\label{supp:sec_I}
\begin{figure*}[ht]
  \centering
   \includegraphics[width=0.8\linewidth]{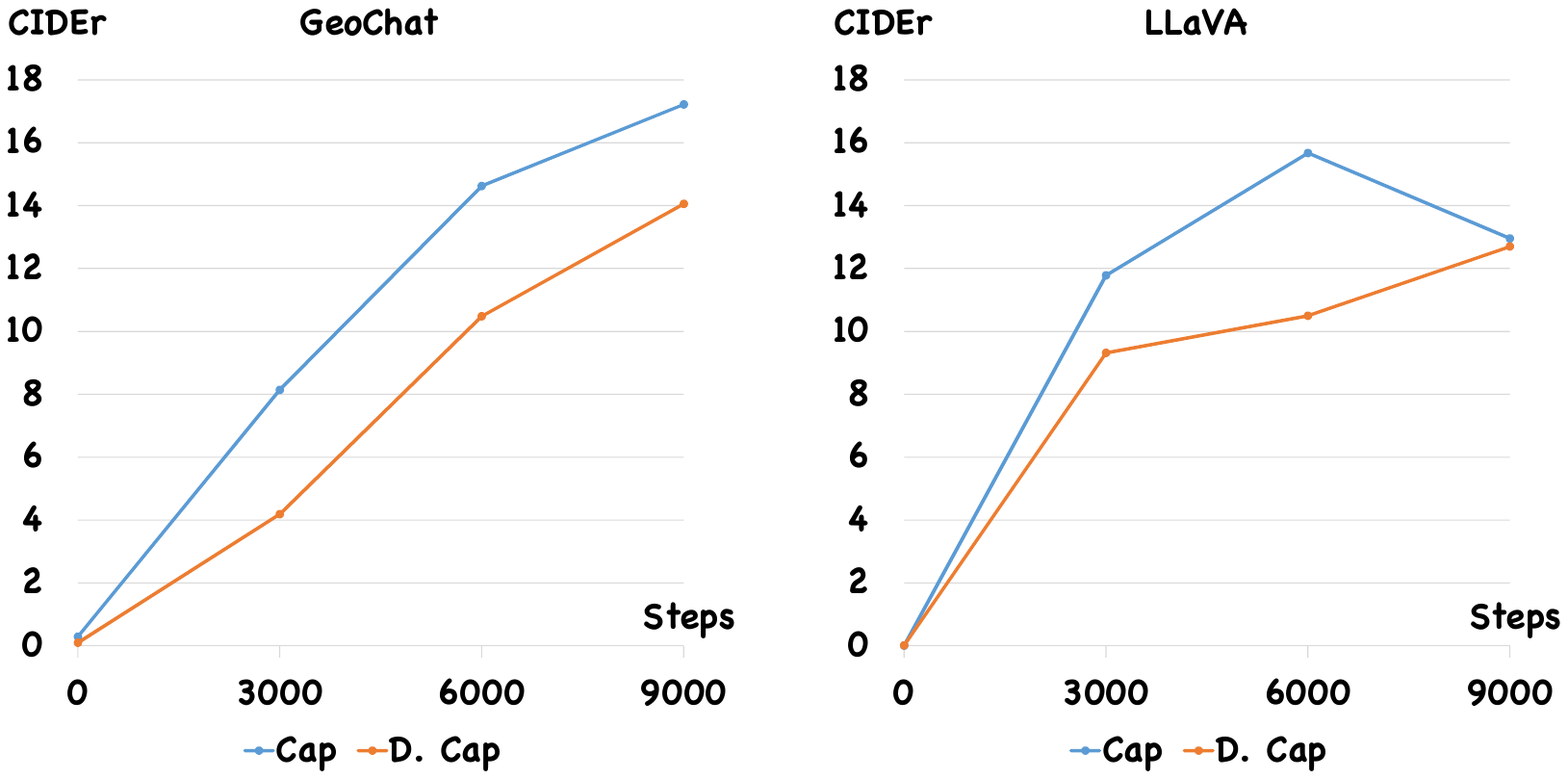}
    \vspace{-20mm}
   \caption{Lora fine-tuning of GeoChat and LLaVA-1.5 on Falcon\_SFT.}
   \label{fig:ablation}
\end{figure*}

This section presents more ablation study results, expecting to demonstrate the effectiveness of our proposed Falcon\_SFT dataset.
Specifically, as shown in Fig.\;\ref{fig:ablation}, we finetune the GeoChat and LLaVA-1.5 on the Falcon\_SFT dataset using LoRA. The performance trends during the first 9000 training steps indicate that the Falcon\_SFT dataset enables the current best models to continue improving their performance.


Additionally, we conducted experiments to verify that Falcon learned robust and generalizable representations.
To this end, we evaluated the model's zero-shot performance while progressively increasing the training data scale. As shown in Tab. \ref{tab:supp_ablation1}, Falcon effectively avoided overfitting, demonstrating its ability to learn stable and transferable representations rather than merely memorizing the training data.

\begin{table}[ht]
\centering
\small
\begin{tabular}{c|c|ccccc}
\hline
\multirow{2}{*}{Data Scale} & Cap (CIDEr) & \multicolumn{1}{c|}{Cls (Accuracy)} & \multicolumn{3}{c|}{Count (Accuracy)} & Seg (mIoU) \\ \cline{2-7} 
 & UCM-Captions & \multicolumn{1}{c|}{AID} & LRBEN & MAR20 & \multicolumn{1}{c|}{NWPU-VHR-10} & GID15 \\ \hline
10\% & 23.630 & \multicolumn{1}{c|}{0.213} & 0.235 & 0.809 & \multicolumn{1}{c|}{0.754} & 0.347 \\
50\% & 26.360 & \multicolumn{1}{c|}{0.252} & 0.223 & 0.820 & \multicolumn{1}{c|}{0.786} & 0.378 \\
100\% & 30.481 & \multicolumn{1}{c|}{0.363} & 0.252 & 0.876 & \multicolumn{1}{c|}{0.798} & 0.390 \\ 
\hline
\hline
\multirow{2}{*}{Data Scale} & D.Cap (CIDEr) & \multicolumn{3}{c|}{Det$^{hbb}$ (AP@IoU=0.5)} & \multicolumn{2}{c}{Det$^{obb}$ (AP@IoU=0.5)} \\ \cline{2-7} 
 & UCM-Captions & GID15 & MAR20 & \multicolumn{1}{c|}{NWPU-VHR-10} & GID15 & MAR20 \\ \hline
10\% & 13.343 & 17.436 & 92.574 & \multicolumn{1}{c|}{77.666} & 9.866 & 77.930 \\
50\% & 28.500 & 19.632 & 92.260 & \multicolumn{1}{c|}{80.771} & 11.232 & 82.370 \\
100\% & 23.554 & 23.548 & 94.189 & \multicolumn{1}{c|}{80.678} & 16.297 & 83.270 \\ \hline
\end{tabular}
\caption{Ablation studies on zero-shot performance of Falcon with increasing training data scale.}
\label{tab:supp_ablation1}
\end{table}

Finally, we conduct ablation experiments on the choice of encoder-decoder
architecture over the decoder-only architecture.
To this end, decoder-only architectures, such as MiniCPM-V \cite{hu2024minicpm}, typically require a large and computationally intensive LLM. 
We believe this type of module is not essential for remote sensing vision tasks and may introduce unnecessary computational overhead. 
As shown in Table \ref{tab:supp_ablation2}, after fine-tuning Falcon and MiniCPM-V \cite{hu2024minicpm} on the same dataset (a subset of FCD), we observed similar performance. 
Notably, Falcon has fewer parameters, which makes it more suitable for deployment.

\begin{table}[ht]
\small
\resizebox{1\columnwidth}{!}{
\begin{tabular}{c|cc|cc|cc|cc|cc|cl}
\hline
\multirow{2}{*}{Models} & \multicolumn{2}{c|}{\begin{tabular}[c]{@{}c@{}}Det$^{hbb}$ \\ AP@IoU=0.5\end{tabular}} & \multicolumn{2}{c|}{\begin{tabular}[c]{@{}c@{}}Det$^{obb}$ \\ Precision@IoU=0.5\end{tabular}} & \multicolumn{2}{c|}{\begin{tabular}[c]{@{}c@{}}Cap\\ Rouge\_L\end{tabular}} & \multicolumn{2}{c|}{\begin{tabular}[c]{@{}c@{}}Cls\\ Accuracy\end{tabular}} & \multicolumn{2}{c|}{\begin{tabular}[c]{@{}c@{}}VQA \\ Accuracy\end{tabular}} & \multicolumn{2}{c}{\begin{tabular}[c]{@{}c@{}}VG \\ AP@IoU=0.5\end{tabular}} \\ \cline{2-13} 
 & DIOR & DOTA2.0 & DIOR & DOTA & RSICD & RSITMD & AID & EuroSAT & HRBEN & LRBEN & \multicolumn{2}{c}{DIOR-RSVG} \\ \hline
\begin{tabular}[c]{@{}c@{}}Decoder-Only\\ (MiniCPM-V {[}20{]})\end{tabular} & 40.277 & 36.440 & 0.533 & 0.454 & 0.474 & 0.290 & 0.984 & 0.971 & 0.816 & 0.740 & \multicolumn{2}{c}{0.749} \\
\begin{tabular}[c]{@{}c@{}}Encoder-Decoder\\ (Falcon)\end{tabular} & 53.666 & 48.262 & 0.886 & 0.831 & 0.507 & 0.361 & 0.987 & 0.975 & 0.817 & 0.752 & \multicolumn{2}{c}{0.871} \\ \hline
\end{tabular}
}
\caption{Ablation studies on the choice of encoder-decoder architecture over the decoder-only architecture.}
\label{tab:supp_ablation2}
\end{table}

\clearpage
\section{Evaluation metric for each task}
\label{supp:sec_J}
\subsection*{J.1 Accuracy}

In our image classification task, we aim to ensure consistent and fair comparisons across different datasets and models, which is challenging due to variations in label naming and class hierarchies in datasets and models. To address this, we leverage BERT (Bidirectional Encoder Representations from Transformers), a powerful language model, to standardize the mapping of classification results into the specific remote sensing classes of each dataset. 
BERT is well-suited for this task because it captures semantic relationships and contextual information in text, making it ideal for aligning class labels that may differ in terminology but share similar meanings. By using BERT to map our classification outputs to consistent class names, we can harmonize class labels across models with different naming conventions or granularities, ensure that semantically similar classes are mapped together, reducing discrepancies in label interpretation, and improve the interpretability of results, especially in multi-source, multi-dataset, multi-model evaluations.

For evaluating classification performance, we use accuracy as our primary evaluation metric, defined as follows:

\begin{equation}
\text{Accuracy} = \frac{\text{Number of Correct Predictions}}{\text{Total Number of Predictions}} = \frac{TP + TN}{TP + TN + FP + FN}
\end{equation}

where \( TP \) (True Positives) and \( TN \) (True Negatives) are the counts of correctly predicted positive and negative instances, respectively. \( FP \) (False Positives) and \( FN \) (False Negatives) are the counts of incorrectly predicted positive and negative instances, respectively.

Using BERT for label mapping and accuracy as the evaluation metric allows us to conduct a more reliable and interpretable comparison of model performance across diverse remote sensing datasets.

\subsection*{J.2 BLEU, METEOR, CIDEr, ROUGE-L}

To evaluate our captioning performance, we use the following metrics: BLEU, METEOR, CIDEr, and ROUGE-L. These metrics provide a comprehensive assessment of the quality of generated captions by measuring different aspects of caption similarity to reference captions. The equations are as follows:

\subsubsection*{J.2.1 \textbf{BLEU (Bilingual Evaluation Understudy)}}
BLEU measures the precision of n-grams (typically up to 4-grams) between the geROUGE-Lnerated captions and reference captions. It is a precision-oriented metric that does not account for recall.

\begin{equation}
\text{BLEU} = \exp\left( \sum_{n=1}^{N} w_n \log p_n \right) \times \text{brevity\_penalty}
\end{equation}

where \( p_n \) is the precision of n-grams, \( w_n \) is the weight for each n-gram level, and the brevity penalty penalizes shorter generated captions.

\subsubsection*{J.2.2 \textbf{METEOR (Metric for Evaluation of Translation with Explicit ORdering) }}
METEOR considers both precision and recall and uses the harmonic mean (F1 score) of these measures. It includes stemming and synonym matching, making it more robust to variations in word choice.

\begin{equation}
\text{METEOR} = F_{\text{mean}} \times (1 - \text{penalty})
\end{equation}

where \( F_{\text{mean}} \) is the harmonic mean of precision and recall, and the penalty term penalizes longer, less precise matches.

\subsubsection*{J.2.3 \textbf{CIDEr (Consensus-based Image Description Evaluation) }}
CIDEr measures consensus between generated and reference captions by applying term frequency-inverse document frequency (TF-IDF) weighting, which emphasizes terms that are more descriptive.

\begin{equation}
   \text{TF-IDF}(t, s) = \text{TF}(t, s) \times \text{IDF}(t)
\end{equation}

\begin{equation}
   \text{CIDEr}_n(g, r) = \frac{\sum_{t \in g \cap r} \text{TF-IDF}(t, g) \times \text{TF-IDF}(t, r)}{\sqrt{\sum_{t \in g} (\text{TF-IDF}(t, g))^2} \cdot \sqrt{\sum_{t \in r} (\text{TF-IDF}(t, r))^2}}
\end{equation}

\begin{equation}
    \text{CIDEr}(g) = \frac{1}{N} \sum_{n=1}^{N} \frac{1}{|R|} \sum_{r \in R} \text{CIDEr}_n(g, r)
\end{equation}

\begin{equation}
    \text{CIDEr} = \frac{1}{|G|} \sum_{i=1}^{|G|} \text{CIDEr}(g_i)
\end{equation}

where \( \text{TF}(t, s) \) is the term frequency of \( t \) in \( s \), and \( \text{IDF}(t) = \log \frac{N}{1 + n(t)} \), with \( N \) as the total number of captions and \( n(t) \) as the number of captions containing \( t \). \( R \) is the set of reference sentences, \( |G| \) is the total number of generated captions.

\subsubsection*{J.2.4 \textbf{ROUGE-L (Recall-Oriented Understudy for Gisting Evaluation)}}
The ROUGE-L metric measures the longest common subsequence (LCS) between a generated caption \( G \) and a reference caption \( R \). Given \( LCS(G, R) \) as the length of the longest common subsequence, calculated as follows:

\begin{equation}
\text{Recall} = \frac{LCS(G, R)}{|R|}
\end{equation}

\begin{equation}
\text{Precision} = \frac{LCS(G, R)}{|G|}
\end{equation}

\begin{equation}
\text{ROUGE-L} = \frac{(1 + \beta^2) \times \text{Precision} \times \text{Recall}}{\text{Recall} + \beta^2 \times \text{Precision}}
\end{equation}

where \( \beta \) is typically set to 1, giving equal importance to precision and recall.



\subsection*{J.3 AP@IoU=0.5}

In our object detection and visual grounding tasks, we employ the evaluation metric AP@IoU=0.5(AP@50) to assess the performance of each model accurately across Horizontal Bounding Box (HBB) detection, Oriented Bounding Box (OBB) detection, and visual grounding. AP@50 is defined as Average Precision (AP) at 50\% Intersection over Union (IoU) threshold. It measures the average precision of detections where the IoU between predicted and ground-truth bounding boxes is at least 50\%. AP@50 balances precision and recall across different confidence thresholds, providing a summary metric of detection quality. It is calculated as:

\begin{equation}
\text{AP@50} = \frac{1}{|\mathcal{D}|} \sum_{d \in \mathcal{D}} \text{Precision}(d) \times \text{Recall}(d)
\end{equation}

where \( \mathcal{D} \) is the set of detections, and Precision and Recall are computed at an IoU threshold of 0.5. AP@50 gives an overall measure of how well the model detects objects with minimal overlap criteria.

\subsection*{J.4 mIoU}

In our segmentation and change detection tasks, we employ mean Intersection over Union (mIoU) as the evaluation metric. mIoU is widely used in semantic segmentation and change detection tasks as it provides a comprehensive assessment of the model's ability to correctly predict class boundaries across the entire dataset. The mIoU is defined as follows:

Mean Intersection over Union (mIoU)
The mIoU measures the overlap between predicted and ground-truth regions for each class, averaged over all classes. It is computed as:

\begin{equation}
\text{mIoU} = \frac{1}{C} \sum_{c=1}^{C} \frac{\text{TP}_c}{\text{TP}_c + \text{FP}_c + \text{FN}_c}
\end{equation}

where \( C \) is the total number of classes, \( \text{TP}_c \) (True Positives) represents the correctly predicted pixels for class \( c \), \( \text{FP}_c \) (False Positives) represents the pixels incorrectly predicted as class \( c \), \( \text{FN}_c \) (False Negatives) represents the pixels of class \( c \) that were not correctly predicted.

In segmentation, mIoU evaluates the overlap quality of predicted regions with respect to the ground truth for each class, making it an effective metric for measuring model performance in pixel-wise classification. For change detection, mIoU helps quantify the accuracy of predicted change regions by comparing the overlap between predicted change areas and actual change areas in the ground truth. This metric is particularly useful in remote sensing applications where precise boundary alignment is essential for detecting changes over time. Using mIoU for both segmentation and change detection allows us to assess how well the model captures class-specific and change-specific information, providing a reliable measure of performance in spatially complex remote sensing tasks.

\clearpage

\end{document}